\documentclass{article}

\usepackage{microtype}
\usepackage{graphicx}
\usepackage{subfigure}
\usepackage{booktabs} 
\usepackage{enumitem}

\usepackage{hyperref}
\usepackage{comment}

\usepackage[accepted]{icml2023}

\usepackage{amsmath}
\usepackage{amssymb}
\usepackage{mathtools}
\usepackage{amsthm}
\usepackage{mathtools}
\usepackage{xspace}
\usepackage{booktabs}
\usepackage{multirow}
\definecolor{asparagus}{rgb}{0.53, 0.66, 0.42}
\usepackage[capitalize,noabbrev]{cleveref}

\theoremstyle{plain}
\newtheorem{theorem}{Theorem}[section]

\theoremstyle{definition}

\theoremstyle{remark}

\usepackage[textsize=tiny]{todonotes}

\icmltitlerunning{Hyperbolic Representation Learning: Revisiting and Advancing}

\begin{document}
\newcommand{\method}{HIE\xspace}
\newcommand{\hlms}{hyperbolic models\xspace}
\newcommand{\Hlms}{Hyperbolic models\xspace}
\newcommand{\hmp}{HC\xspace}
\newcommand{\hdo}{HDO\xspace}
\newcommand{\myalgorithmic}{\textbf}

\twocolumn[
\icmltitle{Hyperbolic Representation Learning: Revisiting and Advancing}
\icmlsetsymbol{equal}{*}

\begin{icmlauthorlist}
\icmlauthor{Menglin Yang}{CUHK}
\icmlauthor{Min Zhou}{log}
\icmlauthor{Rex Ying}{Yale}
\icmlauthor{Yankai Chen}{CUHK}
\icmlauthor{Irwin King}{CUHK}

\end{icmlauthorlist}

\icmlaffiliation{CUHK}{Department of Computer Sciences and Engineering, The Chinese University of Hong Kong}
\icmlaffiliation{log}{LOGS AI}
\icmlaffiliation{Yale}{Yale University}

\icmlcorrespondingauthor{Menglin Yang}{mlyang.cuhk@outlook.com}

\icmlkeywords{Hyperbolic Representation Learning, Neural Network, Graph Neural Network}

\vskip 0.3in
]



\printAffiliationsAndNotice{}  

\begin{abstract}
The non-Euclidean geometry of hyperbolic spaces has recently garnered considerable attention in the realm of representation learning. 
Current endeavors in hyperbolic representation largely presuppose that the underlying hierarchies can be automatically inferred and preserved through the adaptive optimization process.
This assumption, however, is questionable and requires further validation.
In this work, we first introduce a position-tracking mechanism to scrutinize existing prevalent \hlms, revealing that the learned representations are sub-optimal and unsatisfactory. 
To address this, we propose a simple yet effective method, hyperbolic informed embedding (HIE), by incorporating cost-free hierarchical information deduced from the hyperbolic distance of the node to origin (i.e., induced hyperbolic norm) to advance existing \hlms. 
The proposed method HIE is both task-agnostic and model-agnostic, enabling its seamless integration with a broad spectrum of models and tasks. 
Extensive experiments across various models and different tasks demonstrate the versatility and adaptability of the proposed method. Remarkably, our method achieves a remarkable improvement of up to 21.4\% compared to the competing baselines.
\end{abstract}

\section{Introduction}
\label{submission}

Hyperbolic space, considered as the continuous analogue of discrete trees~\cite{2010hyperbolic}, exhibits a natural advantage for modeling data with implicit or explicit tree-like layouts, such as hierarchical structures and power-law distributed data~\cite{adcock2013tree,zhou2022telegraph}.
The superiority of hyperbolic space in representation learning, e.g., low distortion~\cite{sarkar2011low}, small generalization error~\cite{suzuki2021generalization1,suzuki2021generalization2}, and impressive performance~\cite{nickel2017poincare,hgcn2019,yang2022hicf}, has been extensively validated in recent works, spanning a plethora of research topics and downstream applications~\cite{peng2021hyperbolic,yang2021hyper,mettes2023hyperbolic}, including graph learning, image, and text understanding.

\begin{figure}[t]
  \centering
  \includegraphics[width=0.9\linewidth]{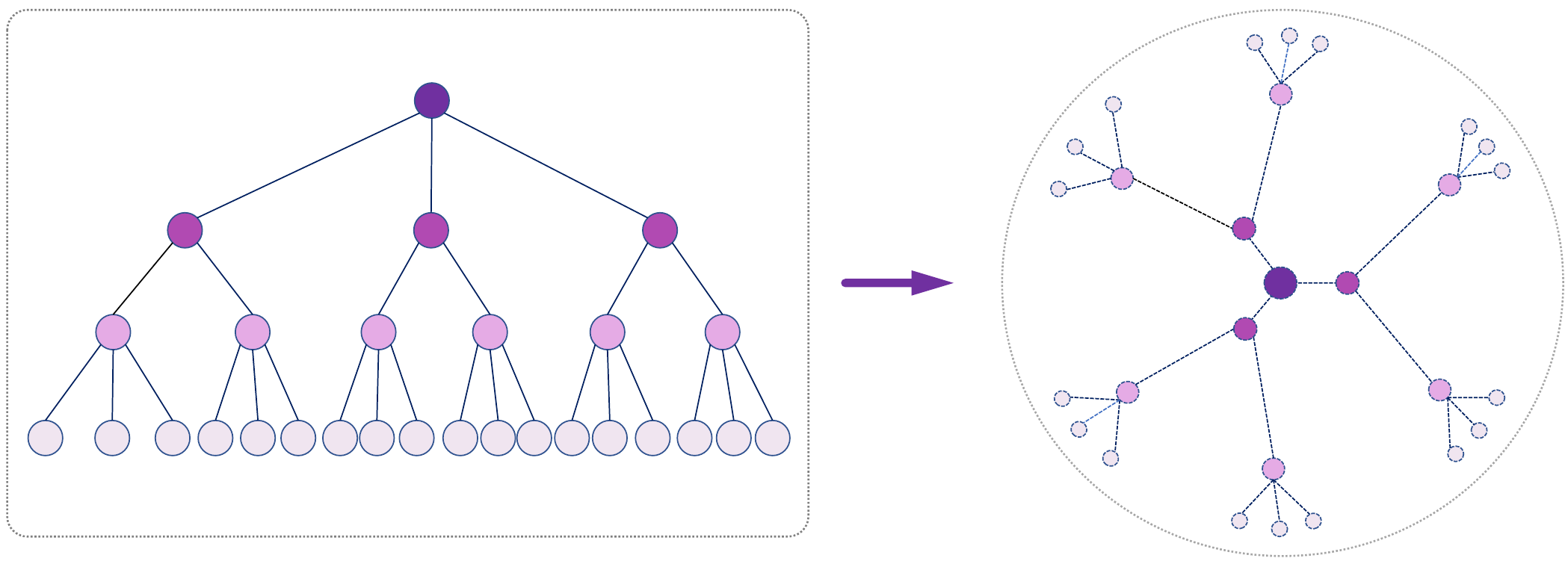}
  \caption{Illustration of an optimal embedding in hyperbolic space that preserves the local dependencies while maintaining a hierarchical relationships of the entities. Left: A tree-like graph. Right: Hyperbolic embedding of the tree-like graph.}
  \label{fig:ideal_example}
  \vspace{-20pt}
\end{figure}

An essential implicit and unspoken objective in utilizing hyperbolic space is to extract intrinsic hierarchical information within data. 
The expected learning objective, as illustrated in \figureautorefname~\ref{fig:ideal_example}, involves optimizing parent nodes above their respective descendants, or from a global perspective, to push the root closer to the hyperbolic origin while simultaneously situating leaf nodes at a greater distance from the hyperbolic origin. This can be understood from the following two aspects.
{Intuitively}, in original data, the parent node occupies a higher hierarchical level than its offspring. Consequently, in the learning space, a high-fidelity embedding necessitates the parent node being positioned relatively higher position, or equivalently in smaller distance to the hyperbolic origin. 
{Mathematically}, let us consider the Poincar\'e ball model as an illustrative example, which is the most frequently used hyperbolic geometry models\footnote{The term ``hyperbolic geometry models'' denotes the mathematical formations for building hyperbolic geometry and creating hyperbolic space, e.g., Poincar\'e ball model and Lorentz model.}. 
When the root node of a tree-like data structure is placed at the hyperbolic origin, it manifests a relatively small distance to all other nodes, as its norm is equal to zero. 
Meanwhile, leaf nodes typically locate closer to the periphery of the ball, as the distance between points increases rapidly as the norm approaches one, thus achieving an optimal and low-distortion embedding that effectively preserves the underlying hierarchical structure~\cite{nickel2017poincare, nickel2018learning}.

The majority of prior studies~\cite{nickel2017poincare,nickel2018learning, HNN, HNN++, hgcn2019,liu2019HGNN,zhang2021hyperbolic} make the assumption that \hlms\footnote{The term ``\hlms'' denotes deep or machine learning models that operate within hyperbolic space, including hyperbolic shallow models, hyperbolic neural networks, and hyperbolic graph neural networks.} are capable of inferring continuous hierarchies from pairwise similarity measurements~\cite{nickel2017poincare,nickel2018learning} or cross-entropy loss~\cite{hgcn2019,lgcn} in a self-optimization process.
However, in the training process, the lack of prior geometric information, such as knowledge about the root, leaves, and hierarchical order, as well as the optimization with geometric-irrelevant objectives raise questions regarding the ability of hyperbolic models to arrange objects in a hierarchical manner and effectively.

To resolve this confusion, we initiate a thorough investigation into the prevalent \hlms
with the intent of ascertaining whether the inherent or latent hierarchical structure is adequately reflected in the resulting embeddings.
We conduct an analysis based on a specially designed position-tracking approach, which is detailed in Section~\ref{sec:tracking embedding position}.
The observations indicate that existing prevalent \hlms fall short in effectively preserving hierarchies and capturing the underlying data structure. 
From a holistic perspective, the root and a significant proportion of leaf nodes are also optimized in unsatisfactory locations, lacking adequate separation in the hyperbolic space. 
To solve these problems, unlike previous works~\cite{nickel2017poincare,nickel2018learning,hgcn2019,liu2019HGNN,HNN,HNN++,lgcn} that solely projected objects into hyperbolic space without considering hierarchical knowledge and assuming that hierarchical dependencies can be automatically extracted in the learning process, we introduce cost-free geometric information to enhance the hierarchical learning process. 
More specifically, we harness the intrinsic geometric characteristics of hyperbolic space, namely hyperbolic distance to origin (HDO), to detect crucial geometric information, such as the root and hierarchical level, and subsequently align the root and ordering. The proposed method is both data and model-agnostic, enabling its facile application to a plethora of datasets and models.

To summarize, our work makes four key contributions:
\begin{itemize}[leftmargin=*]
    \item We propose a position-tracking strategy to investigate current prevalent \hlms, revealing a significant discrepancy between the hyperbolic learning process and conventional understanding, shedding light on the process of hyperbolic representation learning.
    \item We introduce a novel method for inferring implicit hierarchy from hyperbolic embeddings. This approach is cost-free, scalable, and highly effective as it extracts hierarchical information directly from the embeddings themselves, eliminating the need for additional inputs or annotations.
    \item We present a simple yet effective approach that leverages the inferred hierarchy for advancing hyperbolic representation learning. Our method seamlessly integrates with existing hyperbolic models and does not introduce any additional model parameters, making it practical and easy to implement.
    \item We conduct extensive experiments on various \hlms, demonstrating the effectiveness of our proposed method. The results show significant improvements over the baselines, with the largest improvement reaching 21.4\%.
\end{itemize}

\section{Preliminary}

\subsection{Related Works}
The field of neural networks has seen a growing interest in the incorporation of hyperbolic geometry~\cite{peng2021hyperbolic,yang2022hyperbolic,zhou2022hyperbolic,vyas2022graphzoo,xiong2022hyperbolic}, in areas like lexical entailment~\cite{nickel2017poincare,gulcehre2019hyperbolicAT,sala2018representation}, knowledge graphs~\cite{chami2020low,bai2021modeling,sun2020knowledge,xiong2022ultrahyperbolic}, image understanding~\cite{khrulkov2020hyperbolic,HistopathologyHyperbolic,atigh2022hyperbolic,hsu2021capturing}, and recommender systems~\cite{HyperML2020,chen2022modeling,sun2021hgcf,yang2022hrcf,yang2022hicf}. In the realm of graph learning~\cite{gulcehre2019hyperbolicAT,hgcn2019,liu2019HGNN,liu2022enhancing,yang2022hyperbolic}, a significant amount of research works generalizing graph convolutions~\cite{gcn2017,gat2018,graphsage,yang2020featurenorm,yang2022curvature,li2022bsal,zhang2019star} in hyperbolic space for a better graph or temporal graph representation~\cite{hgcn2019,liu2019HGNN,lgcn,yang2021discrete,yang2022hyperbolic,bai2023hgwavenet,bai2022h,sun2021hyperbolic}, which has achieved impressive performance.

The choice of optimization targets or loss functions in hyperbolic learning models, is typically driven by specific downstream tasks or applications. For example, cross-entropy is utilized in node classification or graph classification tasks~\cite{hgcn2019,liu2019HGNN}, while binary cross-entropy is commonly used for link prediction~\cite{hgcn2019,nickel2018learning,HNN}, and ranking margin loss is employed for ranking process in recommender systems~\cite{sun2021hgcf,HyperML2020,yang2022hicf,wang2021hypersorec}. 
However, many of these loss functions are task-specific and locally defined, without explicitly considering the underlying geometry of the hyperbolic space. 
This raises a concern regarding how hyperbolic learning approaches can effectively capture and learn the hierarchical structure of data without explicit or implicit guidance on hierarchical information.
 

\subsection{Riemannian Geometry and Hyperbolic Space}
\label{sec:Riemannian geometry and hyperbolic space}
Here are some preliminary {definitions} for Riemannian geometry and hyperbolic space. We refer the interested readers to ~\cite{spivak1973comprehensive,JohnMLee1997RiemannianMA,lee2013smooth} for more details.

\textbf{Riemannian Geometry.} 
A Riemannian manifold $\mathcal{M}$ with $d$ dimensions is a topological space that possesses a metric tensor $g$, denoted as $(\mathcal{M}, g)$. At any arbitrary point $\mathbf{x}$ in $\mathcal{M}$, the manifold can be locally approximated by a local linear space called the tangent space $\mathcal{T}_\mathbf{x}\mathcal{M}$, which is isometric to $\mathbb{R}^d$.
The shortest path between two points on the manifold is known as a geodesic, and its length is referred to as the induced distance $d_\mathcal{M}$.
In particular, the distance of a node to the origin in hyperbolic space (HDO) can be construed as its corresponding induced hyperbolic norm.
To project a tangent vector onto the manifold, the exponential map $\exp_\mathbf{x}$ is utilized, while its inverse function is known as the logarithmic map $\log_\mathbf{x}$. Another significant operation is parallel transport $P_{\mathbf{x}\to\mathbf{y}}$, which allows for the transportation of geometric data from $\mathbf{x}$ to $\mathbf{y}$ along the unique geodesics while preserving the metric tensors.

\textbf{Hyperbolic Space.} Hyperbolic space refers to a Riemannian manifold with a constant negative curvature, and its coordinates can be represented using various isometric models\footnote{\url{https://en.wikipedia.org/wiki/Hyperbolic_space}}. In the fields of machine learning and deep learning, the Poincar\'e ball model and the Lorentz model are widely studied.
     A $d$-dimensional \textit{Poincar\'e ball model} with a constant negative curvature $\kappa (\kappa<0)$ is defined as a Riemannian manifold ($\mathcal{B}_\kappa^d, g_{\mathcal{B}}^\kappa$), where $\mathcal{B}_\kappa^d=\{\mathbf{x}\in\mathbb{R}^d~|~ \|\mathbf{x}\|^2<-1/\kappa\}$ and its metric tensor $g_{\mathcal{B}}^\kappa = (\lambda_\mathbf{x}^\kappa)^2\mathbf{I}_d$ and $\mathbf{I}_d$ is $d$ dimensional identity matrix. 
    Here, $\mathcal{B}_\kappa^d$ is an $d$-dimensional open ball of radius $(-\kappa)^{-\frac{1}{2}}$, and $\lambda_\mathbf{x}^\kappa=2(1+\kappa\|\mathbf{x}\|_2^2)^{-1}$ is the conformal factor.
    The Lorentz model of the hyperbolic space is another popular model. In this model, the space is visualized as one sheet of a two-sheeted hyperboloid in a ($d$+1)-dimensional Minkowski space. 
    A $d$-dimensional \textit{Lorentz model} (also called hyperboloid model) with a constant negative curvature $\kappa(\kappa<0)$ is defined as a Riemannian manifold ($\mathcal{L}_\kappa^d, g_{\mathcal{L}}^\kappa$), where $\mathcal{L}_\kappa^d = \{\mathbf{x}\in\mathbb{R}^{d+1}~|~\langle\mathbf{x},\mathbf{x}\rangle_\mathcal{L}= {1}/{\kappa}\}$ and its metric tensor $g_{\mathcal{L}}^\kappa = \mathrm{diag}([-1,1,\cdots,1])$. 
    It is important to note that the study applies to both the Poincaré ball model and the Lorentz model, and for simplicity, a unified notation $\mathcal{H}$ is used. For the hyperbolic formula about exponential map $\exp_\mathbf{x}^\kappa$, logarithmic map $\log_{\mathbf{x}}^\kappa$, hyperbolic distance $d_\mathcal{H}$, and parallel transport $P_{\mathbf{o}\to \mathbf{x}}^\kappa$, and other related information, please refer to Appendix~\ref{appendix:Riemannian_geometry}.

\subsection{Hyperbolic Shallow Models}
The hyperbolic shallow models encode entities into hyperbolic space and utilize the relative distance or similarity as a means to infer their ordering. For example, \citet{nickel2017poincare,nickel2018learning} proposed to embed data in Poincar\'e ball model as well as Lorentz model to learn embeddings by optimizing the following objective:
\begin{equation}
    \max_{\mathbf{x}\in\mathcal{H}^d} {\sum_{{(i,j)}\in E}} \log \frac{\exp{(-d_\mathcal{H}(\mathbf{x}_i,\mathbf{x}_j)})}{\sum_{j'\in {Neg}(i)}\exp{(-d_\mathcal{H}(\mathbf{x}_i,\mathbf{x}_{j'})})},
\end{equation}
where $E$ is a set consisting of linked or related object pairs, ${Neg}(i) = \{j|(i,j)\notin E\}\cup\{i\}$ is the set of negative examples for $i$, $\mathbf{x}\in\mathcal{H}^d$ is trainable node embedding ($\mathbf{x}_i$ indicates the specific object $i$), and $d_\mathcal{H}$ is the hyperbolic distance. The optimizing target encourages connected or related object pairs in the original structured data to have a small hyperbolic distance while pushing unconnected or unpaired nodes far apart.


\begin{figure*}[!tp]
\small
\centering
\subfigure{
\begin{minipage}[t]{0.290\linewidth}
\centering
\includegraphics[width=1.4in]{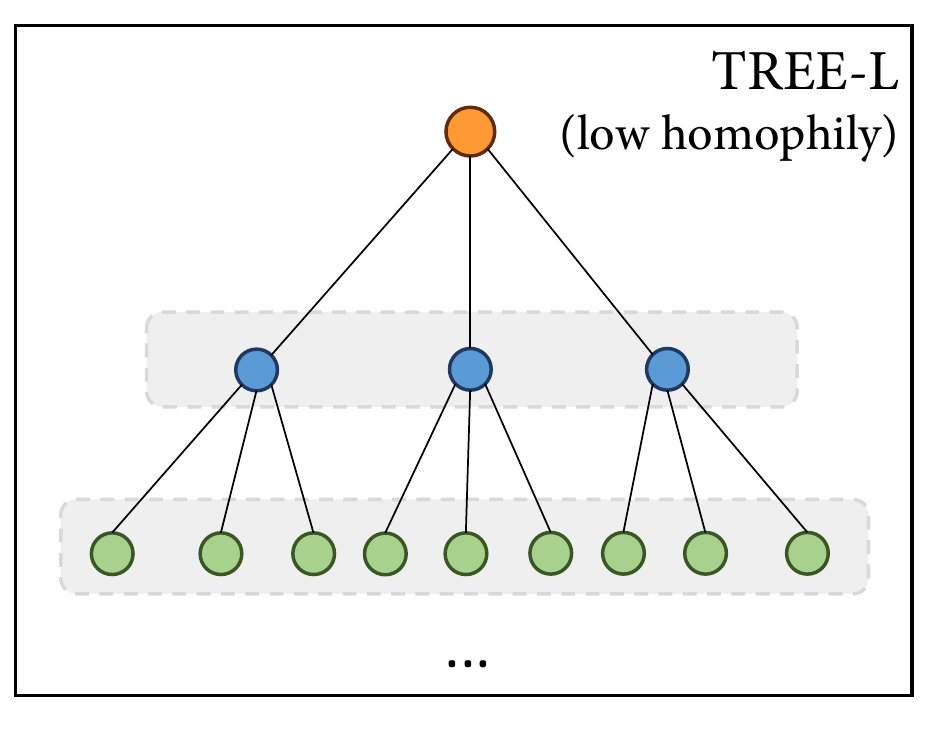}
\end{minipage}%
}%
\subfigure{
\begin{minipage}[t]{0.290\linewidth}
\centering
\includegraphics[width=1.5in]{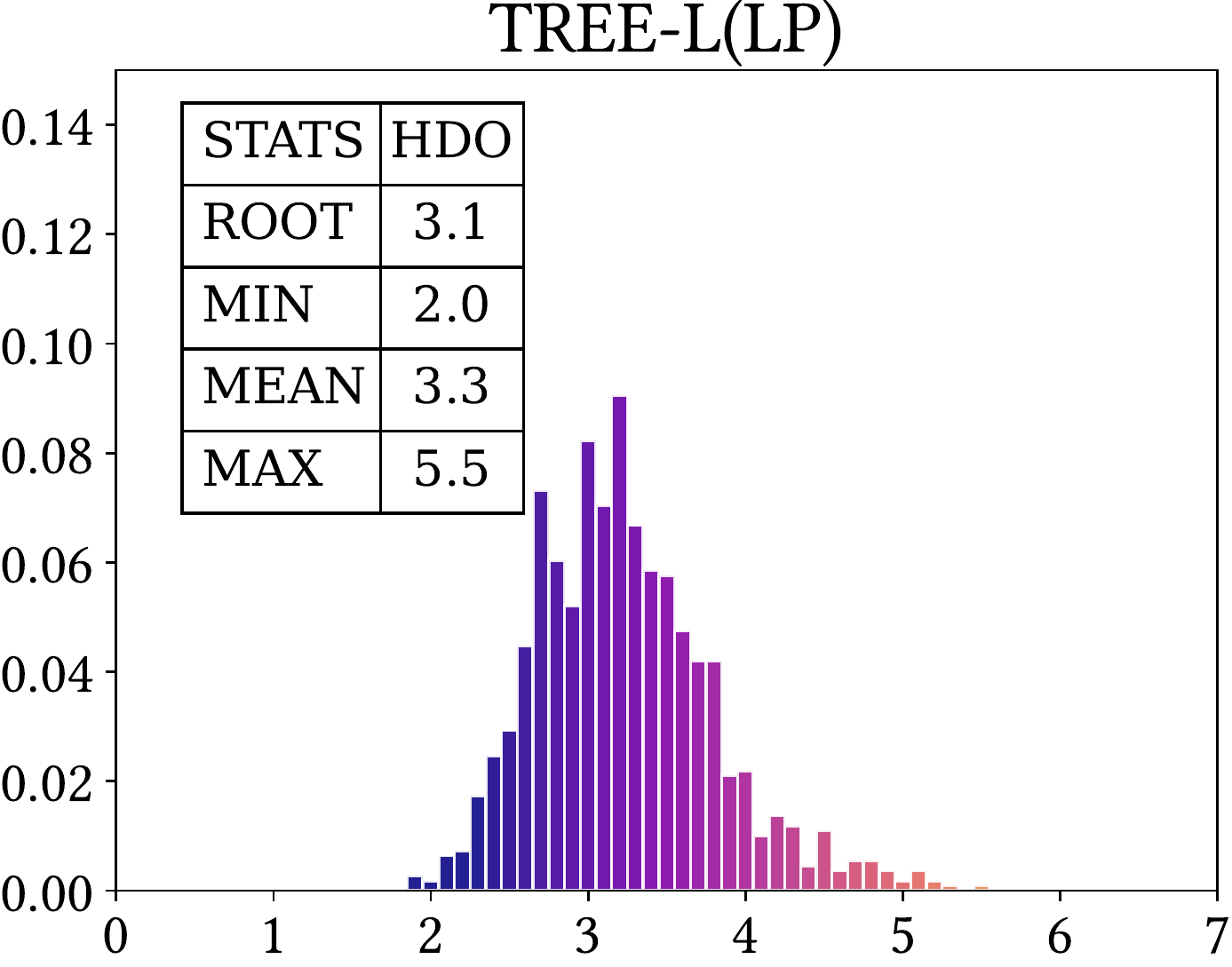}
\end{minipage}%
}%
\subfigure{
\begin{minipage}[t]{0.290\linewidth}
\centering
\includegraphics[width=1.5in]{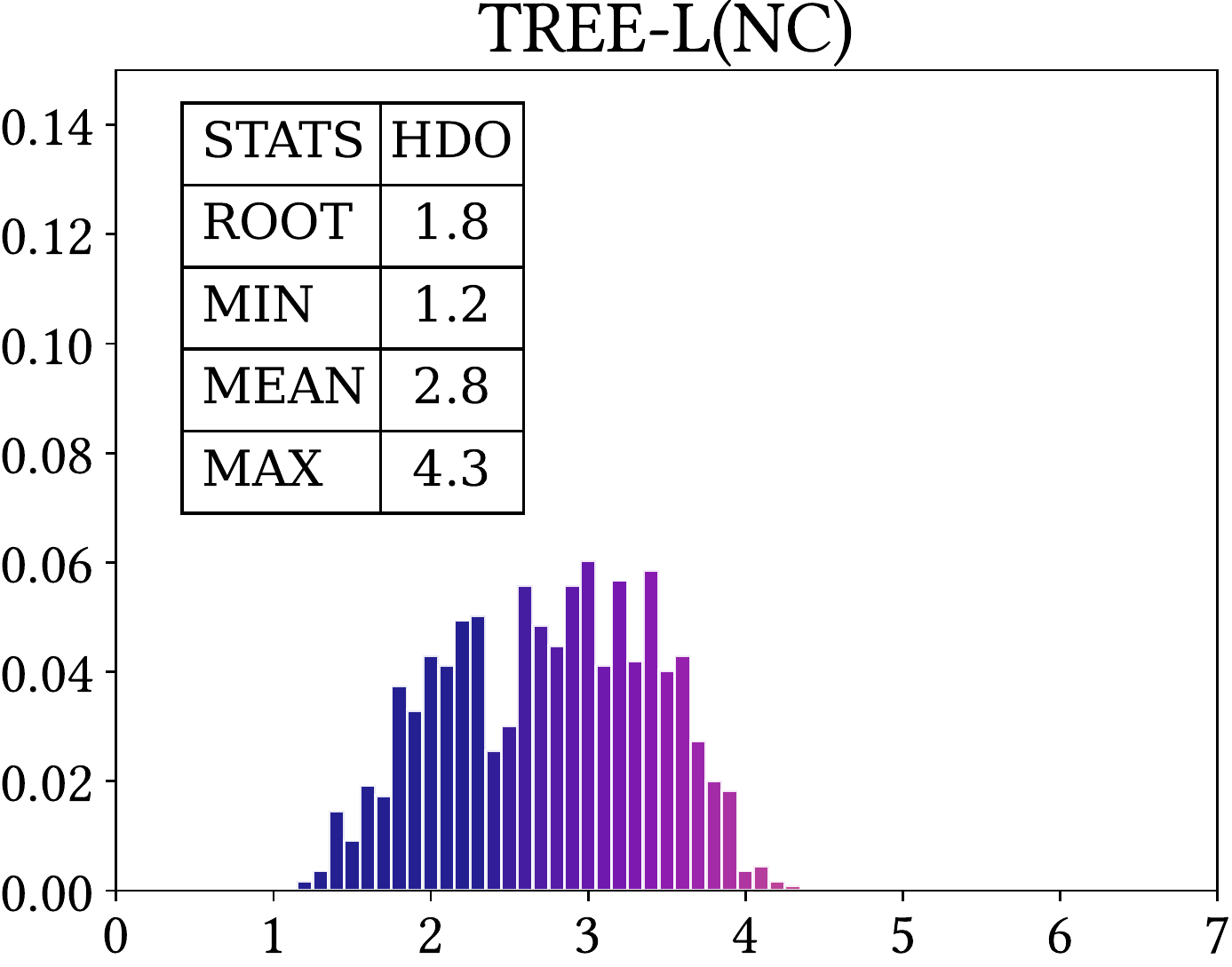}
\end{minipage}
}

\subfigure{
\begin{minipage}[t]{0.290\linewidth}
\centering
\includegraphics[width=1.4in]{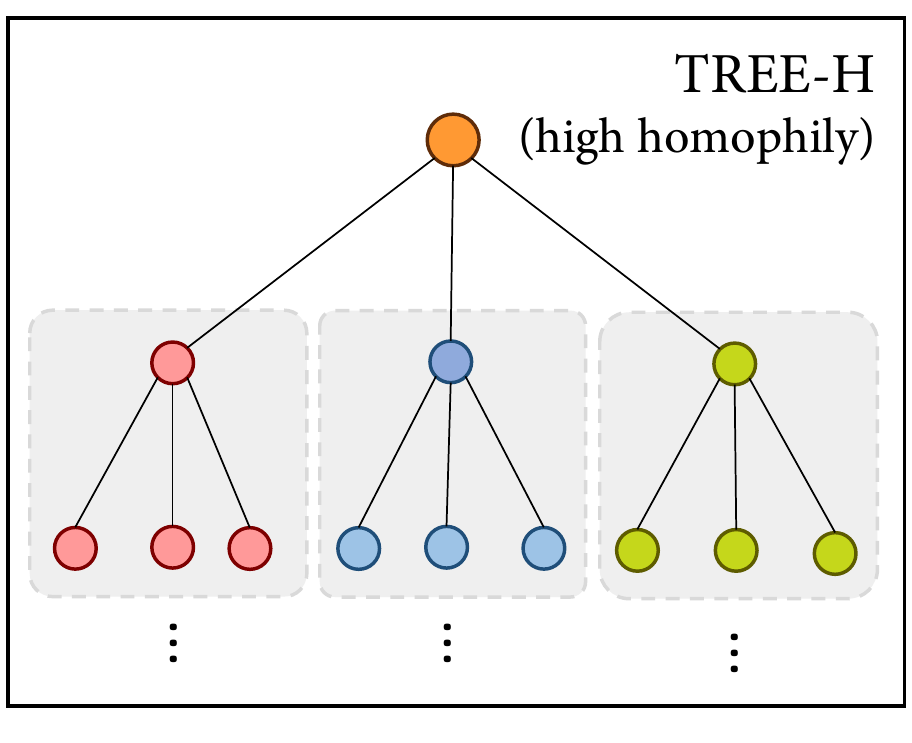}
\end{minipage}%
}%
\subfigure{
\begin{minipage}[t]{0.290\linewidth}
\centering
\includegraphics[width=1.5in]{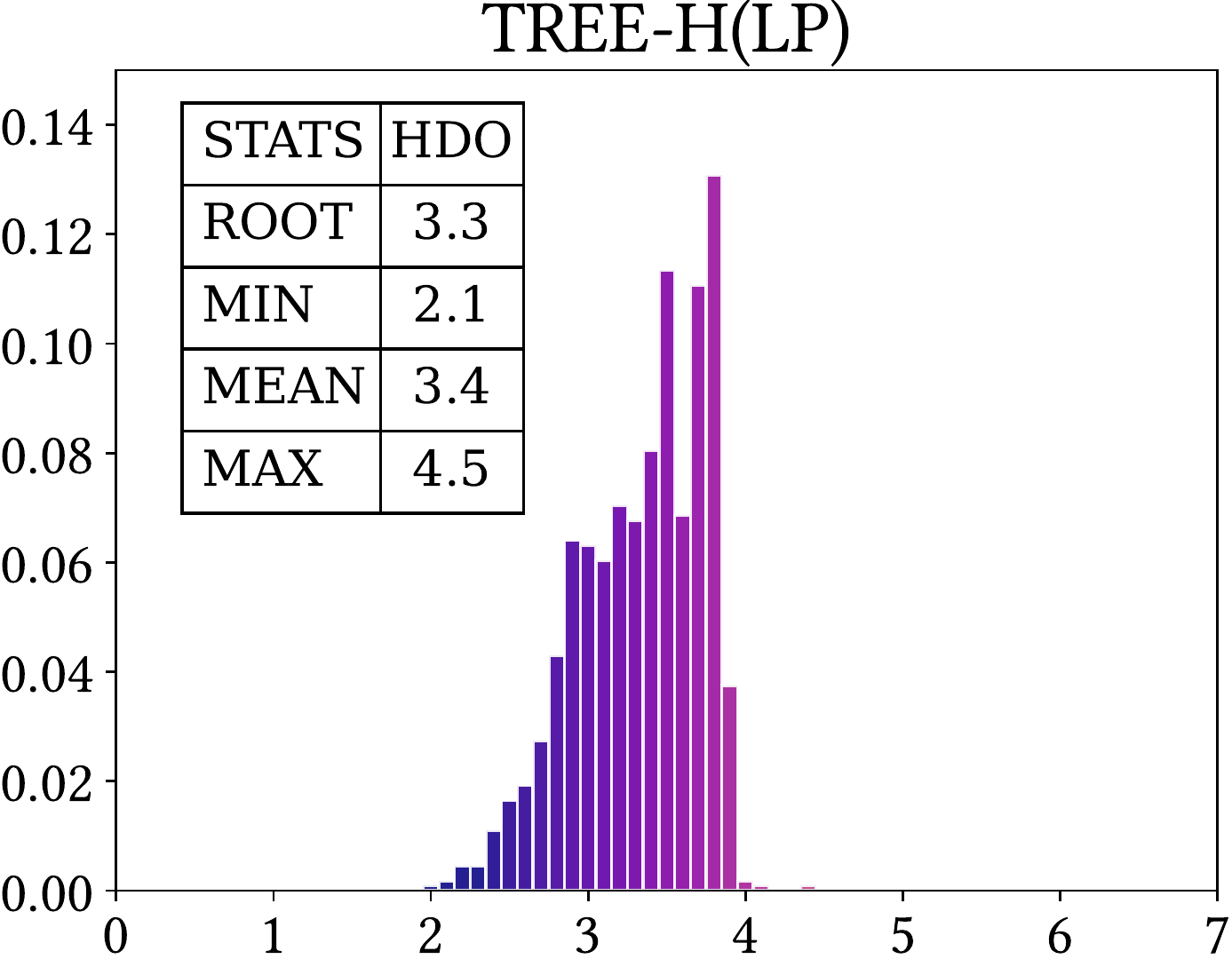}
\end{minipage}%
}%
\subfigure{
\begin{minipage}[t]{0.290\linewidth}
\centering
\includegraphics[width=1.5in]{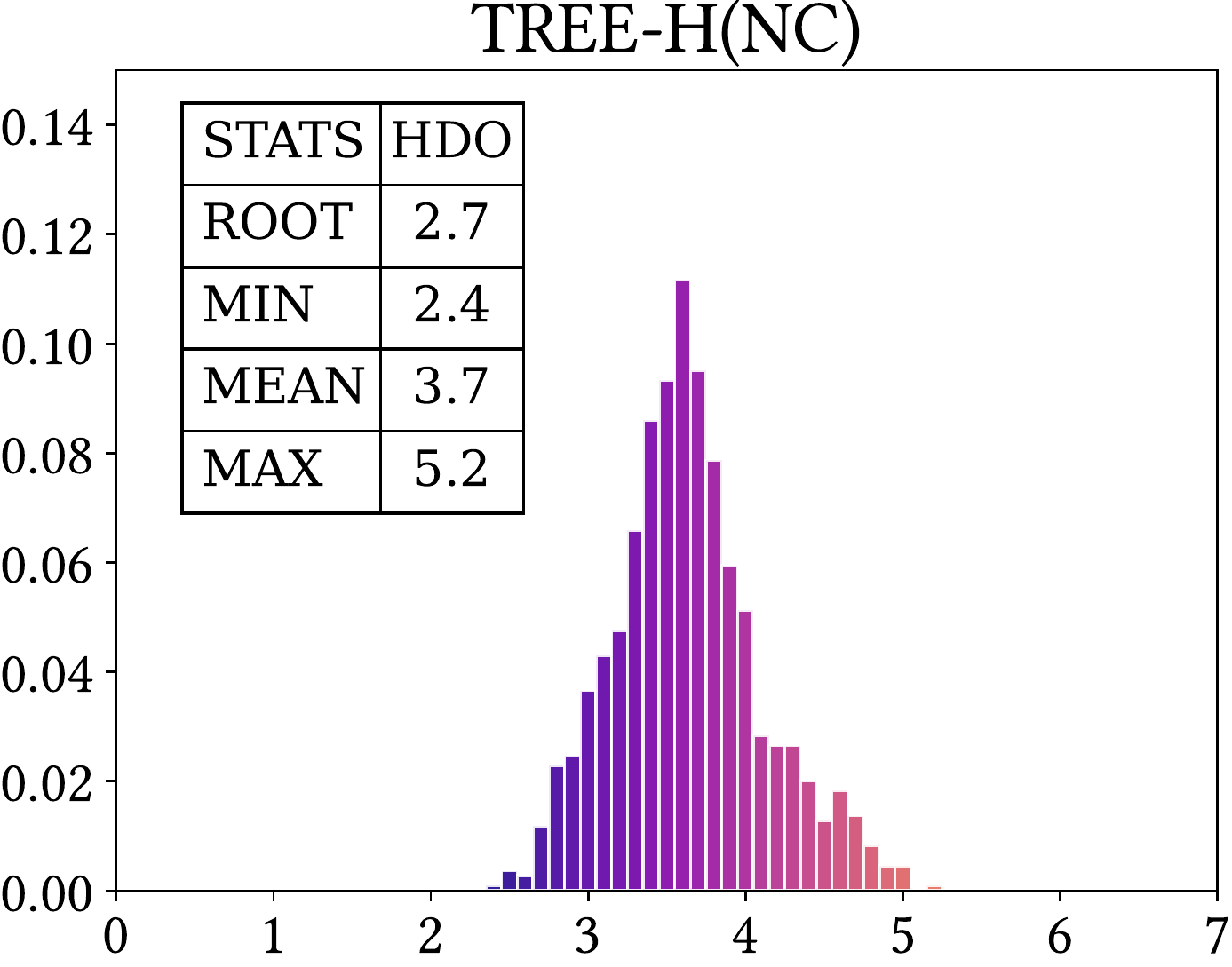}
\end{minipage}%
}%
\vspace{-10pt}
\centering
\caption{Illustration of the structure of synthetic TREE-L/H (first column), and the corresponding distribution of \hdo by HGCN on link prediction (LP) task (second column) and node classification (NC) (third column). 
}
\label{fig:syn_tree_exp}
\vspace{-10pt}
\end{figure*}
\subsection{Hyperbolic Neural Networks}
Hyperbolic neural networks (HNNs)~\cite{HNN,HNN++} represent a generalization of traditional neural networks in hyperbolic space. The fundamental architecture of the HNN encompasses several components, including multinomial logistic regression (MLR), fully-connected (FC) layers, and recurrent neural networks (RNNs). 
For brevity, we mainly focus on the core module, namely FC layer, in this study. The corresponding hyperbolic FC is composed of a linear transformation and non-linear activation.
Hyperbolic linear transformation begins with matrix multiplication, followed by bias translation. In general, this process can be achieved in the tangent space of origin~\cite{HNN,liu2019HGNN,hgcn2019,lgcn}. Given a hyperbolic point $\mathbf{x}\in\mathcal{H}^d$, it is first  projected to the tangent space of origin $\mathcal{T}_\mathbf{o}\mathcal{H}^d$ using the logarithmic map and multiplied by the weight matrix, $\mathbf{W}\in\mathbb{R}^{d'\times d}$. Hyperbolic matrix multiplication is defined as, $\mathbf{W}\otimes^\kappa\mathbf{x} := \exp_\mathbf{o}^\kappa(\mathbf{W}\log_\mathbf{o}^\kappa(\mathbf{x}))$.
For bias addition, let us use a bias $\mathbf{b}$ located at $\mathcal{T}_\mathbf{o}\mathcal{H}^d$, parallel transport it to the hyperbolic point of interest, and map it to the manifold,
$
    \mathbf{x}\oplus^\kappa\mathbf{b}:=\exp_{\mathbf{x}}^\kappa(P_{\mathbf{o}\to\mathbf{x}}^\kappa(\mathbf{b})).
$
The non-linear activation is also achieved at the tangent space, and for simplicity, we use hyperbolic origin, and it is given as,
$
    \sigma_\mathcal{H}^{\kappa_1,\kappa_2}(\mathbf{x}):=\exp_\mathbf{o}^{\kappa_1}{(\sigma(\log_\mathbf{o}^{\kappa_2}(\mathbf{x})})).
$
Finally, given $\mathbf{x} \in\mathcal{H}^d$, the $\ell$-th hyperbolic neural network layer is formulated as:
\begin{equation}
\label{eq:hnn}
    \mathbf{x}^{\ell} = \sigma_\mathcal{H}^{\kappa_{\ell},\kappa_{\ell-1}}{\left(\mathbf{W}\otimes^{\kappa_{\ell-1}}\mathbf{x}^{\ell-1})\oplus^{\kappa_{\ell-1}}\mathbf{b}\right)},
\end{equation}
where $\kappa_\ell$ and $\kappa_{\ell-1}$ is the curvature at layer $\ell$ and $\ell-1$, respectively.

\subsection{Hyperbolic Graph Convolutional Neural Networks}
\label{sec:hgnn}
The cornerstone of Hyperbolic Graph Convolutional Neural Networks (HGCNs) is the execution of graph convolutions within the bounds of hyperbolic space. HGNNs employ the message passing scheme, including linear transformation, neighbors aggregation, and non-linear activation.
The research presented in this paper is applicable to both of the aforementioned cases.
Compared with the HNNs given by Equation~(\ref{eq:hnn}), the HGCN layer has one neighborhood aggregation \textcolor{black}{more}. Let $j\in \mathcal{N}(i)$ be the neighbors of node $i$ with self-loop, the hyperbolic neighborhood aggregation is given as,
$
    \text{{\sc Agg}}^\kappa(\mathbf{x}_i):= \exp_{\mathbf{x}_r}^{\kappa}\left(\sum_{j\in \mathcal{N}{(i)}} \tilde{a}_{ij}\log_{\mathbf{x}_r}^{\kappa}
    ({\mathbf{x}}_j)\right),
    \label{eq:hyperbolic message passing}
$
where $\mathbf{x}_r$ is a reference point in hyperbolic space, which is set as the origin point for simplicity in this study, and $\tilde{a}_{ij}$ is the normalized edge weights, which is computed either by an attention-based method~\cite{hgcn2019}, i.e.,
$
    \tilde{a}_{ij} = \mbox{Softmax}_{j\in \mathcal{N}(i)}(\mbox{MLP}(\log_\mathbf{o}^{\kappa}(\mathbf{x}_i)||\log_{\mathbf{o}}^{\kappa}(\mathbf{x}_j)))
    \label{eq:att-based aggregation}
$
or by the symmetric normalized Laplacian matrix~\cite{gcn2017,liu2019HGNN}, i.e.,
$
\label{eq:degree-based aggregation}
    {\tilde{a}_{ij} = 1/{\sqrt{\tilde{d}_i\tilde{d}_j}}},
$
where the $\tilde{d}_v$ denotes the degree of node $v$ including the self-loop.
Then, one HGNN layer is formulated as:
\begin{equation}
\begin{aligned}
\mathbf{x}_{i}^{\ell} &=\sigma_\mathcal{H}^{{\kappa_{\ell-1}, \kappa_{\ell}}}(\tilde{\mathbf{h}}_{i}^{\ell}),\\
\tilde{\mathbf{h}}_{i}^{\ell} &=\textsc{Agg}^{\kappa_{\ell-1}}(\mathbf{h}^{\ell}_{i}),\\
\mathbf{h}_{i}^{\ell} &=(\mathbf{W}^{\ell} \otimes^{\kappa_{\ell-1}} \mathbf{x}_{i}^{\ell-1}) \oplus^{\kappa_{\ell-1}} \mathbf{b}^{\ell}.
\end{aligned}
\end{equation}

\begin{algorithm}[h]
\caption{Hyperbolic Informed Embedding (HIE)}
\label{alg:hie}
\begin{algorithmic}[1]
    \STATE \textbf{Input:} (i) Input data $\mathcal{X}$ which can be either $n$ general input samples $\mathcal{X}=\{\mathbf{x}_1, \cdots \mathbf{x}_n\}$ or $n$ nodes in a graph with adjacency matrix, i.e., $\mathcal{X}=\{\mathbf{x}_1, \cdots \mathbf{x}_n, \mathbf{A}\}$; $\mathbf{x}_i$ can be randomly sampled and trainable or pre-defined.
    (ii) Learning model $\mathcal{F}$ which is optionally parameterized by $\mathbf{W}$; 
    (iii) Hyperparameter $\lambda>0$.
    \STATE \textbf{Output:} Embeddings of $n$ objects and optimal $\mathbf{W}$.
    \STATE Initialize input data $\mathcal{X}$ and learnable weight $\mathbf{W}$ (optional) of $\mathcal{F}$;
    \STATE Compute embedding $\mathbf{Z} \leftarrow \mathcal{F}([\mathbf{W}],\mathcal{X})$;
    \IF{partial root-alignment}
    \STATE Calculate task-specific loss: $L_\text{task}$;
    \STATE {\color{black!50!blue}Calculate root aligned embedding $\bar{\mathbf{Z}}$};
    \STATE {\color{black!50!blue}Calculate hyperbolic distance to origin loss:$L_\text{hyp}$};
    \STATE Optimize with overall loss function:
    $L_{\text{task}}+\lambda L_\text{hyp}$;
    \STATE {Return} $\mathbf{Z}$,$\mathbf{W}$
    \ENDIF
    \IF{whole root-alignment}
    \STATE {\color{black!50!blue}Calculate root aligned embedding$\bar{\mathbf{Z}}$};
    \STATE {\color{black!50!blue}Calculate hyperbolic distance to origin loss:$L_\text{hyp}$};
        \STATE {Calculate task-specific loss: $L_\text{task}$};
    \STATE Optimize with overall loss function:
    $L_{\text{task}}+\lambda L_\text{hyp}$;
    \STATE {Return} $\bar{\mathbf{Z}}$,$\mathbf{W}$
    \ENDIF
\end{algorithmic}
\end{algorithm}
\section{Investigation}
\label{sec:tracking embedding position}
\subsection{Hyperbolic Distance to Origin (\hdo)}
\label{sec:introdo-of-hdo}
Before delving into the analysis of the tracking of embedding trajectory, it is imperative to introduce an essential tool for the investigation.

The hyperbolic distance from a point $\mathbf{x}$ to the origin $\mathbf{o}$ (\hdo, also referred to as the induced hyperbolic norm), $d_\mathcal{H}(\mathbf{x}, \mathbf{o})$, has found considerable usage as an interpretative post-hoc indicator, particularly for explicating the intricacies of embedding levels across diverse domains. For instance, in WordNet\footnote{\url{https://wordnet.princeton.edu/}} embedding, \citet{nickel2017poincare} employed the difference of embedding norms to define a score function for evaluating the effectiveness of Poincar\'e embedding in representing semantic relationships. 
In image embedding, \citet{khrulkov2020hyperbolic} incorporated \hdo as an uncertainty quantifier and found that hard-to-classify images are embedded near the hyperbolic origin, whereas easy-to-classify images are placed around the boundary of the Poincar\'e. Hard-to-classify images often encompass patterns from various categories, which are associated with high-level positions in the hierarchy. Conversely, the easy-to-classify images exhibit specific patterns, which are related to low-level positions in the hierarchy. A subsequent study by~\citet{sun2021hgcf} leveraged \hdo to analyze the alignment of hyperbolic user-item embeddings with popularity. Their analysis concluded that smaller HDO values were indicative of higher popularity levels. A higher popularity of a item means that the item is favored by a substantial users, with the most popular item functioning as a network hub or a tree root.

In a $k$-regular tree, the number of nodes at level $r$ proliferates exponentially, specifically, $(k+1)k^{r-1}$, and the number of nodes at levels less than or equal to $r$ also follows a similar trend, specifically, $((k+1)k^{r}-2)/(k-1)$. This implies that the number of nodes escalates with the distance from the root of the tree. Analogously, in hyperbolic space, the area of a disc also expands exponentially with distance from the hyperbolic origin. Therefore, strategically placing the nodes of level $r$ of a tree at a distance $R$ (where $R$ is proportionate to $r$)~\cite{2010hyperbolic,krioukov2009curvature} from the hyperbolic origin results in a hierarchical embedding that captures the underlying tree-like structure.
Building on this insight, in this study, we design a position tracking analysis by \hdo. Besides, we also utilize it as a tool to guide representation learning in hyperbolic space.

\subsection{\hdo Investigation}
Due to the lack of crucial geometric information, such as the root label, level information, and leaf nodes, in real-world graph datasets, we synthesized two pure-tree datasets. To ensure our findings do not lose generality, we created two types of trees with varying degrees of homophily\footnote{Homophily is a metric used to differentiate assortative and disassortative graphs, as defined by~\cite{gemo-gcn}.}: TREE-H (homophily rate of 0.998) and TREE-L (homophily rate of 0.018). Detailed information regarding the construction of TREE-L and TREE-H can be found in Appendix~\ref{appendix:data_construction_treelh}.  We conduct an experimental evaluation of the HGCN on node classification and link prediction tasks and depict the distribution of the \hdo for the node states in the final embedding layer as illustrated in Figure~\ref{fig:syn_tree_exp}. It is noteworthy that other hyperbolic models have demonstrated comparable results.
Furthermore, we provide essential statistics regarding \hdo, such as the minimum ({MIN}), maximum ({MAX}), mean ({MEAN}), and \hdo of the root node ({ROOT}), in the corresponding subfigures for the convenience of readers.

It can be observed from the experimental results that in both cases, root nodes exhibit a large disparity with the minimum \hdo (\textit{c.f.}, MIN and ROOT values). For instance, in the link prediction task, the ROOTs of TREE-L and TREE-H are 3.1 and 3.3, respectively, while the MINs are 2.0 and 2.1. Moreover, the distribution of \hdo exhibits a more normal pattern. However, in trees, leaf nodes constitute the majority and are expected to be located farther from the origin, following a power-law distribution. 
Moreover, the nodes are not fully spreading, which can be inferred from the shape of the distribution and the gap between min, mean, and max \hdo. It is worth noting that the hyperbolic space expands exponentially, with the area far from the origin being significantly more spacious and accommodating.

Overall, the aforementioned investigation reveals the following: (1) The root node is not optimized to occupy or close to the highest level, potentially resulting in an inaccurate hierarchical structure; (2) The \hdo of the learned embeddings is normally distributed, which inadequately captures the tree-like structure; (3) The overall embeddings are not maximally scattered, indicating that the model fails to fully leverage the expansive nature of the hyperbolic space.
These limitations motivate us to reshape the optimization objective by explicitly incorporating root alignment and hierarchical stretching.


\section{Method}
\label{sec: do}

To address the above problem, we proposed the HIE algorithm which is demonstrated in Algorithm~\ref{alg:hie}. 
HIE can function both as a novel learning paradigm incorporating \texttt{whole root-alignment} settings, and as a flexible plug-in module by \texttt{partial root-alignment}. The core of HIE is highlighted in blue in Algorithm~\ref{alg:hie}, and we provide further details in the following.

\begin{figure}
\centering
\includegraphics[width=0.450\textwidth]{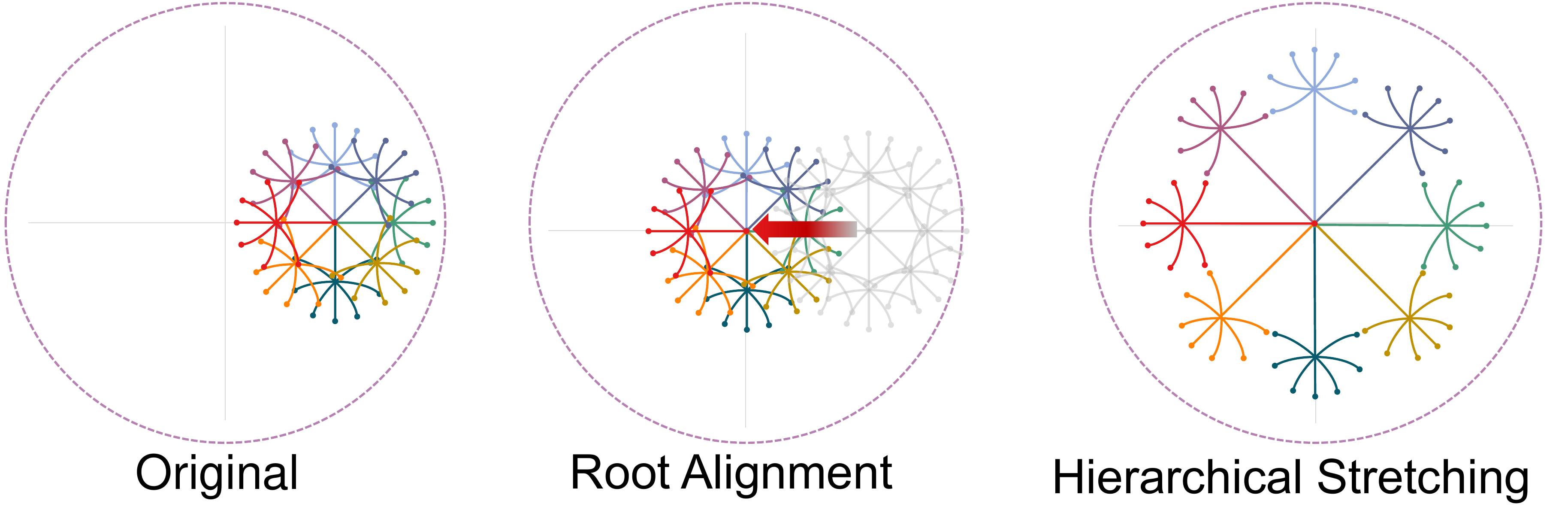}
\centering
\caption{Illustration of the basic idea of \method, which can be decomposed into two critical steps: root alignment and level-aware stretching.}
\label{fig:idea}
\vspace{-15pt}
\end{figure}

As sketched in Figure~\ref{fig:idea}, the intuition of our idea starts with identifying the root node of the overall data and aligning it with the origin of the hyperbolic space. 
Then we optimize each node away from the origin according to its level information. This not only facilitates the formation of the correct hierarchies but also makes full use of the hyperbolic space. However, there are two challenges to achieving this target:\\
\textbf{(i) Defining and locating the root node:} Numerous indicators, such as centrality\footnote{https://en.wikipedia.org/wiki/Centrality}, can describe a node's role in the graph. However, these methods solely rely on the graph topology and overlook the node's features. Additionally, some of these indicators are computationally expensive when dealing with large-scale graphs. Moreover, tree-like datasets often contain multiple subtrees or roots. \textbf{(ii) Efficiently accessing level information and guiding hierarchical learning}: This hierarchical information is rarely on hand in real-world scenarios and is labor-consuming to label in the large-scale dataset.

(1) {To address the first challenge}, we propose 
 utilizing the hyperbolic embedding center (\hmp) as the root node and making an alignment with the hyperbolic origin. HC is computed through a manifold-specific method and serves as the optimal solution for minimizing the sum of squared distances with all nodes, as outlined by Theorem~\ref{theorem: hmp-minimzation}. This conforms to the characteristics of the root node in a regular tree, that is, the node with the smallest sum of distances to other nodes.

\textit{Remark}. Utilizing the \hmp as the root node presents several key advantages: (i) Computation of the \hmp is relatively straightforward, eliminating the need for extensive topology search. The computational cost is $O(|V|)$, or even $O(1)$ when utilizing parallel computing with a GPU, which is significantly more efficient than other centrality computations;
(ii) The \hmp is derived from hyperbolic embeddings, which encode both structural and feature information. It effectively handles scenarios with multiple root nodes since the \hmp always represents a unique point that can be seen as a supernode connecting all subtrees; 
{(iii)} \hmp can be adjusted adaptively according to downstream tasks during the learning process. 

Let $\mathbf{Z}$ be the final layer embedding matrix consisting of $n$ hyperbolic node vectors $(\mathbf{z}_1, \cdots, \mathbf{z}_i, \cdots, \mathbf{z}_n)$ where $\mathbf{z}_i\in\mathcal{H}^d_\kappa$ and let $v_i$ denote the weight assigned to $\mathbf{z}_i$. The function $f_c$ is used for computing the weighted center, and $\mathbf{z}_c$ represents this center, which is computed as:
$
\mathbf{z}_c = f_c(\{\mathbf{z}_i, {v}_i|i\in V\}),
$
where $f_c$ is detailed in Appendix~\ref{appendix:hyperbolic midpoint}. Following this, we employ the \textit{root alignment} strategy as defined in Equation~\eqref{eq:hyperbolic_alignment}.

(2) {To address the second challenge}, we propose a hierarchical stretching of the nodes by leveraging the \hdo, which serves as a means of encapsulating implicit hierarchical information, as discussed in Section~(\ref{sec:introdo-of-hdo}). Specifically, by aligning the hyperbolic embedding center (i.e., the root node) with the origin of the hyperbolic space, the \hdo more accurately reflects the relative distance of a node to the root node, thus indicating its hierarchical level, which is given by Equation~\eqref{eq:hyperbolic_HDOcomputation}. For implicitly \textit{hierarchical stretching}, it is incorporated into the loss function~\eqref{eq:hyperbolic_stretching}.

\begin{table}[!tp]
\vspace{-5pt}
\centering
\caption{Comparisons of shallow models for link prediction task on {\sc Disease}. The first and second rows correspond to the AUC and AP metrics, respectively.}
\vspace{4pt}
\label{tab:shallow_model_comparsion}
\resizebox{0.48\textwidth}{!}{%
\begin{tabular}{@{}l|ccc|ccc@{}}
\toprule
    Link         & \multicolumn{3}{c|}{75\%Training Links}                               & \multicolumn{3}{c}{25\%Training Links}                                \\
 Dim& 8                     & 64                    & 256                   & 8                     & 64                    & 256                   \\ \midrule \midrule
Euclidean    & $61.1\pm2.4$          & $72.2\pm0.5$          & $73.5\pm0.4$          & $53.5\pm0.2$          & $54.3\pm0.3$          & $54.1\pm0.2$          \\
Hyperbolic   & $65.2\pm3.2$          & $74.9\pm0.4$          & $77.0\pm0.3$          & $52.9\pm0.5$          & $53.7\pm0.5$          & $55.0\pm0.4$          \\
Ours         & $\textbf{72.0}\pm0.9$ & $\textbf{80.7}\pm0.4$ & $\textbf{82.5}\pm0.2$ & $\textbf{58.1}\pm1.1$ & $\textbf{63.9}\pm0.4$ & $\textbf{66.8}\pm0.4$ \\
$\Delta$(\%) & $+10.4$               & $+7.7$                & $+7.1$                & $+8.6$                & $+17.7$               & $+21.4$               \\ \midrule
Euclidean    & $58.7\pm1.3$          & $69.7\pm0.5$          & $71.4\pm0.7$          & $52.2\pm0.2$          & $52.8\pm0.6$          & $52.3\pm0.1$          \\
Hyperbolic   & $62.3\pm2.7$          & $71.3\pm0.7$          & $72.7\pm0.4$          & $51.9\pm0.7$          & $52.4\pm0.8$          & $54.1\pm0.7$          \\
Ours         & $\textbf{66.6}\pm1.0$ & $\textbf{75.0}\pm0.4$ & $\textbf{76.4}\pm0.3$ & $\textbf{56.1}\pm1.0$ & $\textbf{62.6}\pm0.6$ & $\textbf{65.2}\pm0.6$ \\
$\Delta$(\%) & $+6.9$                & $+5.2$                & $+5.0$                & $+7.4$                & $+18.7$               & $+20.4$               \\ \bottomrule
\end{tabular}%
}
\vspace{-10pt}
\end{table}
\begin{align}
    &\bar{\mathbf{z}} =  \mathbf{z}\oplus_\kappa (-\mathbf{z}_c), &\quad ({\small \text{root alignment}}) \label{eq:hyperbolic_alignment}\\
    &z_\text{hdo} = \frac{1}{|V|}\sum_{i\in V}w_id_{\mathcal{H}}(\bar{\mathbf{z}}_i, \mathbf{o}), &\quad ({\small\text{HDO computation}}) \label{eq:hyperbolic_HDOcomputation}\\
    &L_\text{hyp} = \sigma(-z_{\text{hdo}}), &\quad ({\small\text{stretching}})\label{eq:hyperbolic_stretching}
\end{align}
where in Equation~\eqref{eq:hyperbolic_alignment}, $\oplus_{\kappa}$ denotes the hyperbolic addition operation and its mathematical expressions is provided in Appendix~\ref{appendix:hyperbolic midpoint}; In Equaiton~\eqref{eq:hyperbolic_HDOcomputation}, $w_i$ indicates the node level in hyperbolic space which is obtained from the HDO ($w_i:=f(d_{\mathcal{H}}(\bar{\mathbf{z}}_i, \mathbf{o}))$) and $f$ is monotonically increasing function, such as linear function, tanh, sigmoid, etc, we use identity function for simplicity; In Equation~\eqref{eq:hyperbolic_stretching},  the stretching is achieved by minimizing $L_\text{hyp}$ with monotonically increasing function $\sigma$ (linear function, tanh, exp, etc).


\textit{Remark}. By minimizing the above loss function, the high-level nodes that are close to the origin will be placed with small weights so that they will not push away; the low-level nodes that are far away from the origin will be given large weights so that they can arrive at a correct position.
In this approach, we optimize by promoting nodes to move away from the origin rather than close to the origin, primarily because the capacity of hyperbolic space increases exponentially, and regions further from the origin bend more, thereby providing more embedding space. 

In the field of hyperbolic learning, a common approach is to model data in the tangent space at the origin of the hyperbolic space. Thus, we introduce an extension to the \method model, referred to as the tangent-version \method\footnote{Experimental studies demonstrate that both the tangent version and hyperbolic version exhibit comparable outcomes.}. Suppose we have the tangential embedding $\mathbf{z}^{\mathcal{T}}$, then we have the Equation~\eqref{eq:tangent_alignment}\eqref{eq:tangent_HDO_computation}\eqref{eq:tangnet_stretching}.
\begin{align}
    &\bar{\mathbf{z}}^{\mathcal{T}} =  \mathbf{z}^{\mathcal{T}}-\mathbf{z}_c^{\mathcal{T}} &\quad ({\small \text{root alignment}}) \label{eq:tangent_alignment}\\ 
    &z_{\text{hdo}}^{\mathcal{T}}  := \frac{1}{|V|}\sum_{i\in V}w_id(\bar{\mathbf{z}}_i^{\mathcal{T}},\mathbf{o}^{\mathcal{T}}) &\quad ({\small\text{HDO computation}}) \label{eq:tangent_HDO_computation}\\
    &L_\text{hyp} = \sigma(-z^{\mathcal{T}}_{\text{hdo}}) &\quad ({\small\text{stretching}}\label{eq:tangnet_stretching}) 
\end{align}
where ${\mathbf{z}}_c^\mathcal{T} := \frac{1}{|V|}\sum_{i\in V}{\mathbf{z}}_i^\mathcal{T}$ denotes the center of the tangential embedding, and $w_i$ derived from $d(\mathbf{\bar{z}}_i,\mathbf{o})$ and we use identity function for simplicity, which signifies the Euclidean distance. The symbol $\sigma$ represents a monotonically increasing function such as linear function, tanh, sigmoid, etc. 
Similarly, the hyperbolic tangent embedding center is the optimal solution for minimizing the weighted sum of squared distance with all nodes in the tangent spaces, as described in Theorem~\ref{theorem: t-hmp-minimzation}.
\\

\begin{theorem}
Given the node embedding $\mathbf{z}\in\mathcal{H}^{d}$ of a graph $G=\{V,E\}$, the hyperbolic embedding center $\mathbf{z}_c\in\mathcal{H}^d$ is a solution of the minimization problem of the sum of weighted squared hyperbolic distance with all nodes in graph $G$, which is expressed as follows:

\begin{equation}\mathbf{z}_c = \min_{\mathbf{z}_a\in\mathcal{H}^{d,\kappa}}\sum_{i\in {V}}v_i d_\mathcal{H}^2(\mathbf{z}_i,\mathbf{z}_a),
\end{equation}
\label{theorem: hmp-minimzation}%
where $\mathbf{z}_a$ is any point in the manifold, and the embedding center includes M\"obius gyromidpoint in Poincar\'e ball model, weighted centroid in the Lorentz model.
\end{theorem}

\begin{theorem}
Given the node embedding $\mathbf{z}\in\mathcal{H}^{d}$ of a graph $G=\{V,E\}$, the hyperbolic tangent embedding center $\mathbf{z}_c^\mathcal{T}\in\mathcal{T}_\mathbf{o}\mathcal{H}^{d}$ is a solution of the minimization problem of the sum of weighted squared distance with all nodes in graph $G$ at the tangent space $\mathcal{T}_\mathbf{o}\mathcal{H}^{d}$, which is expressed as follows:
\label{theorem: t-hmp-minimzation}
\begin{equation}\mathbf{z}_c^\mathcal{T}=\min_{\mathbf{z}_a^\mathcal{T}\in\mathcal{T}_\mathbf{o}\mathcal{H}^{d,\kappa}}\sum_{i\in {V}}v_id^2(\mathbf{z}_i^\mathcal{T},\mathbf{z}_a^\mathcal{T}),
\end{equation}
where $\mathbf{z}_a^{\mathcal{T}}$ is any point in the tangent space of the manifold.
\end{theorem}

Note that the proposed method can be implemented in two ways: one is a hard operation, which involves replacing the all original embeddings, and the other is generating the desired update gradients by implementing it partially in the $L_{\text{hyp}}$ loss function as illustrated in Algorithm~\ref{alg:hie}. 
After conducting extensive experiments, we have found that both approaches yield comparable results. In the following section, for a better visualization comparison, we mainly report the results obtained using the partially root-alignment setting.

\begin{table}[!tp]
\vspace{-5pt}
\small
\caption{Comparisons of the different neural networks on node classification tasks. $\delta_{\text{({\sc Disease})}}$ is 1.0 and $\delta_{\text{({\sc Citeseer})}}$ is 2.5.}
\vspace{2pt}
\label{tab:hnn_node_classification}
\centering
\resizebox{0.48\textwidth}{!}{%
\begin{tabular}{@{}lcccccc@{}}
\toprule
Dataset                      & Dim   & MLP           & HNN           & HNN++~           & Ours          & $\Delta(\%)$  \\ \midrule\midrule
\multirow{3}{*}{{\sc Disease}}  & $8$   & $23.5\pm23.7$ & $46.6\pm12.6$ & $59.1\pm9.5$  & $\textbf{65.6}\pm6.6$   & $+{11.0}$ \\
                                & $64$  & $63.3\pm7.8$  & $68.4\pm4.4$  & $67.4\pm2.0$  & $\textbf{78.4}\pm0.7$  & $+{14.6}$ \\
                                & 256   & $70.9\pm3.5$  & $75.1\pm2.8$  & $76.1\pm5.7$  & $\textbf{77.3}\pm0.9$  & $+{1.60}$  \\ \midrule
\multirow{3}{*}{{\sc Citeseer}} & $8$   & $53.6\pm1.8$  & $52.5\pm1.6$  & $48.0\pm3.9$  & $\textbf{63.7}\pm2.1$  & $+{18.8}$ \\
                                & $64$  & $59.0\pm0.7$  & $57.2\pm0.7$  & $59.1\pm1.1$  & $\textbf{67.0}\pm1.0$  & $+{13.4}$ \\
                                & 256   & $59.1\pm0.7$  & $58.2\pm0.9$  & $60.2\pm1.0$  & $\textbf{68.2}\pm0.5$  & $+{13.3}$ \\ \bottomrule
\end{tabular}%
}
\vspace{-10pt}
\end{table}
\section{Experiments}


\subsection{Experimental Settings}
\textit{Datasets.} In terms of {experimental datasets}, we perform evaluations on four public available datasets, namely {\sc Disease}, {\sc Airport}, {\sc Cora}, {\sc Citeseer} and. 
For more details about these datasets, please refer to Appendix~\ref{sec:statistics}. \textit{Experimental Setups.} To ensure fairness, we employed the same data split for all comparison models in each experiment. The averaged results and standard deviation presented in these tables were computed from 12 runs, with the maximum and minimum values removed. The experiments were executed with the PyTorch~\cite{paszke2017automatic} on the NVIDIA GPUs using Adam~\cite{kingma2014adam} or Riemannian Adam~\cite{becigneul2018riemannian} optimizer. Due to the page limitation, we only list key findings here. Additional details regarding the experimental settings and results can be found in Appendix~\ref{sec:training_details},~\ref{appendix:Opposite_stretching} and~\ref{appendix:more_experimental_results}.

\begin{table*}[ht]
\small
\vspace{-5pt}
\centering
\caption{Comparisons on graph neural networks in Euclidean and hyperbolic space}
\vspace{2pt}
\resizebox{0.95\textwidth}{!}{%
\begin{tabular}{@{}l|ccc|ccc|ccc|ccc@{}}
\toprule
Dataset & \multicolumn{3}{c|}{{\sc Disease }{($\delta=0.0$)}}               & \multicolumn{3}{c|}{{\sc Airport }{($\delta=1.0$)}}               & \multicolumn{3}{c|}{{\sc Citeseer }{{($\delta=2.5$)}}}             & \multicolumn{3}{c}{{\sc Cora }{{($\delta=11.0$)}}}                   \\ 
Dim     & $8$            & $64$           & $256$          & $8$            & $64$           & $256$          & $8$            & $64$           & $256$          & $8$            & $64$           & $256$          \\ \midrule\midrule
SGC     & $70.6\pm6.3$  & $77.4\pm0.5$ & $78.7\pm0.4$ & $71.9\pm1.5$ & $83.3\pm1.4$ & $81.9\pm0.7$ & $70.1\pm0.6$ & $72.2\pm0.4$ & $72.2\pm0.4$ & $80.2\pm0.8$ & $82.2\pm0.5$ & $81.9\pm0.5$ \\
SAGE    & $66.5\pm8.2$ & $71.3\pm5.3$ & $72.0\pm4.1$ & $75.4\pm1.1$ & $87.1\pm1.6$ & $85.2\pm1.7$ & $71.0\pm0.8$ & $70.0\pm0.8$ & $70.9\pm0.8$ & $81.0\pm0.8$ & $80.5\pm0.6$ & $80.4\pm0.6$ \\
GCN     & $76.5\pm 2.9$  & $77.7\pm1.3$ & $78.8\pm1.5$ & $71.2\pm1.8$ & $85.2\pm0.5$ & $84.1\pm0.6$ & $69.6\pm0.3$ & $70.9\pm0.5$ & $71.0\pm0.5$ & $81.4\pm0.6$ & $81.9\pm0.3$ & $82.0\pm0.2$ \\
GAT    & $76.8\pm 1.2$  & $78.7\pm1.5$ & $80.2\pm2.0$ & $73.7\pm1.5$ & $87.5\pm0.4$ & $90.4\pm1.2$ & $69.8\pm0.3$ & $72.0\pm0.4$ & $71.7\pm0.5$ & $81.5\pm0.4$ & $82.9\pm0.3$ & $\textbf{83.0}\pm0.3$ \\
LGCN    & $86.6\pm0.7$     & $87.1\pm0.8$     &  $88.2\pm0.5$            & $88.2\pm1.7$     & $91.8\pm 1.3$              & $92.4\pm 0.3$             & $66.8\pm0.7$     & $69.3\pm0.8$     &  $70.5\pm0.5$            & $78.0\pm2.0$     & $81.2\pm1.0$     & $81.3\pm0.7$     \\
HGCN    & $86.0\pm3.2$     & $90.6\pm1.2$     & $92.2\pm1.8$     & $89.8\pm1.2$     & $94.0\pm0.6$     & $94.1\pm0.5$     & $65.6\pm1.5$     & $67.6\pm0.4$     & $67.6\pm0.7$     & $78.2\pm0.6$     & $78.5\pm0.6$     & $79.1\pm0.4$     \\
 \midrule
Ours    & $\textbf{94.4}\pm0.6$     & $\textbf{95.0}\pm0.8$     & $\textbf{95.4}\pm0.8$     & $\textbf{89.8}\pm1.1$     & $\textbf{94.1}\pm0.8$     & $\textbf{94.7}\pm0.4$     & $\textbf{72.5}\pm1.1$     & $\textbf{74.1}\pm0.3$     & $\textbf{74.2}\pm0.3$     & $\textbf{81.8}\pm0.6$     & $\textbf{83.0}\pm0.3$     & $\textbf{83.0}\pm0.2$     \\ 
\bottomrule
\end{tabular}%
}
\label{tab:hgcn_results}
\end{table*}

\begin{figure*}[!tp]
\centering
\subfigure{
\begin{minipage}[t]{0.245\textwidth}
\centering
\includegraphics[width=1.0\textwidth]{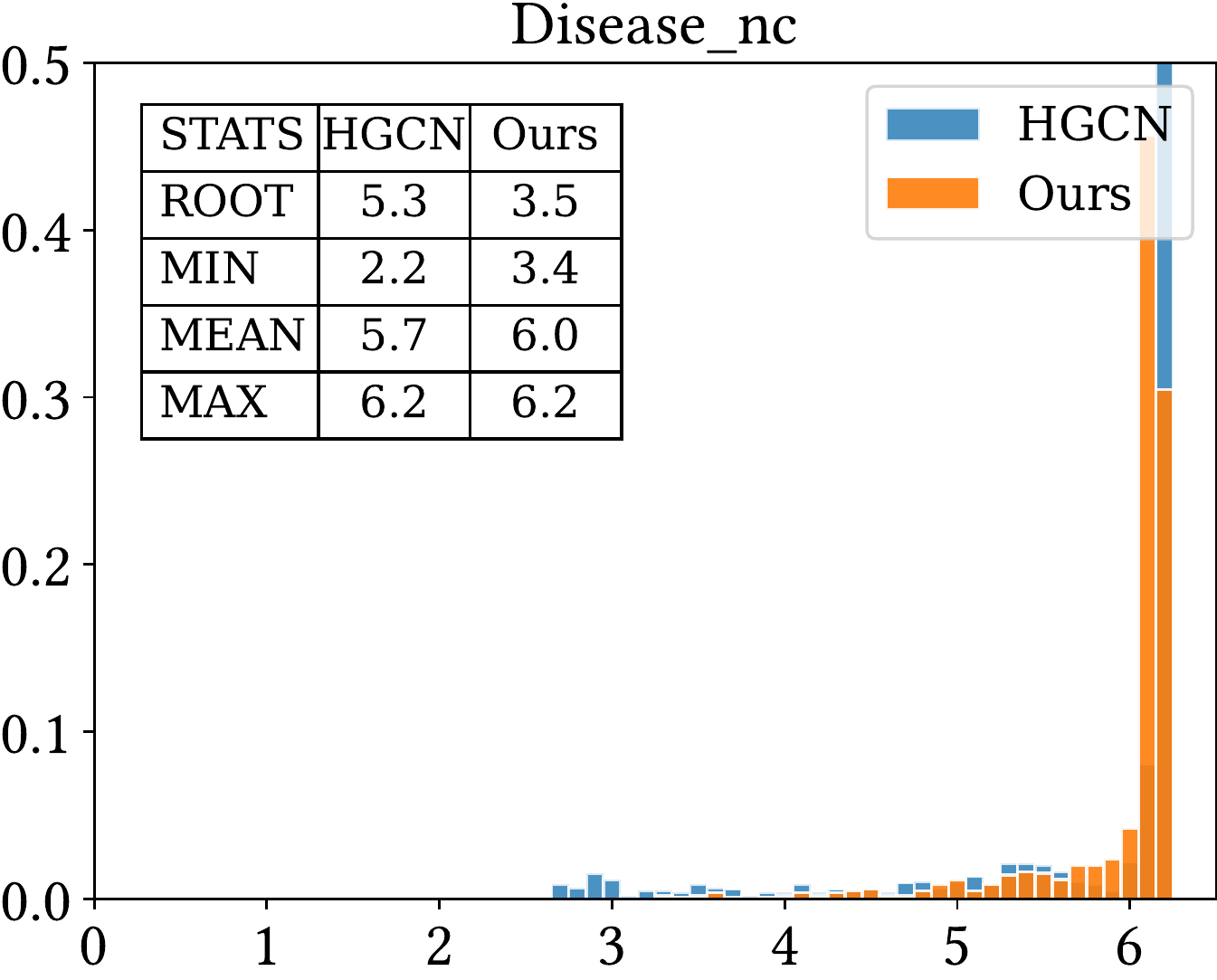}
\end{minipage}%
}%
\subfigure{
\begin{minipage}[t]{0.245\textwidth}
\centering
\includegraphics[width=1.0\textwidth]{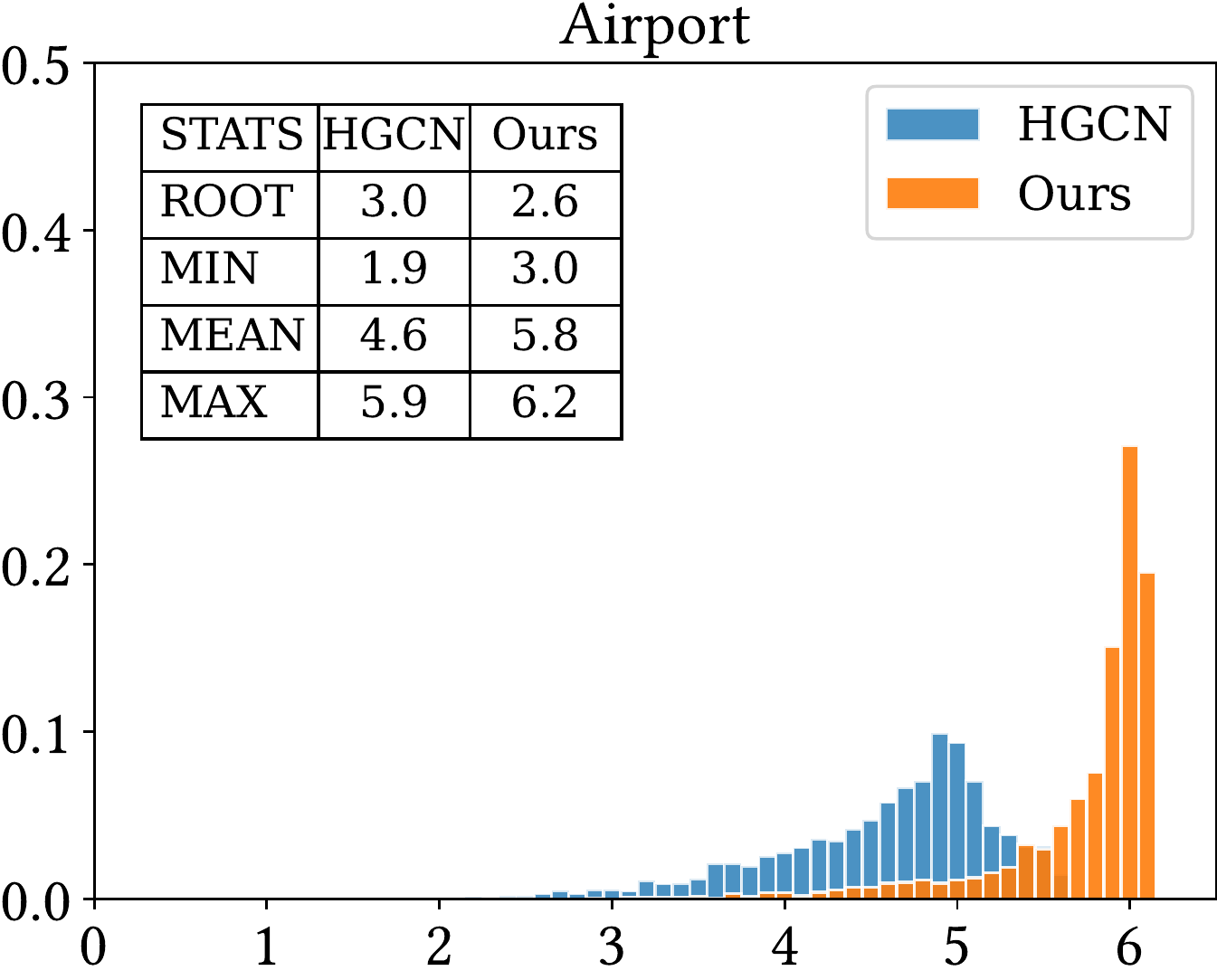}
\end{minipage}%
}%
\subfigure{
\begin{minipage}[t]{0.245\textwidth}
\centering
\includegraphics[width=1.0\textwidth]{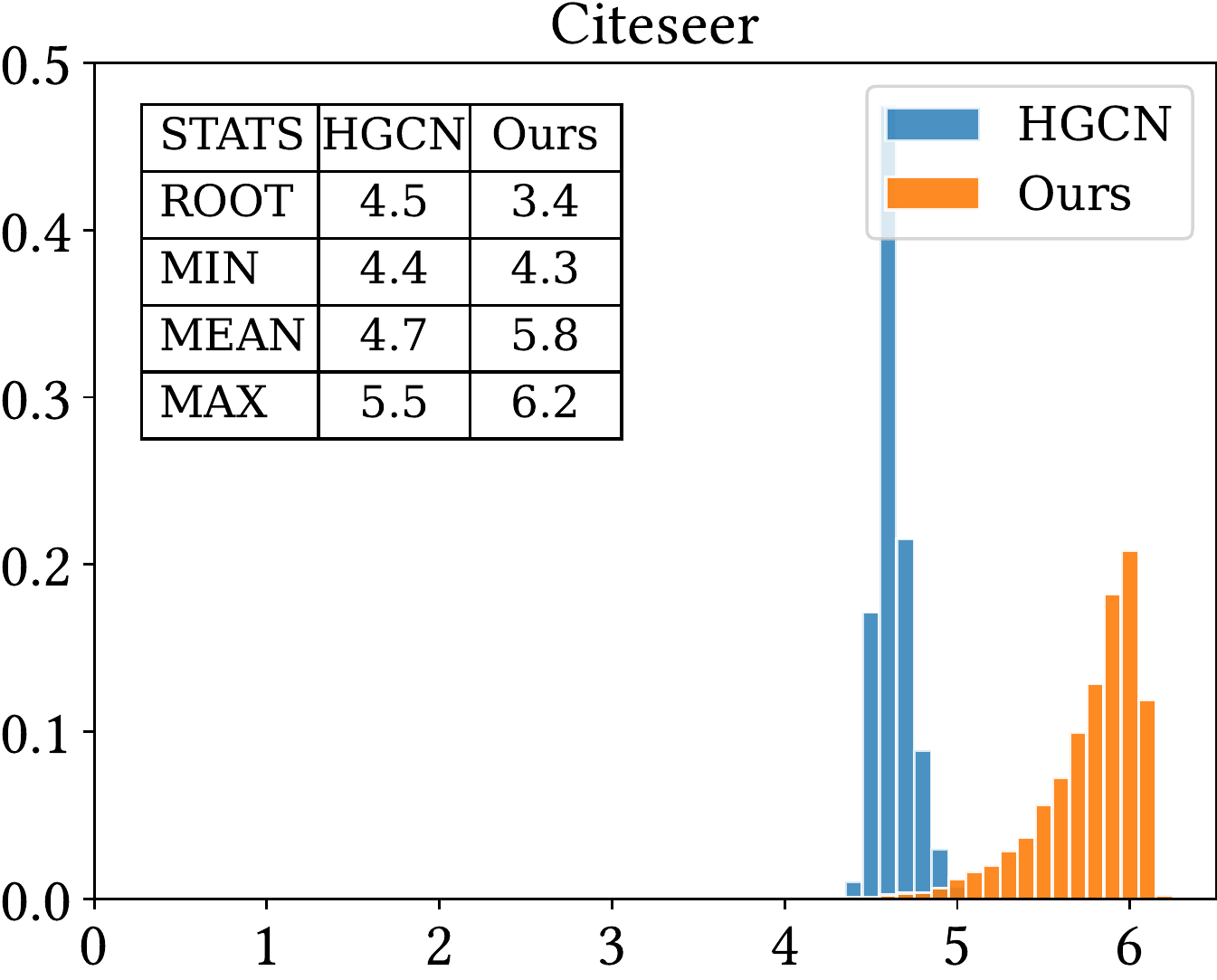}
\end{minipage}%
}%
\subfigure{
\begin{minipage}[t]{0.245\textwidth}
\centering
\includegraphics[width=1.0\textwidth]{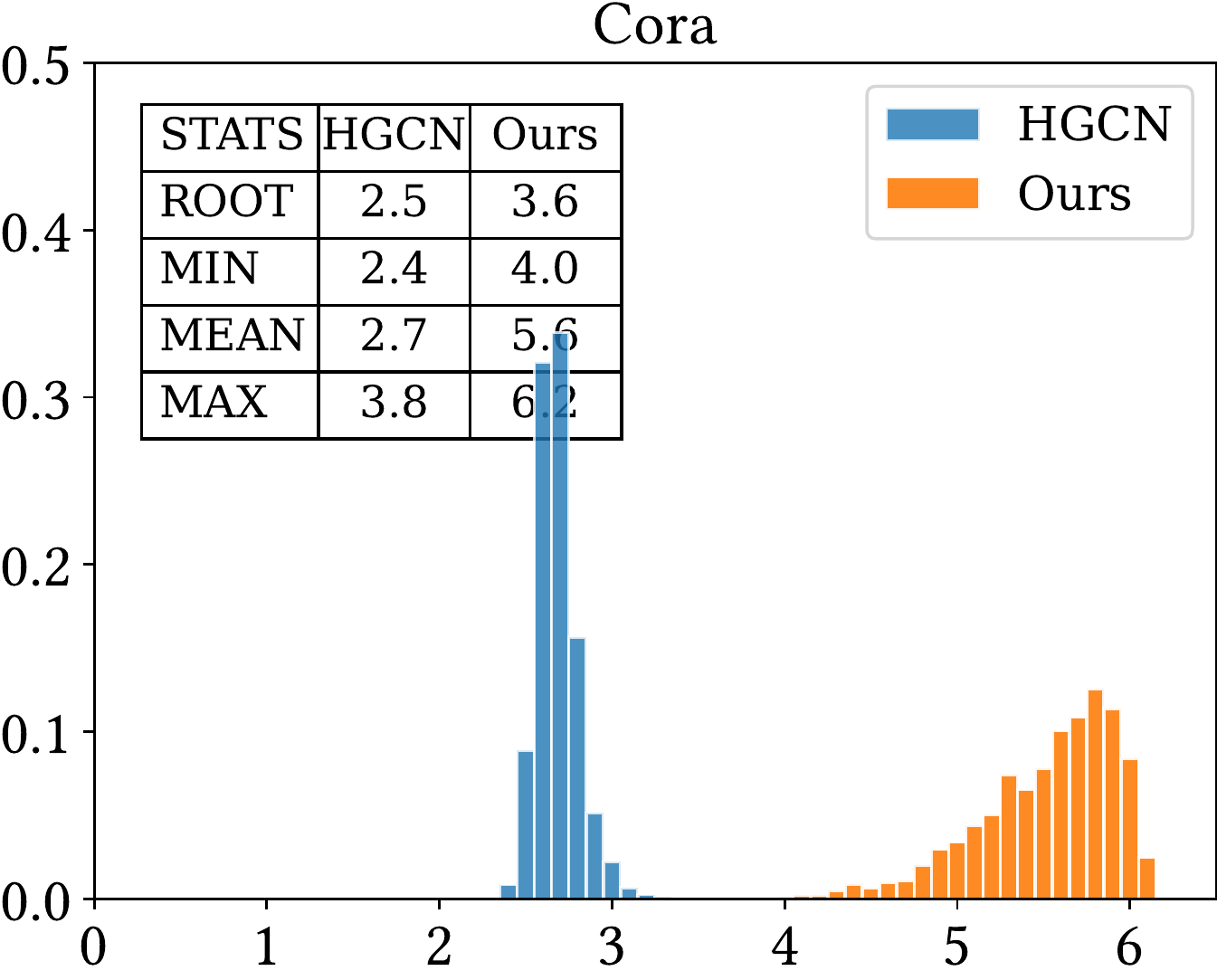}
\end{minipage}%
}%

\caption{Illustration \hdo distribution on  {\sc Disease}, {\sc Airport}, {\sc Citeseer} and {\sc Cora} where x-axis denotes the value of \hdo and y-axis is the corresponding ratio. For a complete comparison, please refer to Appendix~\ref{appendix:more_experimental_results}. Here the Root denotes HC.}
\vspace{-15pt}
\label{hdo_dist}
\end{figure*}
\subsection{Experiments on Shallow Models}
The shallow models optimize node embeddings as learnable parameters.
In line with prior research, we mainly evaluate their generalization capability on tree-structured graphs through link prediction tasks.
To comprehensively ascertain the effectiveness of the proposed method, we conduct a thorough experimental evaluation across a diverse range of training settings. Specifically, the training edge ratio is varied at 75\% and 25\%. Intuitively, as the amount of available structure information is reduced, abstracting the intrinsic hierarchical structure becomes increasingly challenging. 

We conduct a comparative evaluation between the Euclidean shallow model and the hyperbolic shallow model (based on the Poincar\'e ball). The experimental results on AUC and AP metrics are presented in \tableautorefname~\ref{tab:shallow_model_comparsion}. The findings from the results are as follows: The hyperbolic shallow models (including Poincar\'e and ours) demonstrate a distinct advantage over their Euclidean counterparts in most cases. However, as the amount of supervision information decreases, i.e., by reducing the training link ratio to 25\%, the performance of the conventional hyperbolic model deteriorates dramatically and even falls below that of the Euclidean model. This is understandable, as the model faces a more arduous task in perceiving the overall structure of the tree-like data when the number of known links is limited. 
On the contrary, our proposed method, with implicit hierarchical learning, effectively addresses this challenge and produces a substantial improvement, with the maximum enhancement reaching 21.4\%.

\subsection{Experiments on HNNs} 
HNNs utilize the feature of nodes instead of the topology information for graph learning, so here we carry out the task of node classification. 
For model comparisons, we include  Euclidean MLP, HNN~\cite{HNN}, and HNN++~\cite{HNN++}. Compared with HNN, HNN++ reduces the number of parameters, so it is more lightweight. Our work is built upon HNN other than HNN++ with the following consideration: without any additions or reductions to the original model, the effectiveness of the proposed method could be more clear. But note that the proposed method is also applicable to HNN++.
In this section, we assess the performance of our proposed model on two datasets: {\sc Disease} with low hyperbolicity and {\sc Citeseer} with high hyperbolicity. Hyperbolicity, a metric from graph theory, quantifies the tree-like structure of a network; lower values indicate a more tree-like structure. By analyzing the performance across datasets of varying hyperbolicity, we establish the robustness and generalizability of our model.

As shown in Table~\ref{tab:hnn_node_classification}, we notice that \hlms are more prominent in low-hyperbolicity tree-like data, i.e., {\sc Disease}, but is poorer in general graph data, like {\sc Citeseer} and even perform lower than the Euclidean model. 
Although the modified HNN++ improves the performance of the HNN model to a certain extent, its gains are modest. In contrast, our method has a significant improvement over the original HNN model, regardless of the low or high hyperbolicity graphs. In addition to being able to extract the underlying hierarchical structure of the data, the hyperbolic space has a larger room for embedding than the Euclidean space. The improvement of the proposed method on a high-hyperbolicity graph is mainly from the stretching operation where the nodes are fully expanded in the hyperbolic space and have nice separability.

\subsection{Experiments on HGNNs}
\label{others}
For {model comparisons}, we compare our proposed method against (1) four Euclidean GCN models, i.e., GCN~\cite{gcn2017}, Graph Attention Networks (GAT)~\cite{gat2018}, SGC~\cite{wu2019simplifying} and SAGE~\cite{graphsage}; and (2) two prominent hyperbolic GCN model, i.e., HGCN~\cite{hgcn2019} and LGCN~\cite{lgcn}, we take two different aggregation methods as described in Section~\ref{sec:hgnn} and report the best results of them. 
Our method is built upon the original model, HGCN. To ensure fairness, we adhered to the parameters and experimental strategies outlined in the original papers of all baselines and conducted extensive experiments in a consistent environment. 
Following the literature, we test on node classification task and report the F1-score for {\sc Disease} and {\sc Airport} datasets and accuracy for the others.  

The experimental results are reported in Table~\ref{tab:hgcn_results}. As observed, the proposed method \textit{consistently} outperforms strong hyperbolic baselines HGCN and LGCN, which demonstrates the effectiveness of the proposed method. We further dig into the details and have the following observations: 
(1) The performance of baselines varies along embedding dimensions. With the proposed \method, the hyperbolic model outperforms the baselines, especially achieving remarkable gains on {\sc Citeseer}, and {\sc Disease}. (2) As noted, the improvement on {\sc Airport} is relatively small. A possible explanation for this is that {\sc Airport} is well hierarchically organized. The original hyperbolic models have already successfully learned the proper embeddings, and the improvement is limited. Nonetheless, the proposal pushes the accuracy to a new level and obtains state-of-the-art performance.

\begin{table}[]
\vspace{-5pt}
    \centering
    \small
    \caption{Ablation Study of \method
    }
    \vspace{2pt}
    \resizebox{0.42\textwidth}{!}{%
    \begin{tabular}{@{}lcccc@{}}
    \toprule
    Dataset  & HGCN     & w/ Stretching & w/ Alignment & Ours  \\ \midrule\midrule
    {\sc Disease }  & $90.6\pm1.2$ & $89.4\pm2.5$   & $94.5\pm0.6$   & $\textbf{95.0}\pm0.8$ \\
    {\sc Airport}  & $94.0\pm0.6$ & $93.6\pm0.9$   & $93.6\pm0.3$   & $\textbf{94.1}\pm0.8$ \\
    {\sc Citeseer} & $67.6\pm0.4$ & $68.5\pm0.6$  & $66.7\pm0.9$   & $\textbf{74.1}\pm0.3$ \\
    {\sc Cora}     & $78.5\pm0.6$ & $81.5\pm0.5$   & $77.5\pm1.0$   & $\textbf{83.0}\pm0.3$ \\
    \bottomrule
    \end{tabular}%
    }
    \vspace{-20pt}
    \label{tab:decoupled_results}
\end{table}

\subsection{Analysis and Discussion}
\textbf{Visualization of HDO Distribution.} 
We conduct a thorough analysis of the changes in HDO between the original HGCN and our proposed method. To provide a clear comparison with the baselines, we implement root centering in the loss function without altering the original node embeddings directly. The results presented in Figure~\ref{hdo_dist} reveal that our method results in an increase in the mean HDO value, indicating that the node spreads more in the hyperbolic space. Additionally, it is observed that the shape of the HDO distribution has undergone a transformation, with a notable increase in the proportion of tail nodes, resulting in a more long-tailed distribution that better captures the tree-like or scale-free structure of the data. Furthermore, the root node (i.e., HC) is optimized to close the highest position (i.e., min HDO), demonstrating higher preservation of hierarchy.

\label{sec: analysis and discussion}


\textbf{{Ablation Study of \method.}} \method contains two steps: root alignment and level-aware stretching. In the following, we perform the decoupling analysis by decomposing \method into the corresponding two components. The experimental results are summarized in Table~\ref{tab:decoupled_results}. Unsurprisingly, the performance decreases by different degrees if we remove each component. With only level-aware stretching, the performance of \method barely satisfies and even degrades on low hyperbolicity datasets such as {\sc Disease} and {\sc Airport}. It is consistent with the expectation that merely stretching without aligning the root to the origin will lead to an incorrect hierarchy and large distortions for a highly tree-like dataset.  Equipped with only alignment, on the other hand, the \method has witnessed apparent improvement on {\sc Disease} but some declines on {\sc Cora} or {\sc Citeseer}. It is because alignment will arrange embeddings to the area near the origin, which is much more cramped than that close to the boundary. 

\textbf{Effectiveness of HC.}
Previously, we proposed the utilization of HC as the root node. For graph data, there exist numerous indicators to define the significance of nodes, such as degree centrality (DC), betweenness centrality (BC), and closeness centrality (CC).
The definitions and computation equations of these metrics can be found in Appendix~\ref{appendix:graph_centrality}. 
We implement these centralities HGCN and evaluate their performance in the context of node classification.
As shown in Table~\ref{tab:centrality}, the experimental results reveal that alternative metrics do not exhibit the same effectiveness as HC. This is probably because these metrics only take into account topological information to determine the weight of the most prominent nodes while neglecting the role of node features. Moreover, the computational complexity of calculating these centralities is high, whereas HC is relatively computationally efficient.

\textbf{Discussion with Non-Hyperbolic Methods}.
When addressing the modeling of trees, it is important to note that the scope is not limited to hyperbolic models, such as the ones presented in the works by Sonthalia et al.~\cite{sonthalia2020tree}, Abraham et al.~\cite{abraham2007reconstructing}, Chepoi et al.~\cite{chepoi2008diameters}, and Saitou et al.~\cite{saitou1987neighbor}. 
One cannot ignore the fact that they primarily focus on the topological structure of the graph, while neglecting the features that the objects carry.
Despite this limitation, these methods can serve as an excellent start point for enhancing the learning of hyperbolic models. Furthermore, they can be seamlessly integrated into the proposed method, thus leveraging their strengths while addressing the need to consider the object features.
\begin{table}[]
\vspace{-5pt}
\centering
\small
\caption{Comparisons with different centralities}
\vspace{2pt}
\resizebox{0.42\textwidth}{!}{%
\begin{tabular}{@{}lcccc@{}}
\toprule
Dataset  & BC           & CC           & DC           & HC(Ours)     \\ \midrule \midrule
{\sc Disease}  & $85.0\pm2.9$ & $81.6\pm3.2$ & $85.0\pm2.9$ & $\textbf{95.0}\pm0.8$ \\
{\sc Airport}  & $93.4\pm0.6$ & $94.0\pm0.5$ & {\sc $93.3\pm0.5$} & $\textbf{94.1}\pm0.8$ \\
{\sc Citeseer} & $66.8\pm0.7$ & $66.5\pm0.4$ & $67.3\pm0.5$ & $\textbf{74.1}\pm0.3$ \\ 
{\sc Cora}     & $81.3\pm0.4$ & $81.3\pm0.4$ & $81.3\pm0.4$ & $\textbf{83.0}\pm0.3$ \\
\bottomrule
\end{tabular}%
}
\vspace{-20pt}
\label{tab:centrality}
\end{table}
\section{Conclusion}
In this work, we first investigated the hierarchical representation ability of the currently popular hyperbolic models, including hyperbolic shallow models, HNNs, and HGNNs. The results indicate that the current hyperbolic models fail to obtain desired hierarchical embeddings. 
With the proposed method, \method, the hyperbolic representation ability has been substantially improved. 
More importantly, the proposed method does not introduce additional model parameters or change the original architecture. 
Considering the fact that our proposed method is task- and model-agnostic to be readily applied in various scenarios, 
we believe that our research holds significant ramifications for the field of hyperbolic representation learning.
In future work, we will investigate the embedding in more expressive manifolds~\cite{gu2019learning,xiong2022ultrahyperbolic,xiong2023geometric,xiong2021pseudo,bachmann2020constant} and explore contrastive learning~\cite{zhang2022costa,zhang2022spectral,liu2022enhancing} to enhance the hierarchical learning.

\section*{Acknowledgements}
We express gratitude to the anonymous reviewers and area chairs for their valuable comments and suggestions.
This work was partly supported by grants from the National Key Research and Development Program of China (No.~2018AAA0100204) and the Research Grants Council of the Hong Kong Special Administrative Region, China (CUHK 14222922, RGC GRF, No. 2151185).

\newpage
\nocite{langley00}
\bibliography{references}
\bibliographystyle{icml2023}

\newpage
\appendix
\onecolumn
\section*{Appendix}
\section{Riemannian Geometry}
\label{appendix:Riemannian_geometry}
In this section, we present the details about the concepts mentioned in the work, i.e., geodesics, distance function, maps and parallel transport of a Riemannian geometry $(\mathcal{M}, g)$ and summarize the formula of hyperbolic space used in the study.

\textbf{Geodesics and Distance Function}. For a curve $\gamma:[\alpha,\beta]\to \mathcal{M}$, the length of $\gamma$, called geodesics, is defined as $L(\gamma)=\int_\alpha^\beta\|\gamma^\prime(t)\|_g dt$. Then the distance of $\mathbf{u}, \mathbf{v} \in \mathcal{M}$ is given by $d_\mathcal{M}(\mathbf{u},\mathbf{v})=\inf L(\gamma)$ where $\gamma$ is a curve that $\gamma(\alpha)=\mathbf{u}, \gamma(\beta) = \mathbf{v}$. 

\textbf{Maps and Parallel Transport}. With assuming the manifolds are smooth, i.e. the maps are diffeomorphic, the map defines the projection between the manifold and the tangent space. 
For a point $\mathbf{x}\in \mathcal{M}$ and a vector $\mathbf{v}\in\mathcal{T}_\mathbf{x}\mathcal{M}$, there exists a unique geodesic $\gamma:[0,1]\to\mathcal{M}$ where $\gamma(0)=\mathbf{x}, \gamma^\prime(0)=\mathbf{v}$. The exponential map $\exp_\mathbf{x}: \mathcal{T}_\mathbf{x}\mathcal{M} \to \mathcal{M}$ is defined as $\exp_\mathbf{x}(\mathbf{v})=\gamma(1)$ and logarithmic map $\log_\mathbf{x}: \mathcal{M} \to \mathcal{T}_\mathbf{x}\mathcal{M}$ is the inverse of $\exp_\mathbf{x}$. The parallel transport $PT_{\mathbf{x}\rightarrow \mathbf{y}}:\mathcal{T}_\mathbf{x}\mathcal{M}\to\mathcal{T}_\mathbf{y}\mathcal{M}$ achieves the transportation from point $\mathbf{x}$ to $\mathbf{y}$ that preserves the metric tensors along the unique geodesic.

\textbf{Hyperbolic Models.}
Riemannian manifolds with different curvatures define different geometries: elliptic geometry (positive curvature), Euclidean geometry (zero curvature), and hyperbolic geometry (negative curvature). Here, we focus on the negative curvature space, i.e., hyperbolic geometry.  Hyperbolic geometry is a  Riemannian manifold with a constant negative sectional curvature.
There exist multiple equivalent hyperbolic models which show different characteristics but are mathematically equivalent. We here mainly consider two widely studied hyperbolic models: the Poincaré ball model \cite{nickel2017poincare} and the Lorentz model (also known as the hyperboloid model) \cite{nickel2018learning}. 

Let $\| . \|$ be the Euclidean norm and  $\left\langle .,. \right \rangle_\mathcal{L}$ represent the Minkowski inner product. The formulas and operations associated with distance, maps, and parallel transport, hyperbolic origin point are summarized in Table~\ref{tab:operation}, where $\oplus_{\kappa}$ and $\operatorname{gyr}[.,.] v$ 
are M\"obius addition and gyration operator~\cite{ungar2007hyperbolic}, respectively, which are given in the following. 

\textit{{M\"obius addition}}: For $\mathbf{x}, \mathbf{y} \in \mathcal{B}_{\kappa}^{n}$, the M\"obius addition~\cite{ungar2007hyperbolic} is
\begin{equation}
\label{equ:mobiuous_addition}
    \mathbf{x} \oplus_{\kappa} \mathbf{y}=\frac{\left(1-2 \kappa\langle \mathbf{x}, \mathbf{y}\rangle_{2}-\kappa\|\mathbf{y}\|_{2}^{2}\right) \mathbf{x}+\left(1+\kappa\|\mathbf{x}\|_{2}^{2}\right) \mathbf{y}}{1-2 \kappa\langle \mathbf{x}, \mathbf{y}\rangle_{2}+\kappa^{2}\|\mathbf{x}\|_{2}^{2}\|\mathbf{y}\|_{2}^{2}}.
\end{equation}
Then the induced M\"obius substraction $\ominus_\kappa$ is given by $\mathbf{x}\ominus_\kappa\mathbf{y} = \mathbf{x}\oplus_\kappa(-\mathbf{y})$.

\textit{{Gyration operator}.} In the theory of gyrogroups, the notion of the gyration operator~\cite{ungar2008gyrovector} is given by
$$\operatorname{gyr}[\mathbf{x}, \mathbf{y}] \mathbf{v}=\ominus_{\kappa}\left(\mathbf{x} \oplus_{\kappa} \mathbf{y}\right) \oplus_{\kappa}\left(\mathbf{x} \oplus_{\kappa}\left(\mathbf{y} \oplus_{\kappa} \mathbf{v}\right)\right).$$

 \begin{table*}[htp!]
    \centering
    \caption{Summary of operations in the Poincar{\'e} ball model and the Lorentz model ($\kappa<0$)}
    \resizebox{1.0\textwidth}{!}{%
    \begin{tabular}{lcc}
    \toprule
    & \textbf{Poincar{\'e} Ball Model} & \textbf{Lorentz Model (Hyperboloid Model)} \\ \midrule \midrule
        \textbf{Manifold}  & $\mathcal{B}_{\kappa}^{n}=\left\{\mathbf{x} \in \mathbb{R}^{n}:\langle \mathbf{x}, \mathbf{x}\rangle_{2}<-\frac{1}{\kappa}\right\}$ 
    &  $\mathcal{L}_{\kappa}^{n}=\left\{\mathbf{x} \in \mathbb{R}^{n+1}:\langle \mathbf{x}, \mathbf{x}\rangle_{\mathcal{L}}=\frac{1}{\kappa}\right\}$          \\
    \textbf{Metric}     &  $g_{\mathbf{x}}^{\mathcal{B}_{\kappa}}=\left(\lambda_{\mathbf{x}}^{\kappa}\right)^{2} \mathbf{I}_n \text { where } \lambda_{\mathbf{x}}^{\kappa}=\frac{2}{1+\kappa\|\mathbf{x}\|_{2}^{2}}$ & $g_{\mathbf{x}}^{\mathcal{L}_{\kappa}}=\eta, \text { where } \eta \text { is } I \text { except } \eta_{0,0}=-1$    \\
\textbf{Distance}   &$d_{\mathcal{B}}^{\kappa}(\mathbf{x}, \mathbf{y})=\frac{1}{\sqrt{|\kappa|}} \cosh ^{-1}\left(1-\frac{2 \kappa\|\mathbf{x}-\mathbf{y}\|_{2}^{2}}{\left(1+\kappa\|\mathbf{x}\|_{2}^{2}\right)\left(1+\kappa\|\mathbf{y}\|_{2}^{2}\right)}\right)$
    &  $d_{\mathcal{L}}^{\kappa}(\mathbf{x}, \mathbf{y})=\frac{1}{\sqrt{|\kappa|}} \cosh ^{-1}\left(\kappa\langle \mathbf{x}, \mathbf{y}\rangle_{\mathcal{L}}\right)$  \\
    \textbf{Logarithmic Map}    &$\log _{\mathbf{x}}^{\kappa}(\mathbf{y})=\frac{2}{\sqrt{|\kappa| \lambda^{\kappa}}} \tanh ^{-1}\left(\sqrt{|\kappa|}\left\|-\mathbf{x} \oplus_{\kappa} \mathbf{y}\right\|_{2}\right) \frac{-\mathbf{x} \oplus_{\kappa} \mathbf{y}}{\left\|-\mathbf{x} \oplus_{\kappa} \mathbf{y}\right\|_{2}}$              
    &$\log _{\mathbf{x}}^{\kappa}(\mathbf{y})=\frac{\cosh ^{-1}\left(\kappa\langle \mathbf{x}, \mathbf{y}\rangle_{\mathcal{L}}\right)}{\sinh \left(\cosh ^{-1}\left(\kappa\langle \mathbf{x}, \mathbf{y}\rangle_{\mathcal{L}}\right)\right)}\left(\mathbf{y}-\kappa\langle \mathbf{x}, \mathbf{y}\rangle_{\mathcal{L}} \mathbf{x}\right)$\\ 
    \textbf{Exponential Map}   & $\exp _{\mathbf{x}}^{\kappa}(\mathbf{v})=\mathbf{x} \oplus_{\kappa}\left(\tanh \left(\sqrt{|\kappa|} \frac{\lambda_{\mathbf{x}}^{\kappa}\|\mathbf{v}\|_{2}}{2}\right) \frac{\mathbf{v}}{\sqrt{|\kappa|\|\mathbf{v}\|_{2}}}\right)$
    & $\exp _{\mathbf{x}}^{\kappa}(\mathbf{v})=\cosh \left(\sqrt{|\kappa|}\|\mathbf{v}\|_{\mathcal{L}}\right) \mathbf{x}+\mathbf{v} \frac{\sinh \left(\sqrt{|\kappa|}\|\mathbf{v}\|_{\mathcal{L}}\right)}{\sqrt{|\kappa||| \mathbf{v}||_{\mathcal{L}}}}$ \\
    \textbf{Parallel Transport} &  $P T_{\mathbf{x} \rightarrow \mathbf{y}}^{\kappa}(\mathbf{v})=\frac{\lambda_{\mathbf{x}}^{\kappa}}{\lambda_{\mathbf{y}}^{\kappa}} \operatorname{gyr}[\mathbf{y},-\mathbf{x}] v $& $ P T_{\mathbf{x} \rightarrow \mathbf{y}}^{\kappa}(\mathbf{v})=\mathbf{v}-\frac{\kappa\langle \mathbf{y}, \mathbf{v}\rangle_{\mathcal{L}}}{1+\kappa\langle \mathbf{x}, \mathbf{y}\rangle_{\mathcal{L}}}(\mathbf{x}+\mathbf{y})$ \\
    \textbf{Origin Point} & $\mathbf{0}_n$ & $[\frac{1}{\sqrt{|\kappa|}},\mathbf{0}_n]$\\
    \bottomrule
    \end{tabular}
    }
    \label{tab:operation}
    \end{table*}   

\section{Dataset Construction about TREE-L and TREE-H}
\label{appendix:data_construction_treelh}
The geometric information of a real-world dataset is not explicitly given and the dataset is also not a pure-tree structure in general, so we synthesize two tree datasets, TREE-L and TREE-H, to track the position of node embedding in hyperbolic space. In particular, the two trees are with the same structures but with different homophily. All nodes except leaf nodes have three child nodes. In total, there are eight layers, 1092 edges, and 1093 nodes for each tree. On the tree with high homophily (i.e., TREE-H), nodes in each largest subtree (total, three largest subtrees) have the same node labels. We set the root node as another class, and then it is easy to know that there are four different classes. For the node features belonging to the same class, we generate a $32$-dimension vector from a Gaussian distribution. The mean and variance of the four classes are (0.0, 1.0), (1.0, 1.0), (2.0, 1.0), (3.0, 1.0), respectively, where (0.0, 1.0) is for the class of root and the other three are for the rest. For TREE-L, on the other hand, nodes at the same level are with the same labels. We further set the first four layers to be the same class to ensure the total classes are four. Then for the four classes in TREE-L, we also use the same method, mean, and variance to generate the node features. 

\begin{figure}[htbp]
\centering
\includegraphics[width=0.65\textwidth]{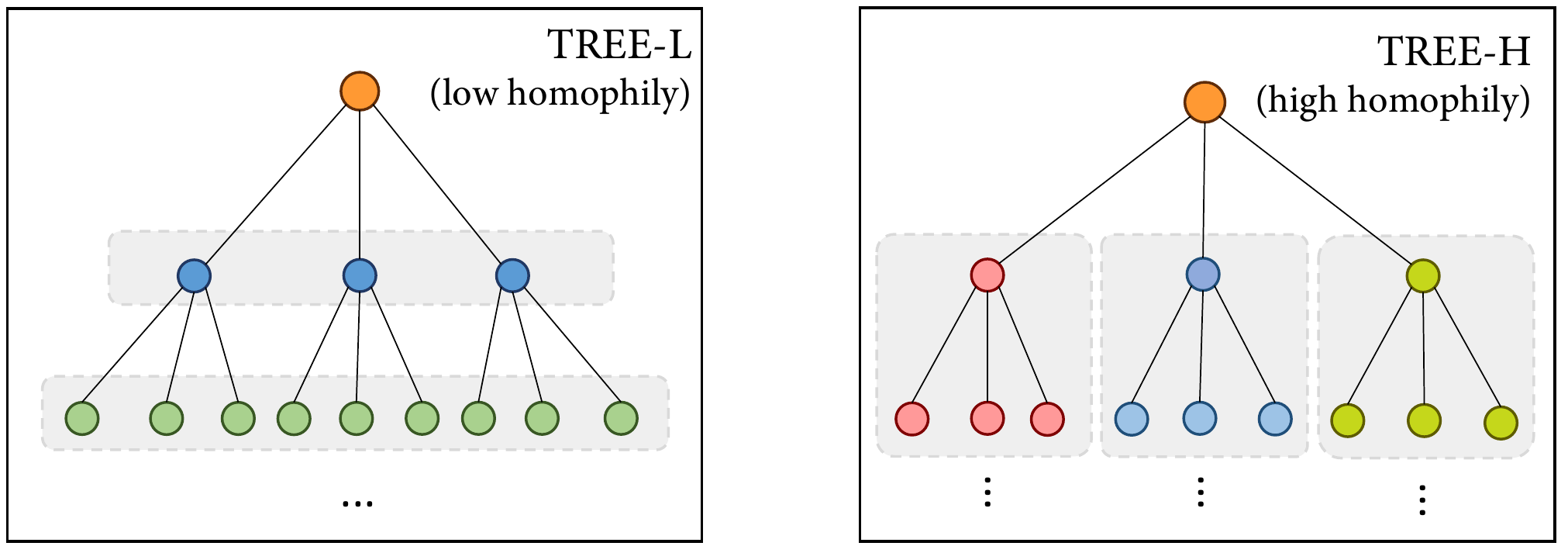}
\centering
\caption{Illustration of the structure of synthetic TREE-L/H. In TREE-L, the node class in the same level is the same, while in TREE-H, the node class in each largest subtree is the same. For clarity, we use the same color to denote the same class in these two trees.}
\label{fig:motivation}
\end{figure}

There is a metric proposed by~Pei et al.~\cite{gemo-gcn} to measure the homophily rate $H_m$ of a graph $G$, which is given by:
\begin{align}
    H_m(G)=\frac{1}{|V|}\sum_{v \in V}\frac{\text{\#Label($v$'s neighbors)==Label of $v$}}{\text{\#$v$'s neighbors}}.
\label{equ: homophily}
\end{align}
Then, the computed homophily rates $H_m$ of TREE-H and TREE-L are $\textbf{0.998}$ and $\textbf{0.018}$, respectively.

\section{Hyperbolic Midpoint Computation and Alignment} 
\label{appendix:hyperbolic midpoint}
Let $\mathbf{Z}$ be the final layer embedding matrix consisting of $n$ hyperbolic node vectors $(\mathbf{z}_1, \cdots, \mathbf{z}_i, \cdots, \mathbf{z}_n)$ where $\mathbf{z}_i\in\mathcal{H}^d_\kappa$, and let $v_i\in \mathbb{R}$ ($v_i\geq 0$ and $\sum_i v_i>0$) be the weight for $\mathbf{z}_i$ and we set it as 1 in this study, the details on the computation of hyperbolic center $\mathbf{z}_c$ and alignment given as follows.

\textit{In the Poincar\'e ball model}, the center, i.e., M\"obius gyromidpoint~\cite{ungar2008gyrovector},  is computed in gyrovector space, which is given by
\begin{equation}
\mathbf{z}_c:=\frac{1}{2} \oplus_{\kappa}\left(\frac{\sum_{i=1}^{n} v_{i} \lambda_{\mathbf{z}_{i}}^\kappa \mathbf{z}_{i}}{\sum_{i=1}^{n} v_{i}\left(\lambda_{\mathbf{z}_{i}}^{\kappa}-1\right)}\right),
\end{equation}
where $\lambda_{\mathbf{z}_i}^\kappa=\frac{2}{1+\kappa\left\|\mathbf{z}_{i}\right\|_2^{2}} .$ The root alignment operation is related to addition in Poincar\'e ball model which is computed by Equation~(\ref{equ:mobiuous_addition}).

\textit{In the Lorentz model}~\cite{law2019lorentzian}, the center (also called Lorentzian centroid) is computed by 
\begin{equation}
    \mathbf{z}_c:=\frac{1}{\sqrt{|\kappa|}} \frac{\sum_{i=1}^{n} v_{i} \mathbf{z}_{i}}{|\left\|\sum_{i=1}^{n} v_{i} \mathbf{z}_{i}\right\|_{\mathcal{L}} \mid},
\end{equation}
The root alignment operation is related to addition operation in Lorentz model which is formulated as:
\begin{equation}
   \mathbf{z}_i\oplus_\kappa({\mathbf{z}_c}) := \exp_{\bar{\mathbf{z}}}^\kappa (PT_{\mathbf{o}\to\bar{\mathbf{z}}}(\log_\mathbf{o}^\kappa(\mathbf{z}_i))). 
\end{equation}

\textit{In the tangent space}, the embedding $\mathbf{Z}$ is first projected to the tangent space at origin, that is $\mathbf{Z}^\mathcal{T} = \log^\kappa_\mathbf{o}(\mathbf{Z}^\mathcal{T})$. Then the center is defined by the following weighted manner,
\begin{equation}
\mathbf{z}_c^\mathcal{T}:= \frac{\sum_{i=1}^n v_i\mathbf{z}_i^\mathcal{T}}{\sum_{i=1}^n v_i},
\end{equation}
The root alignment is given as,
\begin{equation}
    \mathbf{z}_i^\mathcal{T}\oplus_\kappa({\mathbf{z}_c}) := \mathbf{z}_i^\mathcal{T} - \mathbf{z}_c^\mathcal{T}.
\end{equation}

\section{Proof of Theorem}
\subsection{Proof of Theorem~\ref{theorem: hmp-minimzation}}
The M\"obius gyromidpoint for Poincar\'e ball is a solution of the minimization problem, which has been proved by~\citet{HNN++} and please check Theorem 2 in work~\cite{HNN++}. The Lorentzian centroid for Lorentz model also minimizes the weighted sum of the squared distance, which has been proved by~\cite{law2019lorentzian}(c.f.,Theorem 3.3)

\subsection{Proof of Theorem~\ref{theorem: t-hmp-minimzation}}
\label{appendix:theorem}
\begin{proof}

For the sake of clarity, we shall disregard the superscript $\mathcal{T}$ in the following proof. Let $\mathbf{z}_c= \frac{\sum_{i=1}^n v_i\mathbf{z}_i}{\sum_{i=1}^n v_i}$ denote the weighted center, where $v_i$ represents the weight associated with $\mathbf{z}_i$, and we use $v_s$ denote $\sum_{i=1}^n v_i$ for breity. Then we easily derive that,
\begin{equation}
\begin{aligned}
\label{equ:sum_tangent_vector}
    \mathbf{z}_c &= \frac{\sum_{i=1}^n v_i\mathbf{z}_i}{\sum_{i=1}^n v_i} \\
    & \Rightarrow \mathbf{z}_c\cdot \sum_{i=1}^n v_i = \sum_{i=1}^n v_i\mathbf{z}_i
    \\
    & \Rightarrow \mathbf{z}_c\cdot v_s = \sum_{i=1}^n v_i\mathbf{z}_i.
\end{aligned}
\end{equation}
Furthermore, let $\mathbf{z}_a$ be a point in the tangent space. Then the minimization problem can be reformulated as:
\begin{equation}
    \begin{aligned}
    \min_{\mathbf{z}_a\in{\mathcal{T}_\mathbf{o}}\mathcal{H}_\kappa^d} \sum_{i=1}^n v_i d^2(\mathbf{z}_i,\mathbf{z}_a)
    &= \min_{\mathbf{z}_a\in{\mathcal{T}_\mathbf{o}}\mathcal{H}_\kappa^d} \sum_{i=1}^n v_i ||\mathbf{z}_i-\mathbf{z}_a||^2\\
    &= \min_{\mathbf{z}_a\in{\mathcal{T}_\mathbf{o}}\mathcal{H}_\kappa^d}\sum_{i=1}^nv_i\left(\|\mathbf{z}_i\|^2 + \|\mathbf{z}_a\|^2-2\mathbf{z}_i^T\mathbf{z}_a\right)\\
    &= \min_{\mathbf{z}_a\in{\mathcal{T}_\mathbf{o}}\mathcal{H}_\kappa^d}\sum_{i=1}^nv_i\|\mathbf{z}_i\|^2 + \sum_{i=1}^nv_i\|\mathbf{z}_a\|^2-\sum_{i=1}^nv_i\cdot2\mathbf{z}_i^T\mathbf{z}_a\\
    &=\min_{\mathbf{z}_a\in{\mathcal{T}_\mathbf{o}}\mathcal{H}_\kappa^d} \sum_{i=1}^nv_i\|\mathbf{z}_i\|^2 + v_s\|\mathbf{z}_a\|^2-2\sum_{i=1}^nv_i\cdot\mathbf{z}_i^T\mathbf{z}_a \\
    &=\min_{\mathbf{z}_a\in{\mathcal{T}_\mathbf{o}}\mathcal{H}_\kappa^d} \sum_{i=1}^nv_i\|\mathbf{z}_i\|^2 + v_s\|\mathbf{z}_a\|^2 -2 v_s\cdot\mathbf{z}_c^T\mathbf{z}_a \text{\quad{\color{asparagus}\%(refer to Equation (16))}}\\ 
    &=\min_{\mathbf{z}_a\in{\mathcal{T}_\mathbf{o}}\mathcal{H}_\kappa^d} \sum_{i=1}^n v_i\|\mathbf{z}_i\|^2 + v_s\left(\|\mathbf{z}_a\|-2\mathbf{z}_c^T\mathbf{z}_a\right) \\
    &=\min_{\mathbf{z}_a\in{\mathcal{T}_\mathbf{o}}\mathcal{H}_\kappa^d} \sum_{i=1}^n v_i\|\mathbf{z}_i\|^2 + v_s\left(\|\mathbf{z}_c-\mathbf{z}_a\|^2-\|\mathbf{z}_c\|^2\right) \\
    &=\min_{\mathbf{z}_a\in{\mathcal{T}_\mathbf{o}}\mathcal{H}_\kappa^d} \sum_{i=1}^n v_i\|\mathbf{z}_i\|^2 + v_s\|\mathbf{z}_c-\mathbf{z}_a\|^2-v_s\|\mathbf{z}_c\|^2 \\
    &=\min_{\mathbf{z}_a\in{\mathcal{T}_\mathbf{o}}\mathcal{H}_\kappa^d} {\sum_{i=1}^n v_i\|\mathbf{z}_i\|^2 - v_s\|\mathbf{z}_c\|^2} + v_s\|\mathbf{z}_c-\mathbf{z}_a\|^2 \\
    &=\min_{\mathbf{z}_a\in{\mathcal{T}_\mathbf{o}}\mathcal{H}_\kappa^d} \underbrace{\sum_{i=1}^n v_i\left(\|\mathbf{z}_i\|^2 -\|\mathbf{z}_c\|^2\right)}_{constants} + v_s\|\mathbf{z}_c-\mathbf{z}_a\|^2.\\
    \end{aligned}
\end{equation}
It is known that the first term in the equation are constant. Furthermore, it can be deduced that the last term, $v_s\|\mathbf{z}_c-\mathbf{z}_a\|^2$, is non-negative. It follows that the total distance sum to all other nodes is minimized when $\mathbf{z}_a=\mathbf{z}_c$. In other words, the embedding center, represented by $\mathbf{z}_c$, is the solution that minimizes the sum of distances to all other nodes.

\end{proof}

\section{Graph Centrality}
\label{appendix:graph_centrality}
In graph theory and network analysis, centrality metrics are utilized to assign numerical or ordinal values to nodes within a graph commensurate with their relative importance and prominence within the network. These metrics find wide-ranging applications, such as identifying key actors in a social network, critical infrastructure nodes in internet or urban networks, primary vectors of disease transmission, and salient nodes in brain networks~\cite{van2013network,saberi2021topological}. In this part, we present some important centrality in graphs for reference. Unless otherwise specified, the graph referred to here is assumed to be an undirected one.

\textbf{Degree centrality}. The degree centrality is a commonly employed metric in graph theory and network analysis, which quantifies the centrality of a node by calculating the number of edges incident upon it, commonly referred to as its degree, that is,
\begin{equation}
C_D(v)=\operatorname{deg}(v).
\end{equation}
 Computing degree centrality of all nodes takes $\Theta(|V|^2)$ in a dense adjacency matrix representation and $\Theta(|E|)$ in a sparse matrix representation. 
Nodes with a higher degree are considered more central in the network, and thus we select the node with the highest degree centrality as the root node for comparison.

\textbf{Betweenness centrality.}
The betweenness centrality of a vertex is based on the number of shortest paths passing through it. This metric is computed by considering all pairs of vertices in a connected graph and identifying the number of shortest paths between them that pass through the given vertex, which is given by
\begin{equation}
   C_B(v)=\sum_{s \neq v \neq t \in V} \frac{\delta{s t}(v)}{\delta{s t}}, 
\end{equation}
where $\delta{s t}$ represents the total number of shortest paths from vertex $s$ to vertex $t$, and $\delta{s t}(v)$ denotes the number of those paths that traverse vertex $v$.  Computing betweenness centrality of all nodes involves the computation of shortest path between all pairs, which takes  $\Theta(|V|^3)$~\cite{cormen2022introduction} for general graphs, and $O(|V|^2\log|V|+|V||E|)$ for sparse graphs. Similarly, we take the node with highest betweenness centrality for experiential comparison.

\textbf{Closeness centrality.} The closeness centrality quantifies how closely connected the node is to all other nodes in the network. The core concept of closeness centrality is a node is central if the average number of links needed to
reach another node is small.
It is computed by taking the reciprocal of the sum of the shortest path distances between the node and every other node in the graph, which can be formulated as,
\begin{equation}
    C_C(v)=\frac{n-1}{\sum_{u=1}^{n-1} d(v, u)},
\end{equation}
where $d(v, u)$ is the shortest-path distance between $v$ and $u$, and $n-1$ is the number of nodes reachable from $v$. The time complexity is similar to betweenness centrality since it requires to compute the shortest distance as well.
Nodes with higher closeness centrality are considered more central as they are closer to all other nodes. Therefore, we treat the highest clustering centrality as the equivalent for comparisons.

\section{Experimental Details}
\label{appendix:experimental_details}
\subsection{Statistics of Datasets}
\label{sec:statistics}
We list the statistics of the dataset used in our work in Table~\ref{tab:statistics}.
In specific, {\sc Disease} is simulated by the SIR disease spreading model~\cite{anderson1992infectious}, where the label indicates whether the node was infected or not.
The {\sc Airport} is a flight network where nodes are the airports, edges represent airline routes, and the label denotes the population of the country that the corresponding airport belongs to. {\sc Cora} and {\sc Citeseer} are standard benchmarks describing citation networks with nodes as scientific papers, edges as citations between them, and node labels as academic areas.

\begin{table}[h]
\centering
\caption{Statistics of the datasets.}
\vspace{5pt}
\resizebox{0.45\textwidth}{!}{%
\begin{tabular}{@{}lllcc@{}}
\toprule
{Dataset}  & Nodes & Edges & Classes & Node features \\ \midrule
{{\sc disease (NC)}}  & 1,044  & 1,043  & 2       & 1,000          \\
{\sc Disease (LP)} & 2,665  & 2,664  & 2       & 1000             \\
{{\sc Airport}}  & 3,188  & 18,631 & 4       & 4             \\
{{\sc Citeseer}} & 3,327  & 4,732  & 6       & 3,703          \\
{{\sc Cora}}     & 2,708  & 5,429  & 7       & 1,433          \\ 
{TREE-L}   & 1,093  & 1,092  & 4       & 32          \\ 
{TREE-H}   & 1,093  & 1,092  & 4       & 32          \\ \bottomrule
\end{tabular} 
}
\label{tab:statistics}
\end{table}

\subsection{Training Details}
\label{sec:training_details}
(1) \textbf{Data split.} 
We conducted experiments on both node classification and link prediction tasks. For the link prediction task, we randomly split the edges in the {\sc Disease} dataset into training (75\%), validation (5\%), and test (20\%) sets for the shallow models. In the semi-supervised learning setting, we split the {\sc Disease} dataset into training (25\%), validation (5\%), and test (70\%) sets.
For node classification, we split the nodes in the {\sc Airport} dataset into 70\%, 15\%, and 15\%, and the nodes in the {\sc Disease} dataset into 30\%, 10\%, and 60\%. For the {\sc Cora} and {\sc Citeseer} datasets, we used 20 labeled examples per class for training. The above splits are the same as those used in previous works, except for the semi-supervised learning.

(2) \textbf{Implementation details.} For all models, we traverse the number of embedding dimensions from 8, 64, 256 and then perform a hyper-parameter search on a validation set over learning rate $\{0.01, 0.02, 0.005\}$, weight decay $\{1e-4, 5e-4, 5e-5\}$, dropout $\{0.1, 0.2, 0.5, 0.6\}$, and the number of layers $\{1, 2, 3,4,5\}$. We also adopt the early stopping strategies based on the validation set with patience in $\{100, 200, 500\}$. 

(3) \textbf{Evaluation metric.}
Following the literature, we report the F1-score for {\sc Disease} and {\sc Airport} datasets and accuracy for the others in the node classification tasks. For the link predictions task, the Area Under Curve(AUC) is calculated. We report the results of 12 random experiments by removing the max and min values. 

(4) \textbf{Loss functions.}
In this study, we conducted two different experimental tasks, link prediction, and node classification. For link prediction, we used the following loss function as~\cite{hgcn2019,nickel2017poincare},
	\begin{equation} 
	\label{HGCN: lp_loss_fun}
	L_\mathrm{lp} = \frac{1}{|E|}\sum_{(i,j)\in E}-\mathrm{log}~p(\mathbf{z}_{i}, \mathbf{z}_{j}) + \frac{1}{|E|}\sum_{(i,j)\notin E'} \mathrm{log}~p(\mathbf{z}_{i}, \mathbf{z}_{j}'),
	\end{equation}
where $E$ is the  edge set and $p(\cdot)$ is the Fermi-Dirac function, indicating the probability of two hyperbolic nodes $\mathbf{u}, \mathbf{v}$ have a link or not, which is defined as:
	\begin{equation}
	\label{equ:femi-dirac}
	p(\mathbf{u}, \mathbf{v})=\left[\exp{(d_\mathcal{H} (\mathbf{u}, \mathbf{v})^{2}-r) / t}+1\right]^{-1},
	\end{equation}
The loss function is to maximize the probability of two nodes if they are linked in the training set while minimizing the probability of two nodes if they are not linked in the training set. We sample the same number of negative sampling for training.

For node classification, we follow the previous work~\cite{hgcn2019}, and classify the node by cross-entropy, which is formulated by,
\begin{equation}
    L_\mathrm{ce} = -\sum_{i\in |V|}q(\mathbf{z}_i)\log q(\mathbf{z}_i)
\end{equation}
where 
\begin{equation}
    q(\mathbf{z}) = \mathrm{Softmax}(\mathrm{MLP}(\log_\mathbf{o}^\kappa(\mathbf{z})))
\end{equation}

Totally, our optimization target is to minimize the following loss function,
\begin{equation}
    L = L_\mathrm{task} + \lambda L_\mathrm{hyp},
\end{equation}
where the task denotes the down-stream task, and $\lambda$ is selected from $\{1, 0.1, 0.01, 0.001\}$

\begin{table}[]
\centering
\caption{Comparisons using stretching, alignment, opposite stretching and our proposed method \method with dimension 64.}
\vspace{5pt}
\resizebox{0.57\textwidth}{!}{%
\begin{tabular}{@{}lccccc@{}}
\toprule
Dataset & HGCN     & w/Stretching & w/Alignment &  w/-Stretching & \method(Ours) \\ \midrule
{\sc disease}  & $90.6\pm1.2$ & $89.4\pm2.5$        & $95.5\pm0.6$            & $85.7\pm4.8$        & $\textbf{95.0}\pm0.8$  \\
{\sc Airport}  & $94.0\pm0.6$ & $93.6\pm0.9$        & $93.6\pm0.3$            & $92.1\pm0.5$        & $\textbf{94.1}\pm0.8$  \\
{\sc Citeseer} & $67.6\pm0.4$ & $68.5\pm0.6$        & $66.7\pm0.9$            & $18.1\pm0.0$        & $\textbf{74.1}\pm0.3$  \\
{\sc Cora}     & $78.5\pm0.6$ & $81.5\pm0.5$        & $77.5\pm1.0$            & $31.9\pm0.0$        & $\textbf{83.0}\pm0.3$  \\
\bottomrule
\end{tabular}%
}
\label{table: inverse_stretching}
\end{table}
\section{Experiments on Opposite Stretching}
\label{appendix:Opposite_stretching}
The capacity of hyperbolic space increases exponentially with the radius, in this work, we propose to push nodes away from the origin and thus introduce a origin-concerned geometric penalty. In this part, we supplement the opposite stretching operation, which pushes the node hidden state close to the origin, to further verify our motivation. Specifically, we use the following equation to achieve the opposite stretching:
\begin{equation}
       {L}_\text{hyp}^{-} = \tanh\left(\frac{1}{|V|}\sum_{i\in V} w_id_\mathcal{H}({\mathbf{z}_i^\mathcal{H}},\mathbf{o})\right).
       \label{equ: direct stretching}
\end{equation}

We ran all experiments 12 times with the same settings as the previous, including the random seeds. The experimental results are shown in Table~\ref{table: inverse_stretching}. To straightforwardly compare the performance of the opposite stretching and other related methods, we list the independent stretching (w/-Stretching), alignment (w/Alignment), and the proposed \method together. The performance of using opposite stretching is shown in the "w/Stretching" column.  

It can be seen that the opposite stretching (w/-Stretching), or equivalently, encouraging the node embedding near the origin, is fatal to {\sc Cora} and {\sc Citeseer} where the performance drops to 31.9 and 18.1. At the same time, the performances on {\sc disease} and {\sc Airport} also drop dramatically. This is not difficult to understand since the embedding space near the origin is relatively small, and it is difficult for the model to obtain a better discriminative representation. This further verifies the effectiveness of our proposal.

\section{More Results}
\subsection{HDO on Different Dimensions}
\label{appendix:more_experimental_results}
\begin{figure}[!tp]
\subfigure{
\begin{minipage}[t]{0.329\linewidth}
\centering
\includegraphics[width=1.75in]{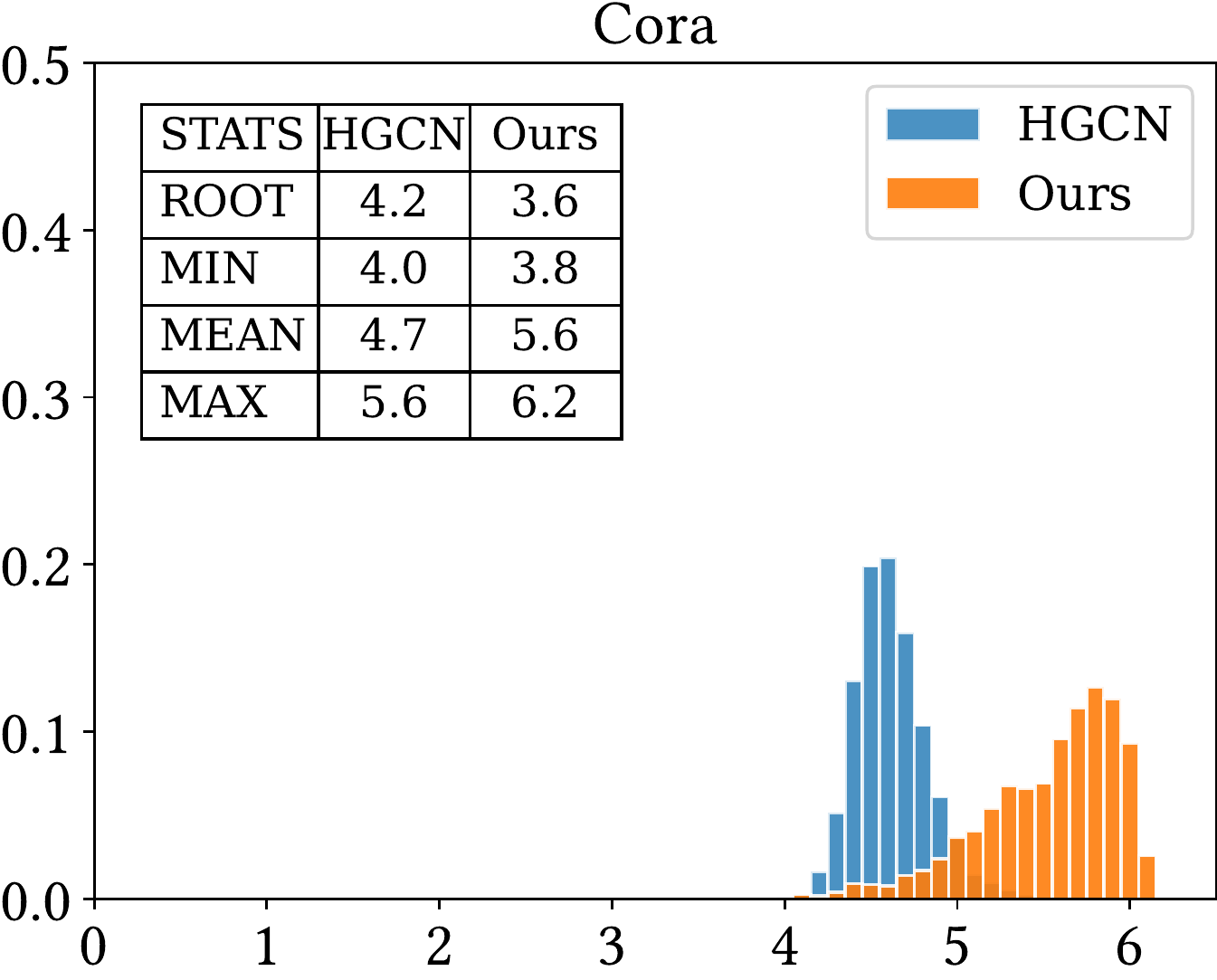}
\end{minipage}%
}%
\subfigure{
\begin{minipage}[t]{0.329\linewidth}
\centering
\includegraphics[width=1.75in]{figures/hdo/icml2023_cora_64.pdf}
\end{minipage}%
}%
\subfigure{
\begin{minipage}[t]{0.329\linewidth}
\centering
\includegraphics[width=1.75in]{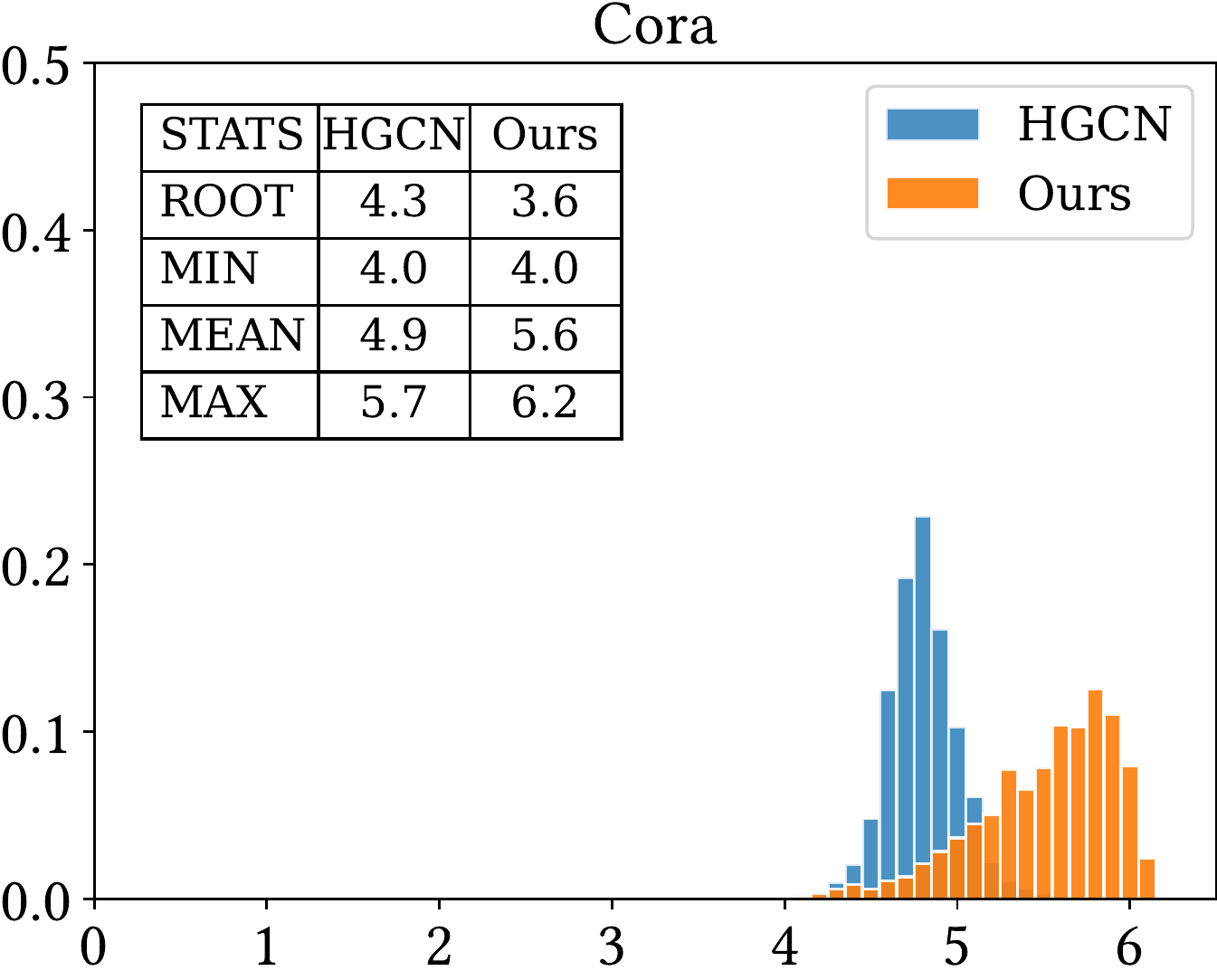}
\end{minipage}%
}%

\subfigure{
\begin{minipage}[t]{0.329\linewidth}
\centering
\includegraphics[width=1.75in]{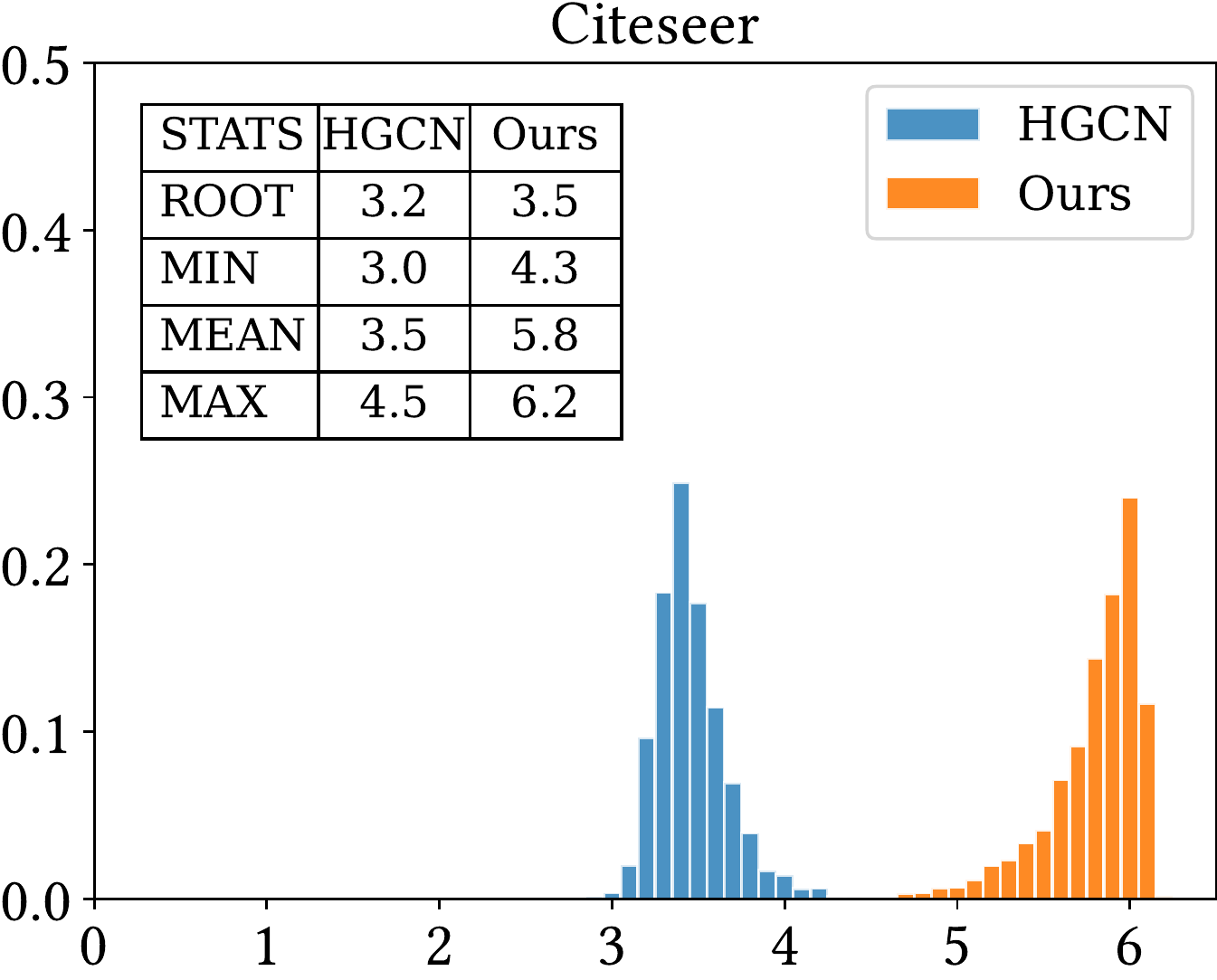}
\end{minipage}%
}%
\subfigure{
\begin{minipage}[t]{0.329\linewidth}
\centering
\includegraphics[width=1.75in]{figures/hdo/icml2023_citeseer_64.pdf}
\end{minipage}%
}%
\subfigure{
\begin{minipage}[t]{0.329\linewidth}
\centering
\includegraphics[width=1.75in]{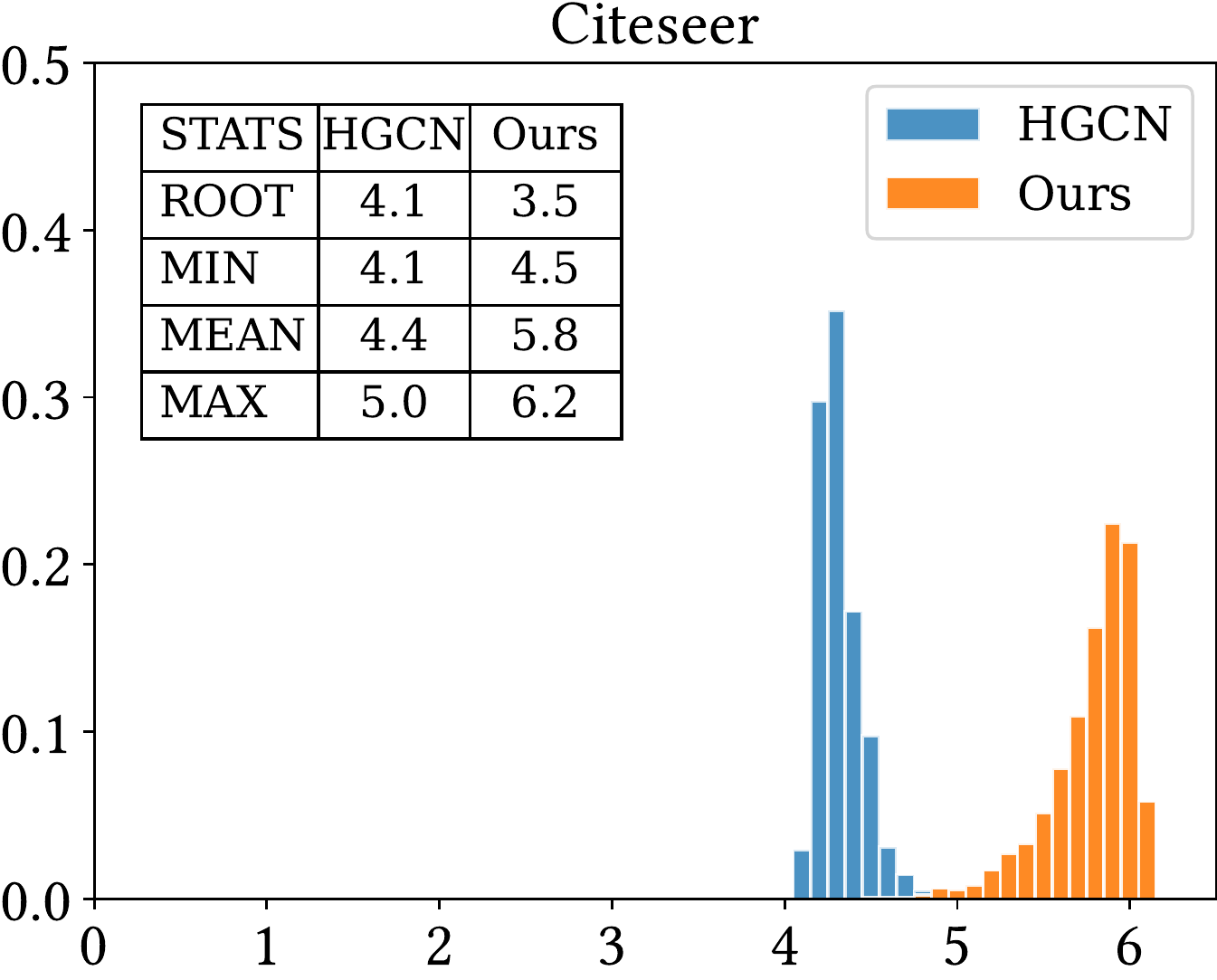}
\end{minipage}%
}%

\subfigure{
\begin{minipage}[t]{0.329\linewidth}
\centering
\includegraphics[width=1.75in]{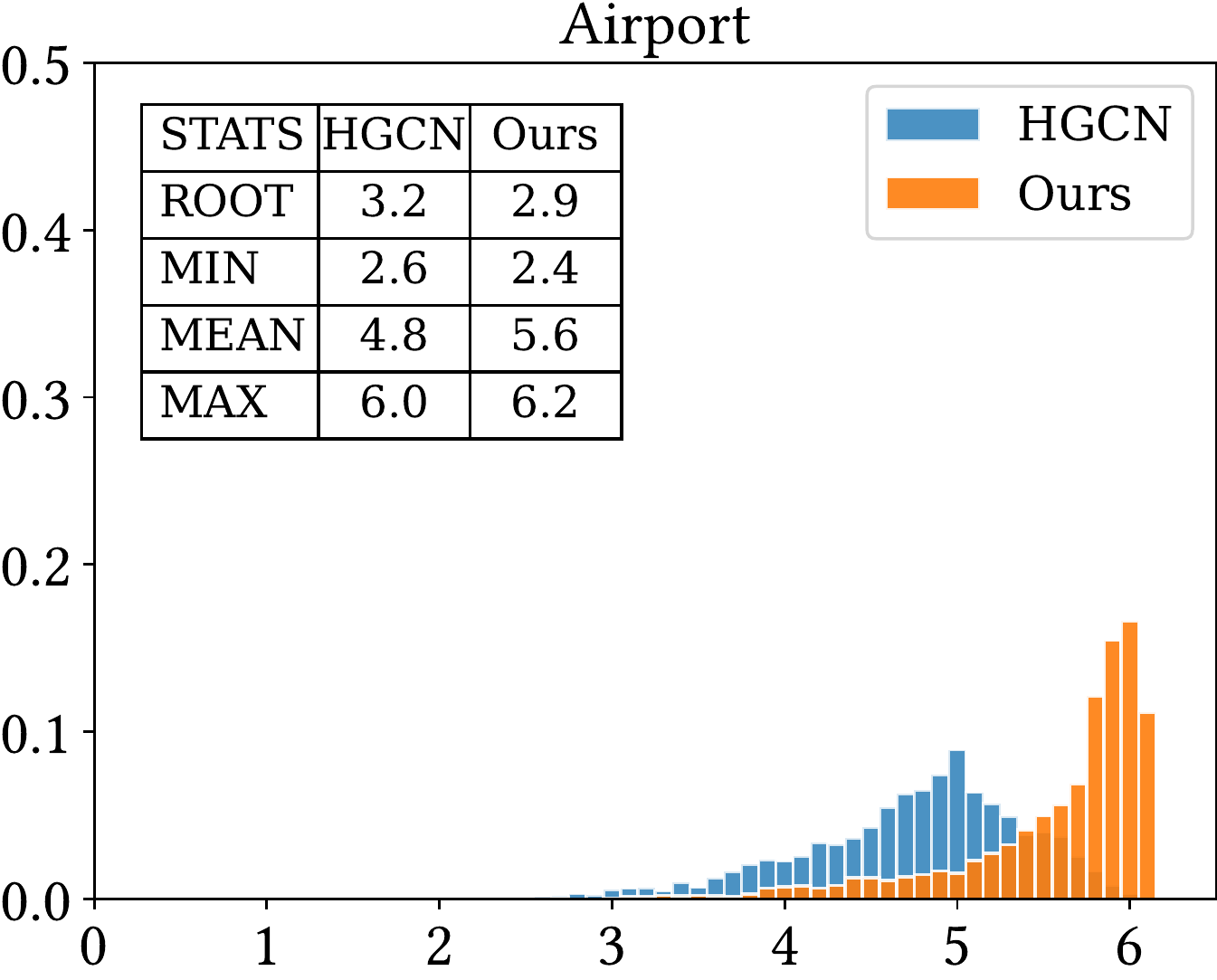}
\end{minipage}%
}%
\subfigure{
\begin{minipage}[t]{0.329\linewidth}
\centering
\includegraphics[width=1.75in]{figures/hdo/icml2023_airport_64.pdf}
\end{minipage}%
}%
\subfigure{
\begin{minipage}[t]{0.329\linewidth}
\centering
\includegraphics[width=1.75in]{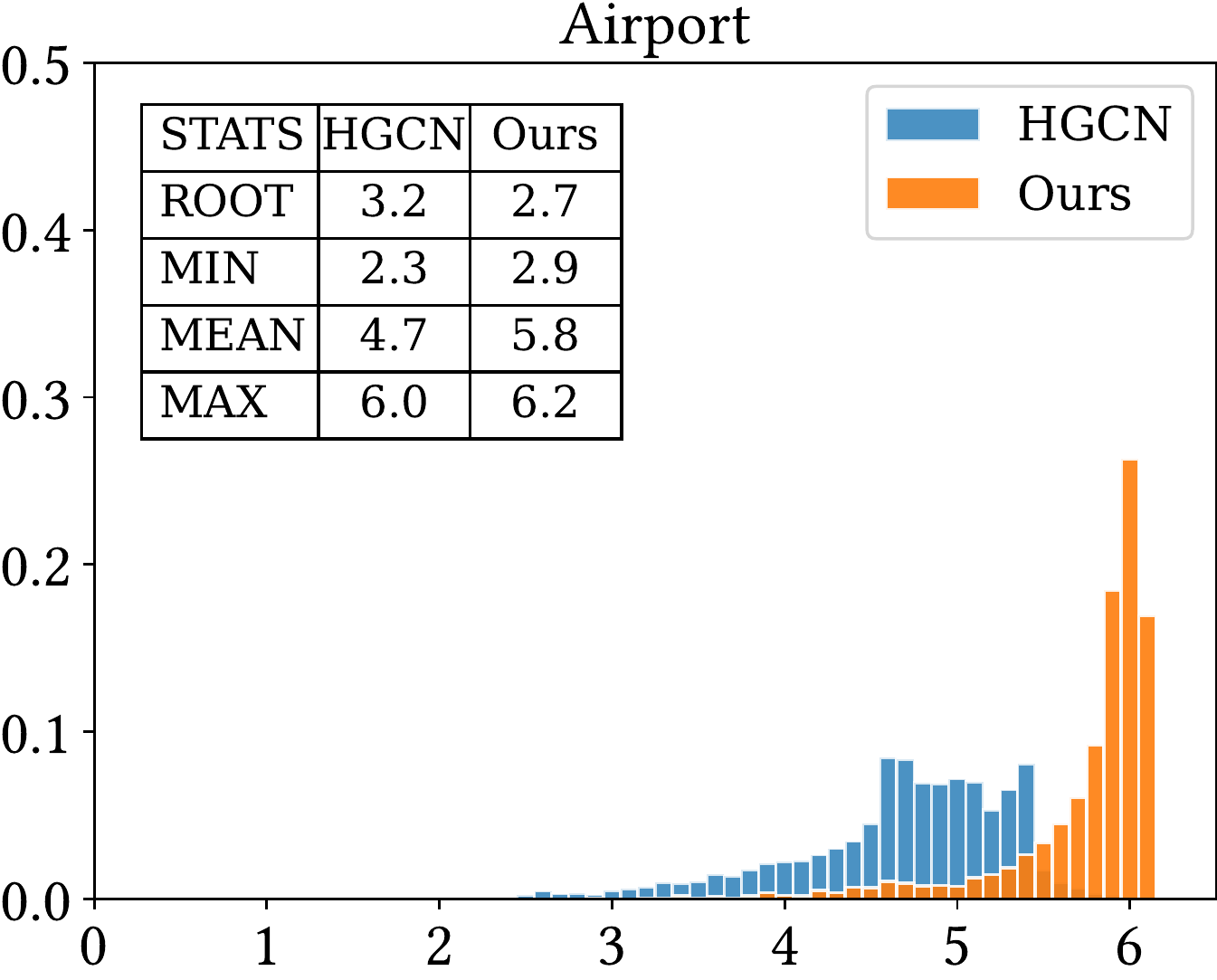}
\end{minipage}%
}%

\subfigure{
\begin{minipage}[t]{0.329\linewidth}
\centering
\includegraphics[width=1.75in]{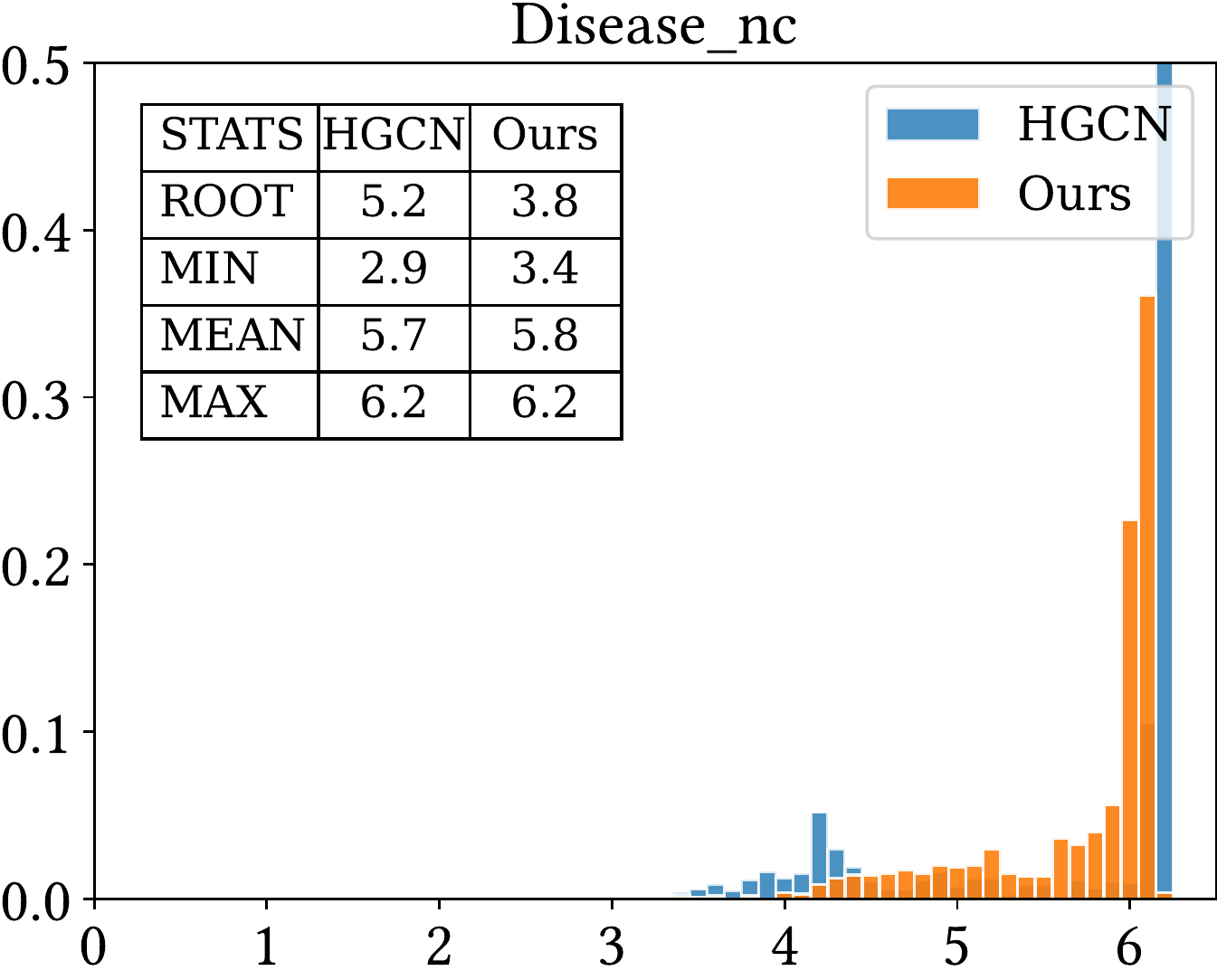}
\end{minipage}%
}%
\subfigure{
\begin{minipage}[t]{0.329\linewidth}
\centering
\includegraphics[width=1.75in]{figures/hdo/icml2023_disease_nc_64.pdf}
\end{minipage}%
}%
\subfigure{
\begin{minipage}[t]{0.329\linewidth}
\centering
\includegraphics[width=1.75in]{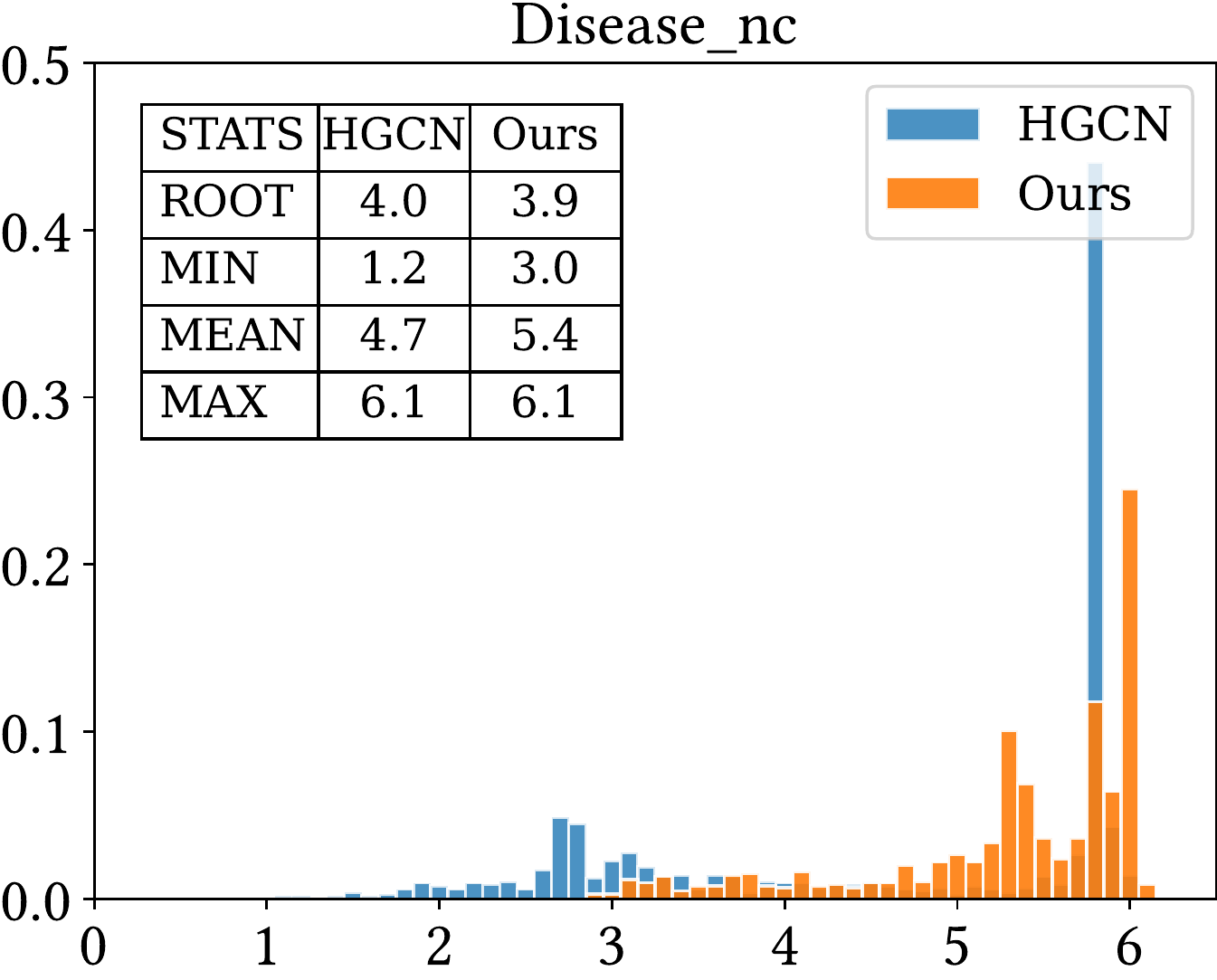}
\end{minipage}%
}%

\caption{Illustration of HDO distribution on  {\sc Cora}, {\sc Citeseer}, {\sc Airport}, Disease where x-axis denotes the value of HDO and y-axis is the corresponding ratio. The figures in the first, second, and third column denotes the dimension 16, 64 and 256, respectively.}
\vspace{-10pt}
\label{fig: hdo_dist_more}
\end{figure}

Here we additionally supplement the HDO distribution on different dimensions,  in Figure~\ref{fig: hdo_dist_more}. It is easy to know that the overall HDO values become larger when applied with our \method, which can be inferred from the shapes of HDO distribution or MEAN values. It is also observed that the  number of the nodes whose embeddings are located near the boundary has increased significantly, as the shape of the HDO distribution looks more like a long-tailed distribution, which well matches the exponential increase in the capacity of the hyperbolic space. 

We observe that the distribution of HDO of {\sc disease} is more concentrated in locations far from the origin in both our method and the original model, which mainly lies in the properties of the {\sc disease} dataset. The {\sc disease} dataset is pure tree-like, which has obvious hierarchies and can well match the hyperbolic space. Therefore, with hyperbolic embedding, the overall HDO is distributed far from the origin. This can explain the reason that only using centering for HGCN achieves considerable improvements on {\sc disease}, that is, the leaf nodes have already embedded near the boundary. However, it is also noted that the embedding position of ROOT (i.e., HC) is quite different: the ROOT in HGCN is far from the origin, while our proposal is at or near the origin. Our method is significantly better than the original HGCN. It further confirms that aligning the embedding center with the hyperbolic origin is of significance for the embedding quality and classification performance.  

\subsection{Hyperbolic Distance to HC}
\begin{figure}[!tp]
\subfigure{
\begin{minipage}[t]{0.329\linewidth}
\centering
\includegraphics[width=1.75in]{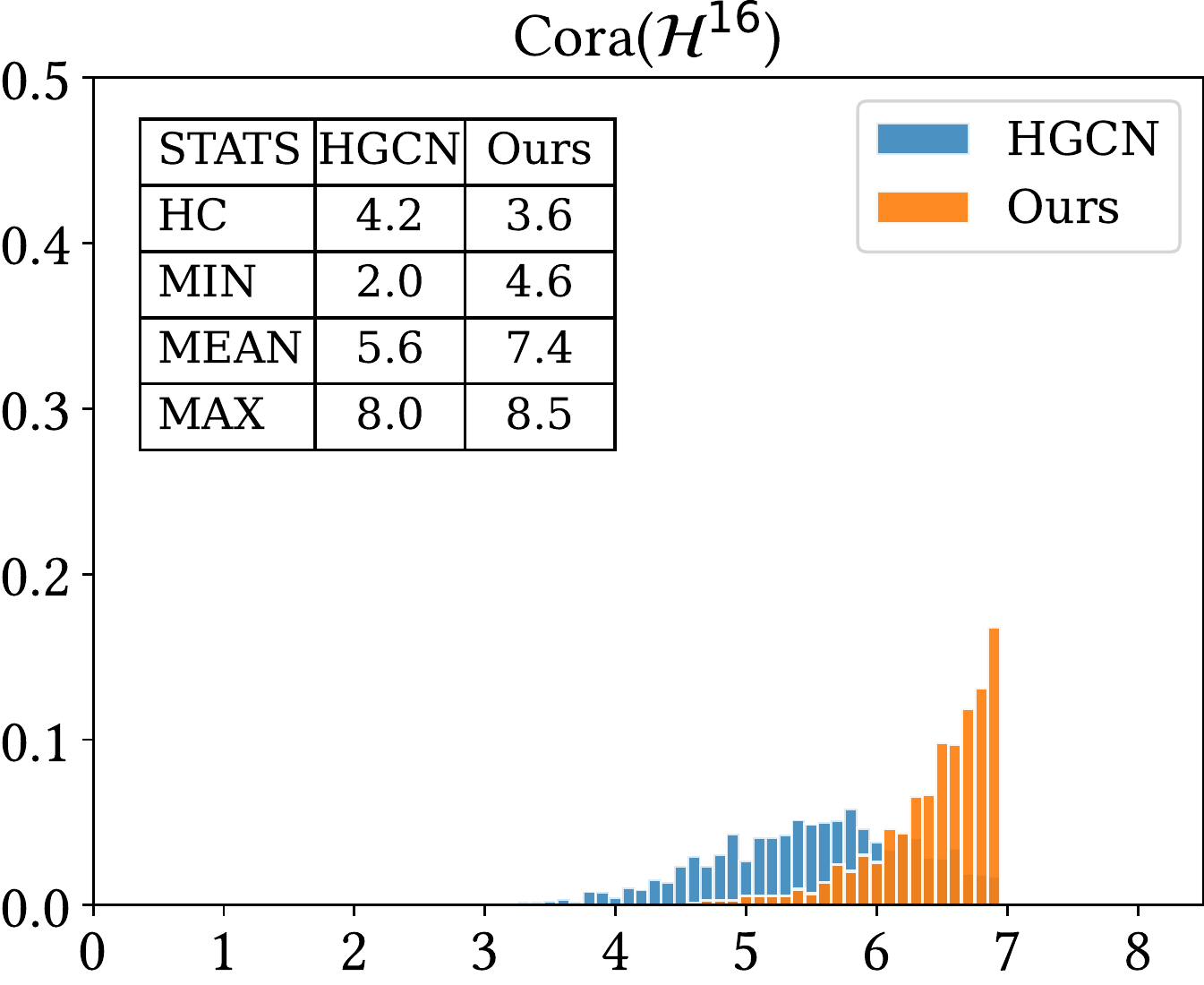}
\end{minipage}%
}%
\subfigure{
\begin{minipage}[t]{0.329\linewidth}
\centering
\includegraphics[width=1.75in]{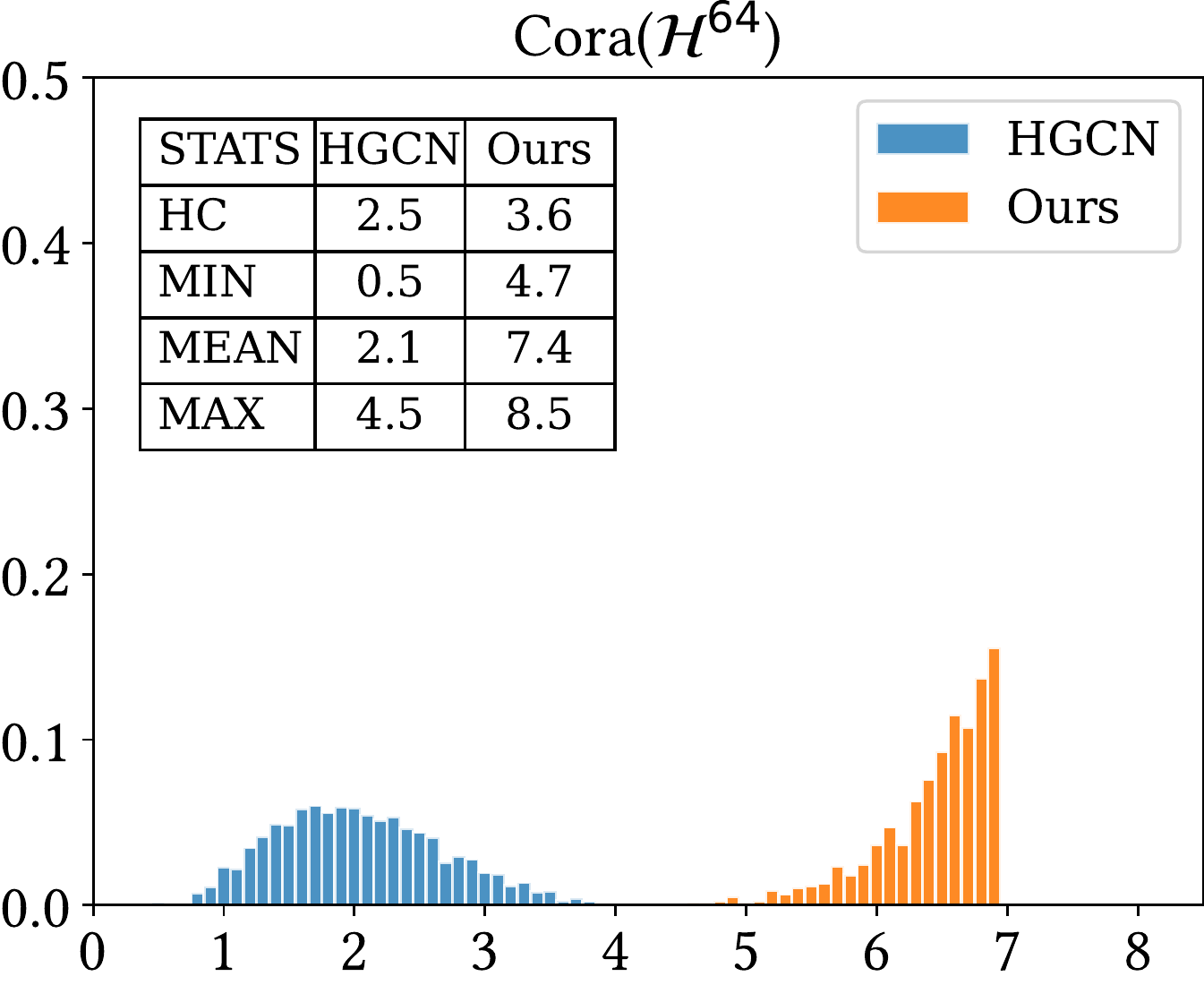}
\end{minipage}%
}%
\subfigure{
\begin{minipage}[t]{0.329\linewidth}
\centering
\includegraphics[width=1.75in]{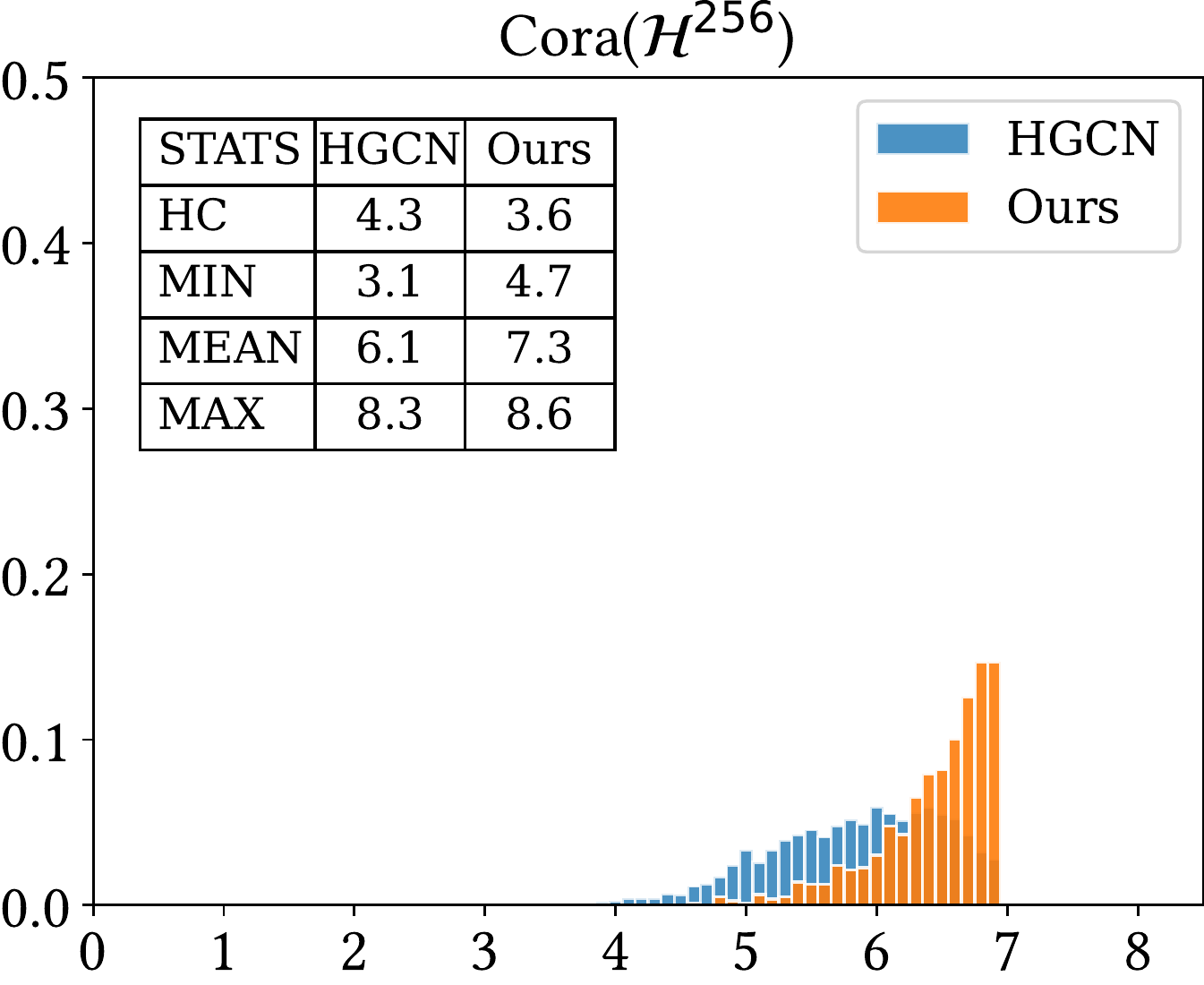}
\end{minipage}%
}%

\subfigure{
\begin{minipage}[t]{0.329\linewidth}
\centering
\includegraphics[width=1.75in]{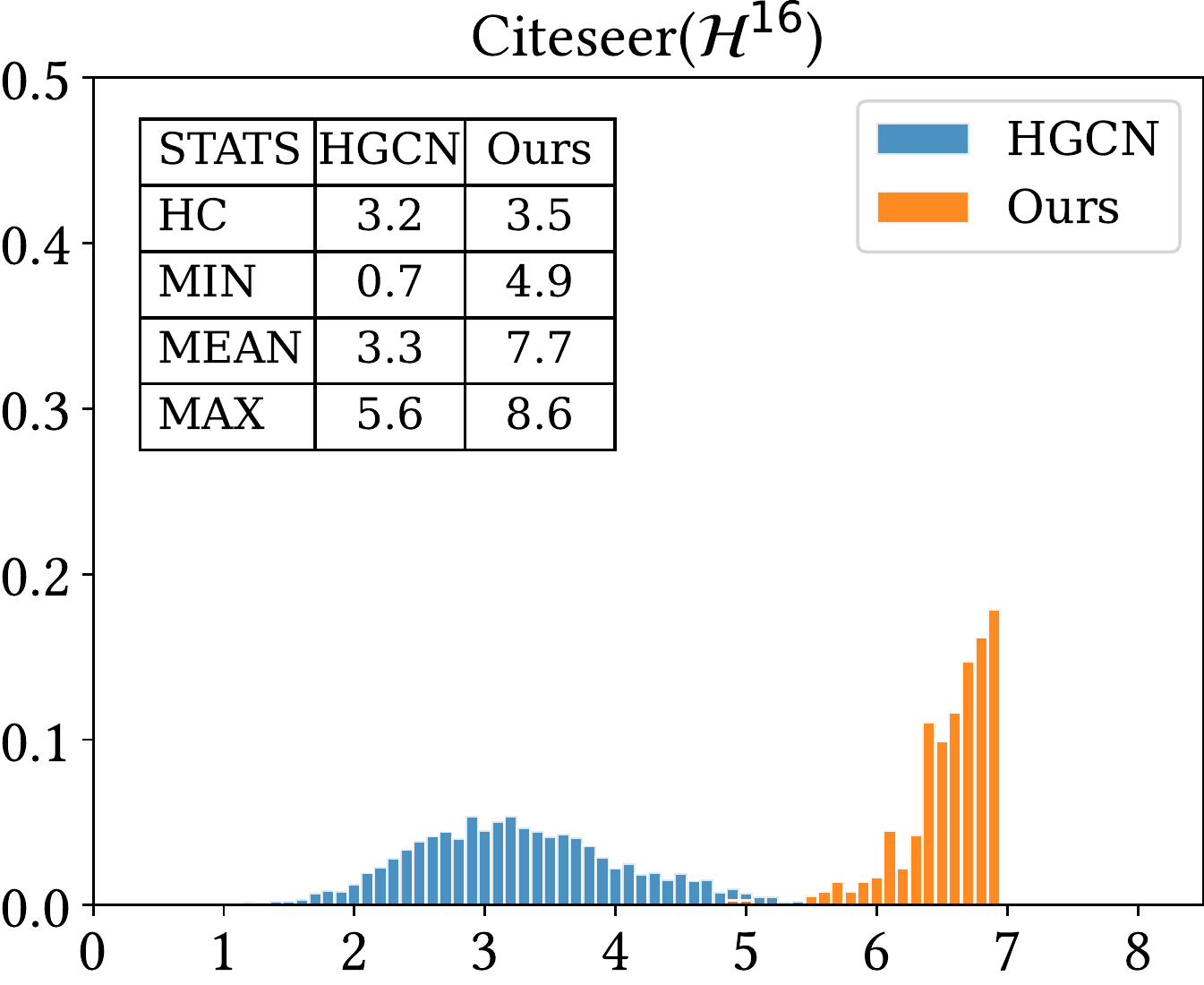}
\end{minipage}%
}%
\subfigure{
\begin{minipage}[t]{0.329\linewidth}
\centering
\includegraphics[width=1.75in]{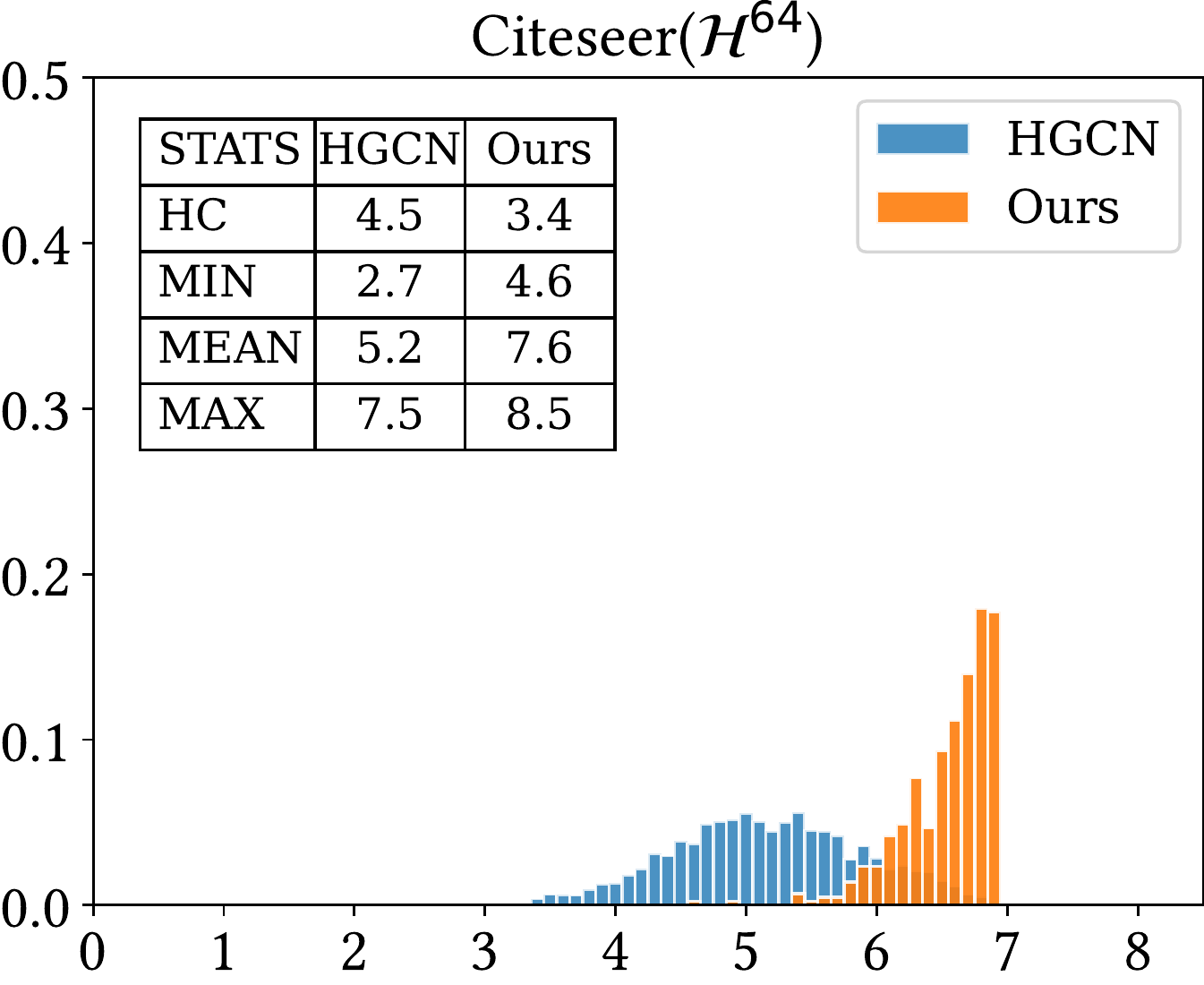}
\end{minipage}%
}%
\subfigure{
\begin{minipage}[t]{0.329\linewidth}
\centering
\includegraphics[width=1.75in]{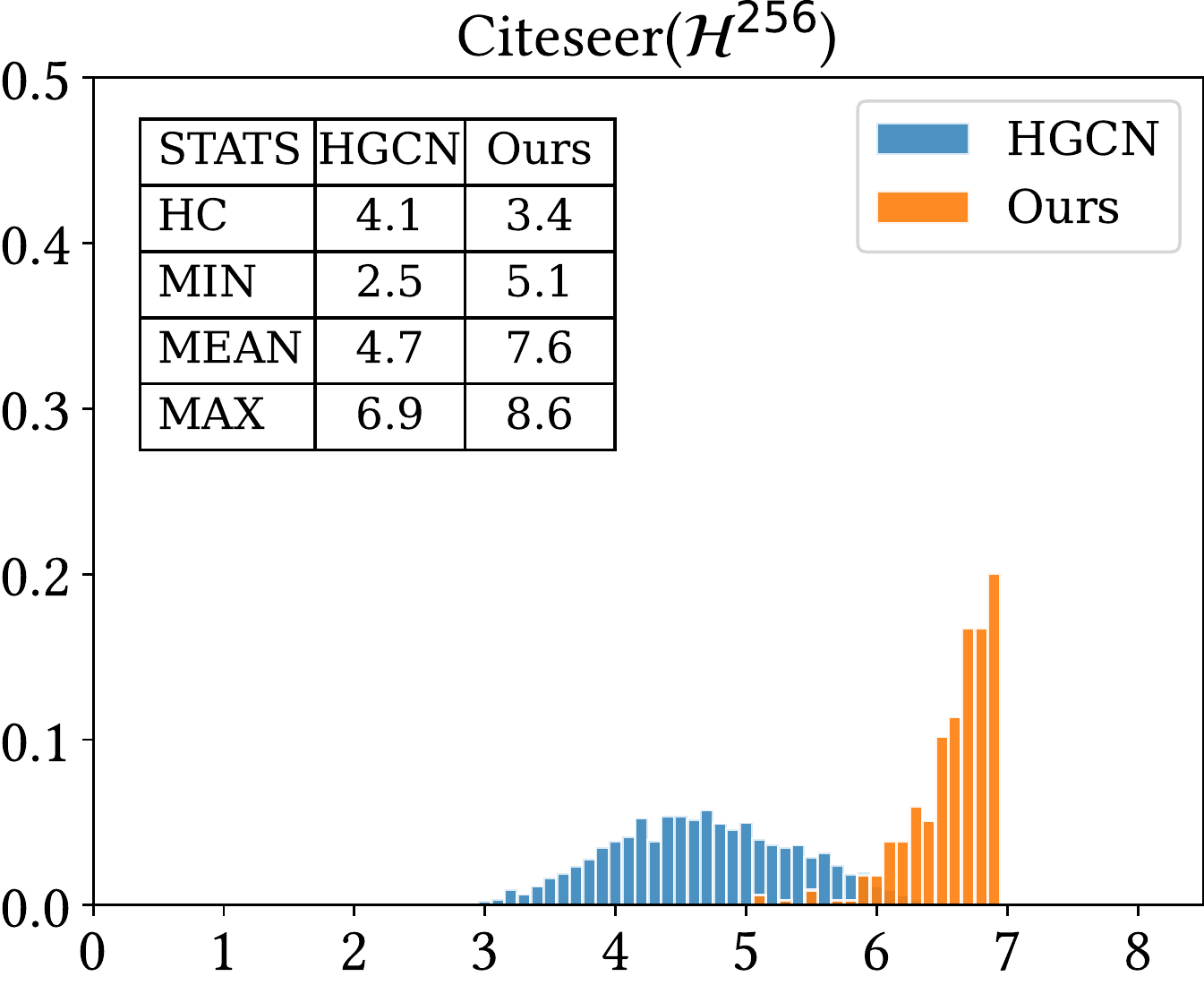}
\end{minipage}%
}%

\subfigure{
\begin{minipage}[t]{0.329\linewidth}
\centering
\includegraphics[width=1.75in]{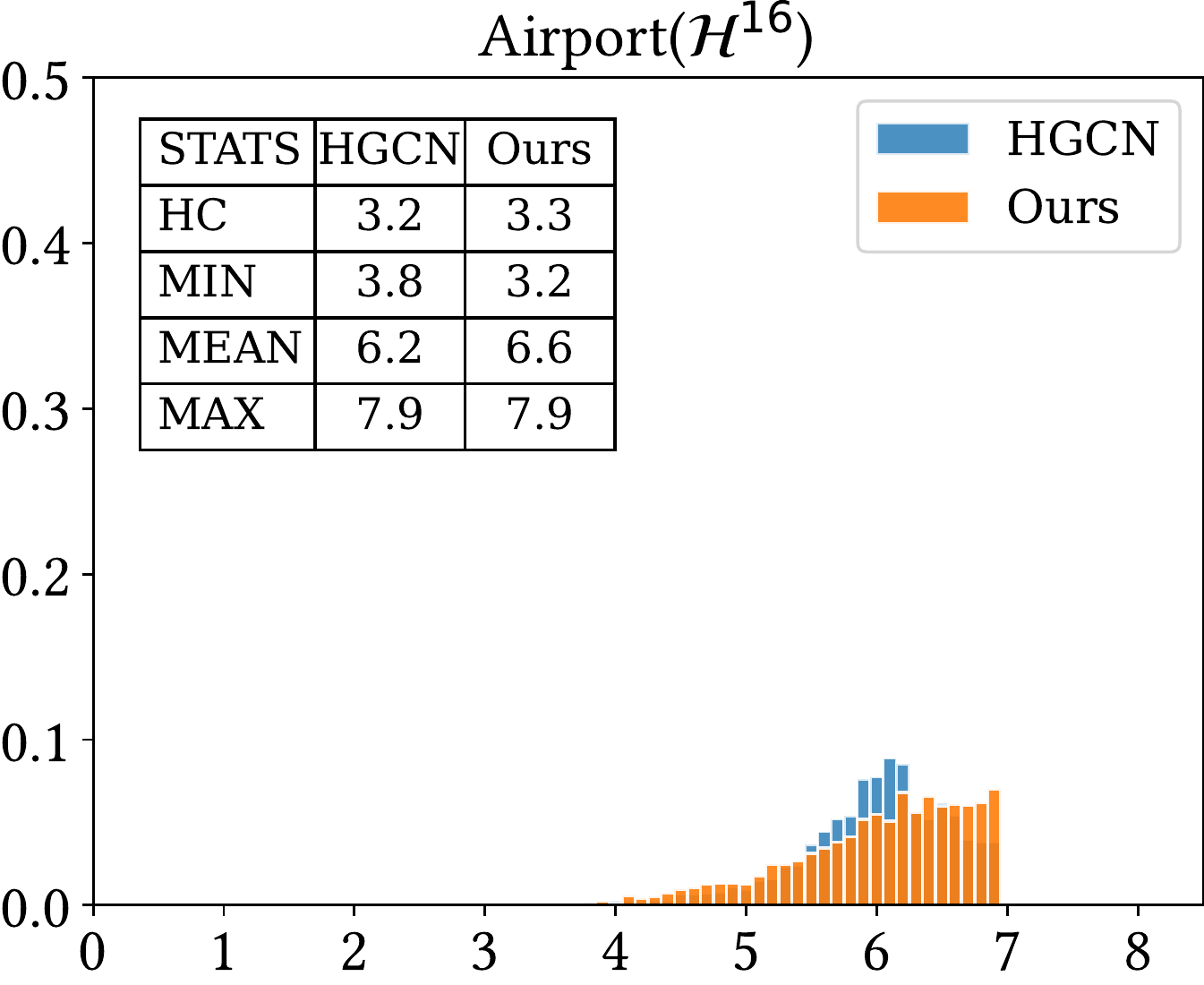}
\end{minipage}%
}%
\subfigure{
\begin{minipage}[t]{0.329\linewidth}
\centering
\includegraphics[width=1.75in]{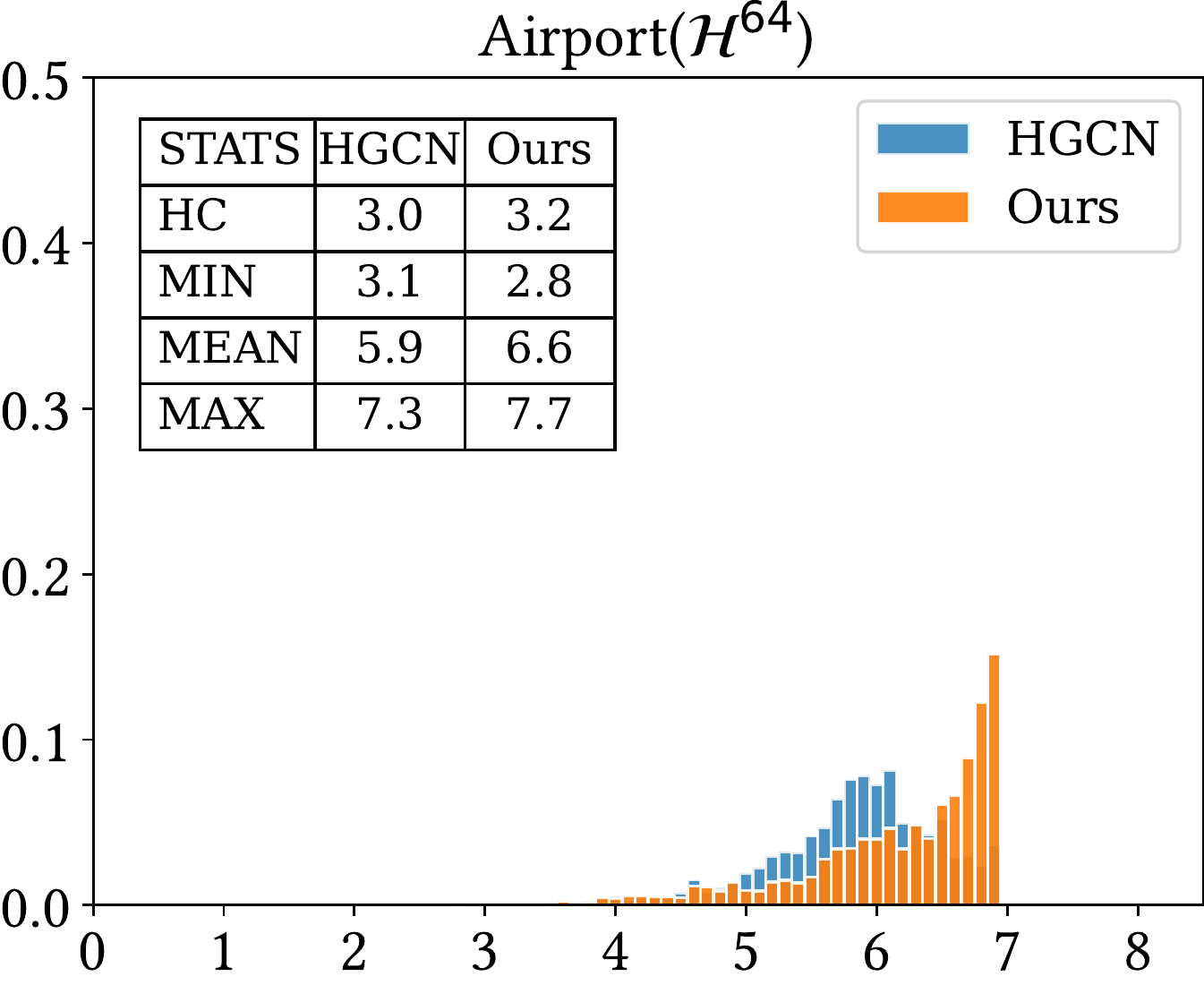}
\end{minipage}%
}%
\subfigure{
\begin{minipage}[t]{0.329\linewidth}
\centering
\includegraphics[width=1.75in]{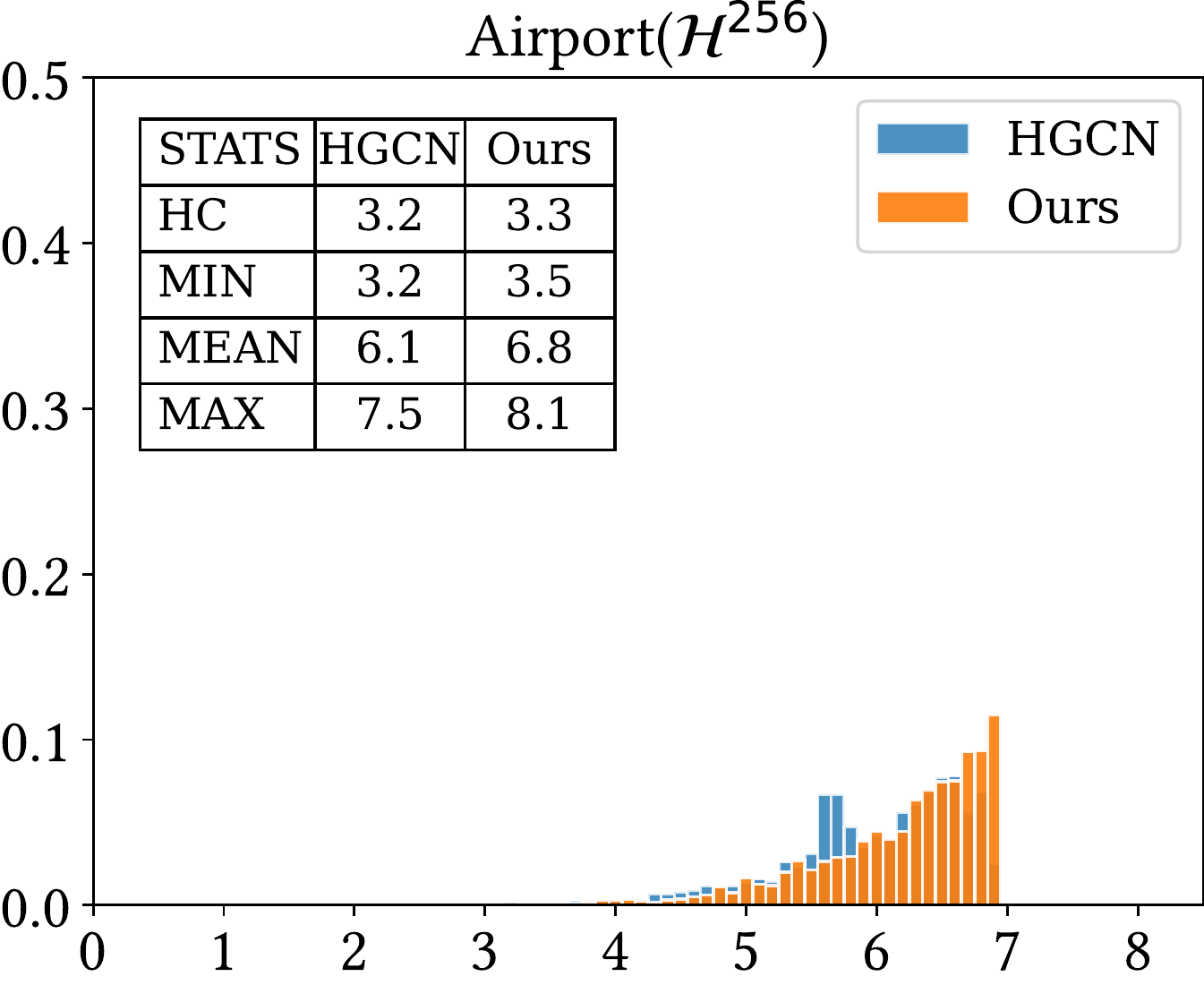}
\end{minipage}%
}%

\subfigure{
\begin{minipage}[t]{0.329\linewidth}
\centering
\includegraphics[width=1.75in]{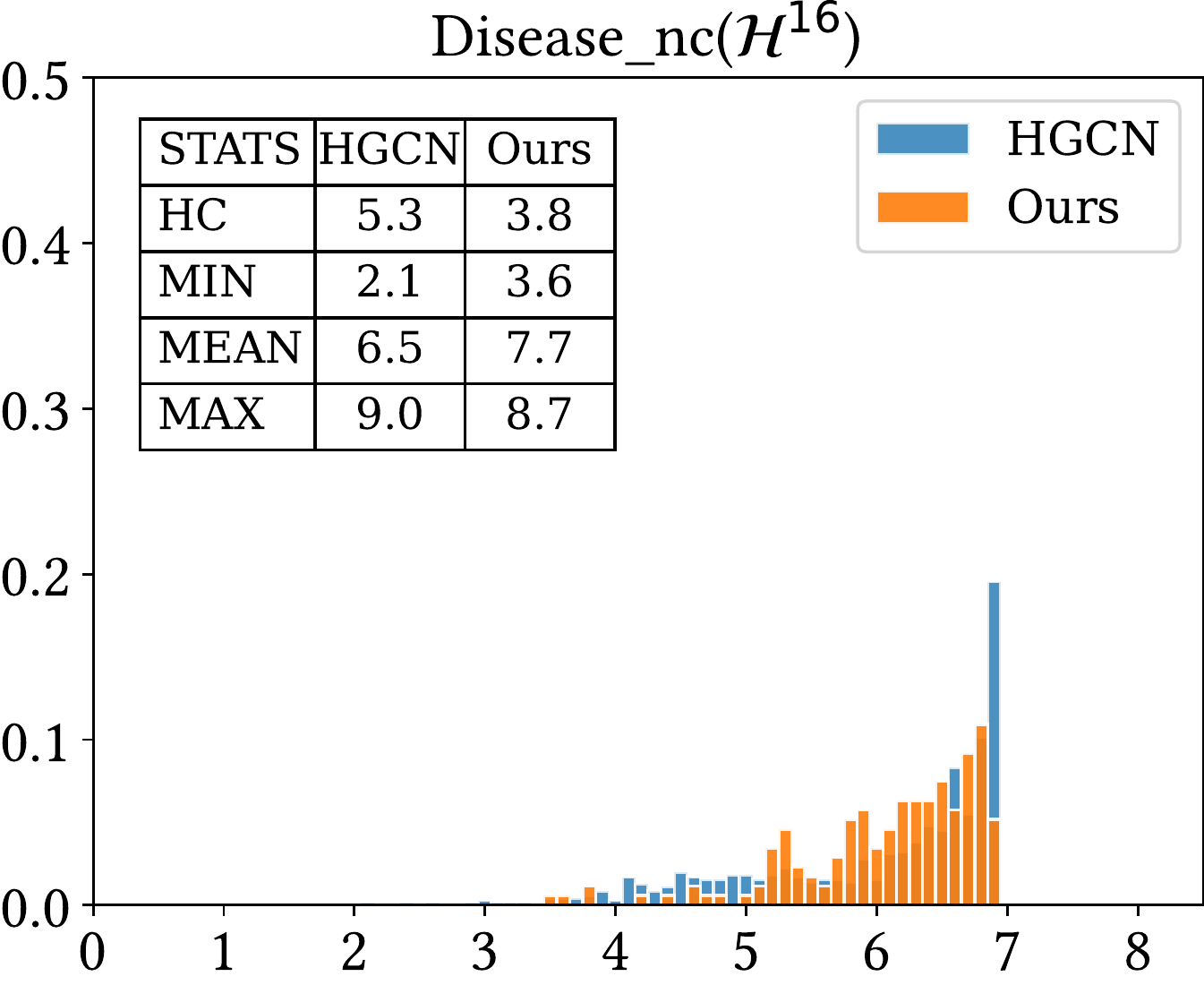}
\end{minipage}%
}%
\subfigure{
\begin{minipage}[t]{0.329\linewidth}
\centering
\includegraphics[width=1.75in]{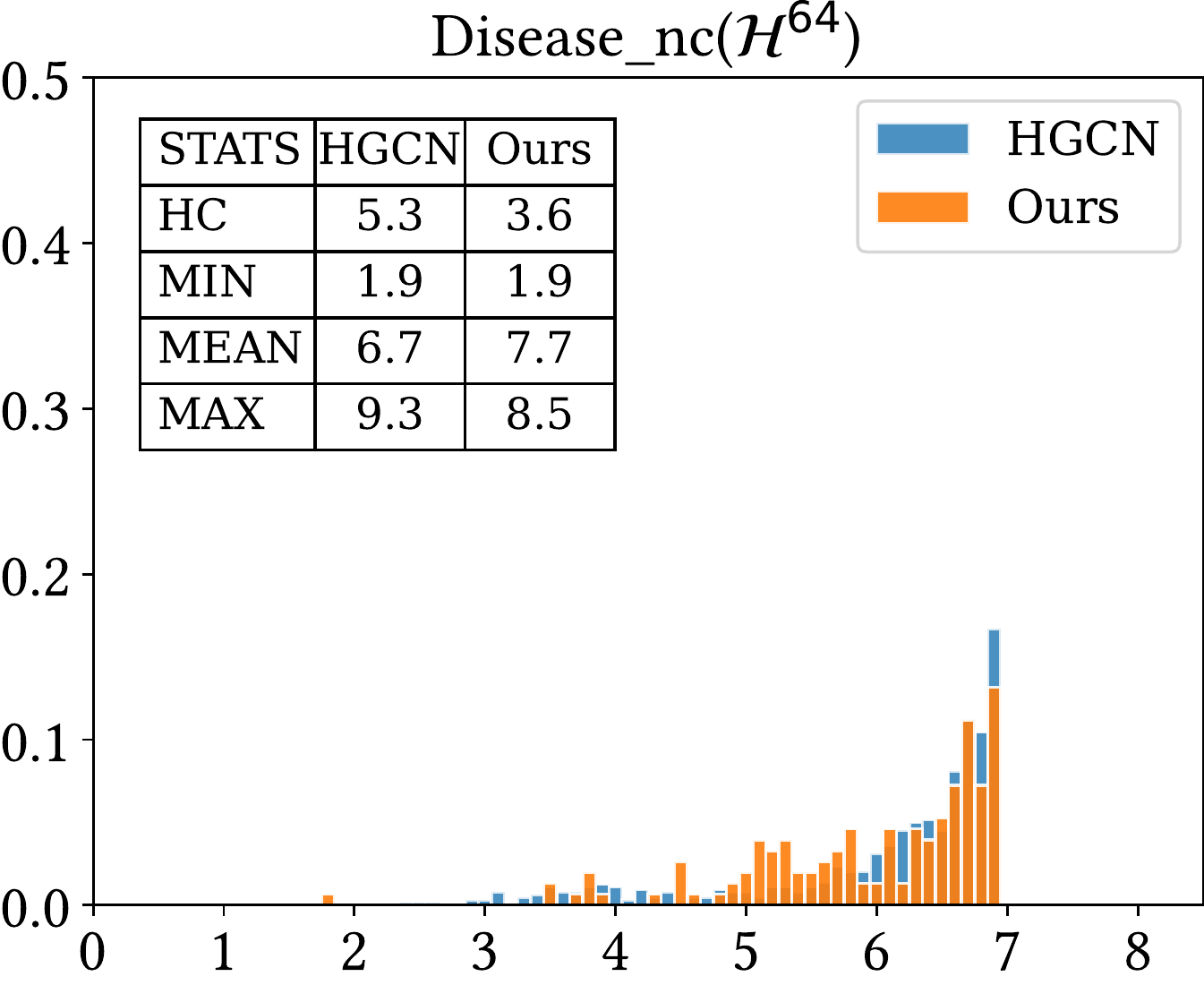}
\end{minipage}%
}%
\subfigure{
\begin{minipage}[t]{0.329\linewidth}
\centering
\includegraphics[width=1.75in]{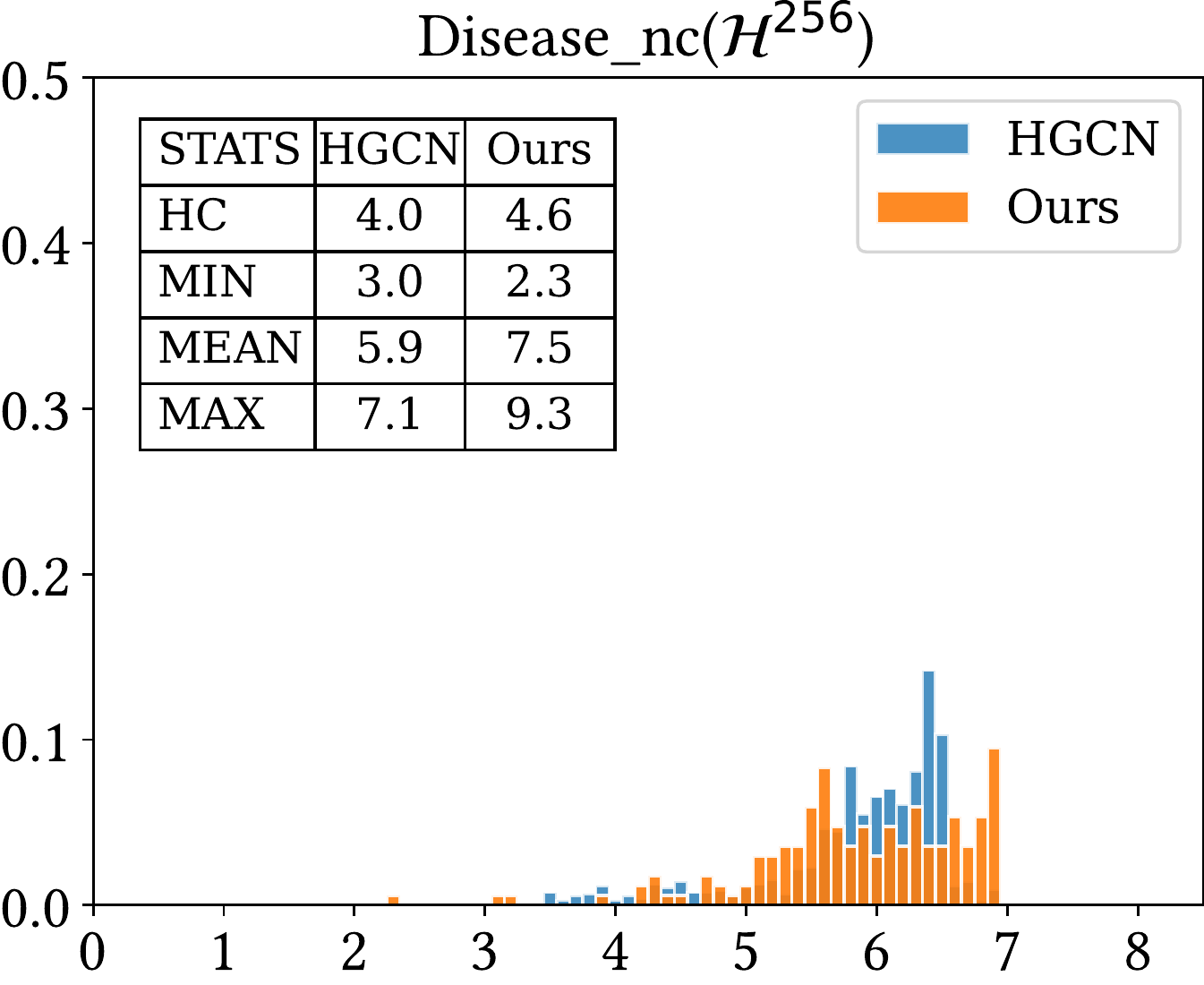}
\end{minipage}%
}%

\caption{Illustration of HDC distribution on  {\sc Cora}, {\sc Citeseer}, {\sc Airport}, Disease where x-axis denotes the value of HDC and y-axis is the corresponding ratio. The figures in the first, second, and third column denotes the dimension 16, 64 and 256, respectively.}
\vspace{-10pt}
\label{fig: hdc_dist_more}
\end{figure}

Besides, we further compute the hyperbolic distance from each node to the hyperbolic center in Figure~\ref{fig: hdc_dist_more}. In the figure above, HC represents an absolute value indicating the distance from the center to the origin, while MIN, MEAN, and MAX represent relative values reflecting the distance to the HC.
Our experiments have revealed that previous models are unable to capture the hierarchical relationships within the data from the HDC distribution effectively, as most of them are normally distributed.
In contrast, our proposed approach consistently produced relatively larger values for the relative MIN, MEAN, and MAX, indicating that it can effectively fit the hyperbolic space.

\subsection{Relative Hierarchies within Nodes}

\begin{table}[ht]
\centering
\caption{Accuracy of relative hierarchies within nodes}
\label{tab:relative_hierarchies}
\vspace{5pt}
\resizebox{0.32\columnwidth}{!}{%
\begin{tabular}{@{}ccccc@{}}
\toprule
Models & HGCN   & Ours   & HGCN   & Ours   \\
Dims   & 16     & 16     & 256    & 256    \\ \midrule
TREE-L & 72.5\% & 75.4\% & 75.3\% & 77.6\% \\
TREE-H & 68.9\% & 80.0\% & 73.7\% & 83.5\% \\ \bottomrule
\end{tabular}%
}
\end{table}

In the following, we compared hierarchies accuracy based on synthetic datasets, which provide easily accessible hierarchies, using both HGCN and our proposed method. We randomly selected 5000 pairs of nodes and computed their distance from the origin as a proxy for their respective hierarchical levels. We labeled a pairing as ``1" if its hierarchical ordering matched the true relationship and ``0" otherwise. Finally, we computed the accuracy of the sampled node pairs and presented the results in Table~\ref{tab:relative_hierarchies}.
The results demonstrate that our proposed method outperforms HGCN in ranking more pairs correctly, indicating its ability to effectively capture the hierarchical relationships within the data.

\subsection{Similar Idea Exp oration in Euclidean Space}
The method proposed in this work is simple and effective. Can the same effect be achieved if the same idea is adopted in the Euclidean space? In the following, we conduct the same algorithm presented in Algorithm~\ref{alg:hie} within Euclidean space, termed as EIE, on the best Euclidean model GAT.

\begin{table}[ht]
\centering
\caption{Comparisons with Euclidean variants}
\vspace{10pt}
\label{tab:eie}
\resizebox{0.65\columnwidth}{!}{%
\begin{tabular}{@{}lcccccccc@{}}
\toprule
Dataset   & {\sc Disease} & {\sc Disease} & {\sc Airport} & {\sc Airport} & {\sc Citeseer} & {\sc Citeseer} & {\sc Cora} & {\sc Cora} \\ \midrule
Dims & 16      & 256     & 16      & 256     & 16       & 256      & 16   & 256  \\
GAT       & 74.4    & 80.2    & 78.2    & 90.4    & 70.9     & 71.7     & 82.5 & 83.0 \\
GAT+EIE   & 79.1    & 81.5    & 78.6    & 90.8    & 71.2     & 72.2     & 82.8 & 83.0 \\
HGAT      & 88.8    & 89.8    & 92.9    & 95.2    & 66.8     & 67.9     & 79.0 & 79.1 \\
HGAT+HIE  & 93.7    & 94.0    & 93.0    & 95.7    & 73.4     & 74.4     & 83.0 & 83.4 \\ \bottomrule
\end{tabular}%
}
\end{table}

The experiments are shown in Table~\ref{tab:eie} and the experiments revealed that penalizing the Euclidean-centered norm of nodes can prevent nodes from getting too close to each other, which can impact classifier performance or downstream tasks, resulting in minor improvements across most cases. However, the capacity limitations of Euclidean space lead to inferior outcomes compared to using \method in hyperbolic space.

Moreover, we observed that while classification accuracy does not see significant improvement with increased dimensionality in Euclidean space, such an increase in hyperbolic space can enhance performance. These phenomena demonstrate important progress in hyperbolic space. First, \method enhances the hyperbolic models' embeddability, allowing them to fully utilize the spacious capacity of hyperbolic space and achieve superior performance across various datasets compared to Euclidean space. Second, as the embedding dimension increases, performance can still be improved, pushing the model to achieve better results, while their Euclidean counterparts show only minor improvements.

Overall, we believe that our study provides novel insights into developing innovative methodologies in hyperbolic space, and highlights the potential benefits of using hyperbolic geometry for machine learning tasks.

\end{document}